\documentclass[preprint,12pt,authoryear]{elsarticle}

\usepackage{amssymb}
\usepackage{amsmath}
\usepackage{natbib}
\usepackage{amsfonts,amssymb} 
\usepackage{multirow}
\usepackage{makecell}
\usepackage{color}
\usepackage[colorlinks,linkcolor=red,anchorcolor=blue,citecolor=green]{hyperref}

\usepackage{longtable}

\usepackage{graphicx}  
\usepackage{float}  
\usepackage{subcaption}  
\usepackage{lineno}

\journal{}

\begin{document}

\begin{frontmatter}



\title{Deep Learning Framework for History Matching CO$_2$ Storage with 4D Seismic and Monitoring Well Data}


\author[label1]{Nanzhe Wang\texorpdfstring{\corref{cor1}}{}}
\author[label1]{Louis J.~Durlofsky}

\affiliation[label1]{organization={Department of Energy Science and Engineering, \\ Stanford University},
            addressline={Stanford}, 
            state={CA},
            postcode={94305}, 
            country={USA}}

\cortext[cor1]{Corresponding author: nzwang@stanford.edu}

\begin{abstract}
Geological carbon storage entails the injection of megatonnes of supercritical CO$_2$ into subsurface formations. The properties of these formations are usually highly uncertain, which makes design and optimization of large-scale storage operations challenging. In this paper we introduce a history matching strategy that enables the calibration of formation properties based on early-time observations. Early-time assessments are essential to assure the operation is performing as planned. Our framework involves two fit-for-purpose deep learning surrogate models that provide predictions for in-situ monitoring well data and interpreted time-lapse (4D) seismic saturation data. These two types of data are at very different scales of resolution, so it is appropriate to construct separate, specialized deep learning networks for their prediction. This approach results in a workflow that is more straightforward to design and more efficient to train than a single surrogate that provides global high-fidelity predictions. The deep learning models are integrated into a hierarchical Markov chain Monte Carlo (MCMC) history matching procedure. History matching is performed on a synthetic case with and without 4D seismic data, which allows us to quantify the impact of 4D seismic on uncertainty reduction. The use of both data types is shown to provide substantial uncertainty reduction in key geomodel parameters and to enable accurate predictions of CO$_2$ plume dynamics. The overall history matching framework developed in this study represents an efficient way to integrate multiple data types and to assess the impact of each on uncertainty reduction and performance predictions.

\end{abstract}



\begin{keyword}
History matching \sep carbon storage \sep 4D seismic \sep deep learning


\end{keyword}

\end{frontmatter}

\section{Introduction}
\label{sec:intro}

Geological carbon storage has the potential to reduce substantially the amount of CO$_2$ emitted to the atmosphere. The ability to predict the performance of injection operations (e.g., pressure buildup, plume location) is essential for the gigatonne-scale deployment of this technology. Flow prediction is challenging, however, in large part because of the high degree of geological uncertainty associated with storage formations. Data assimilation procedures, often referred to as history matching in this setting, are typically applied in subsurface flow modeling to calibrate the geomodel based on observations. Our goal in this work is to introduce a new deep-learning-based methodology able to efficiently perform data assimilation in situations with a high level of prior uncertainty, and measurements in the form of in-situ (downhole) monitoring data and 4D (time-lapse) seismic data. Data of these types are expected to be available in large-scale carbon storage operations. Our focus here is on early-time predictions, when uncertainty is high but performance assessment is essential, though our methodology is applicable in any phase of a storage project.

Both carbon storage modeling and data assimilation for subsurface operations have been considered extensively in the literature. Reviews of carbon storage, including discussion of challenges, current operations, and modeling strategies, have been provided by \citet{aminu2017review} and \citet{eigbe2023general}. Our review here will focus on history matching. History matching algorithms can be divided into gradient-based and derivative-free methods \citep{rwechungura2011advanced}. Gradient-based methods, like steepest descent method \citep{chavent1975history}, Gauss-Newton method \citep{li2003history}, conjugate gradient method \citep{lee1986history}, etc., often converge quickly, but they require the computation of gradient information, which can be cumbersome or expensive. In addition, these methods essentially search locally, and may therefore not provide a fully representative set of posterior models. Derivative-free methods comprise a broad category of approaches. These include ensemble-based methods, e.g., the well-known ensemble Kalman filter (EnKF) \citep{evensen2007using,zovi2017identification} and ensemble smoother with multiple data assimilation (ESMDA) \citep{emerick2013ensemble,todaro2021ensemble} algorithms, which are widely applied. These methods are efficient, though they entail some underlying assumptions of Gaussianity that may limit their applicability. Markov chain Monte Carlo (MCMC) methods are another type of derivative-free method that have also been used for history matching \citep{oliver1997markov,han2024surrogate}. These methods are more general as they involve fewer underlying assumptions, though they may require large numbers of function evaluations to achieve convergence.

Different types of surrogate models, applied in place of computationally intensive forward simulation runs, have been used with many different history matching methods.
For example, \citet{chen2018geologic} constructed a multivariate adaptive regression spline (MARS) proxy for use in CO$_2$ storage modeling and data assimilation. \citet{chen2020development} constructed a bagging MARS (BMARS) surrogate for calibration of a CO$_2$ reservoir model. \citet{rana2018efficient} applied a Gaussian-process-based surrogate model to history match a coalbed methane reservoir. \citet{wang2021deep} utilized a differentiable neural network surrogate model to obtain gradient information for use in history matching. 
\citet{aslam2024novel} introduced a reduced-order model (ROM) to accelerate history matching of fractured simulation models. In this approach, the fractures are represented by a CGNet-Frac model. The ROM was history matched to well production data by adjusting model parameters using a gradient-based optimization method.

The development of deep learning models for subsurface flow has been a very active research area, and a wide range of network architectures have been considered. These include convolutional neural networks (CNN) \citep{mo2019deep}, recurrent neural networks (RNN) \citep{zhang2023robust}, graph neural networks (GNN) \citep{tang2024graph}, and Fourier neural operators (FNO) \citep{yan2022robust,tang2024multi}, among many others. Deep learning surrogates have been extensively explored for history matching of subsurface flow problems. \citet{tang2020deep}, for example, constructed a recurrent R-U-Net surrogate model based on a residual U-Net and a long short term memory (LSTM) network for 2D oil-water flow problems, which they used to history match channelized systems. \citet{han2024surrogate} extended this model to treat  CO$_2$ storage problems with multiple geological scenarios. They combined the surrogate model with a hierarchical Markov chain Monte Carlo (MCMC) method to estimate posterior scenario parameters (metaparameters) and realizations. \citet{wang2022surrogate} incorporated domain knowledge into CNNs to construct theory-guided CNN (TgCNN) surrogates for two-phase subsurface flow problems, which they combined with an iterative ensemble smoother for history matching. \citet{zhang2023robust} constructed an LSTM-based surrogate model to predict the oil recovery factor. This was combined with Bayesian MCMC for history matching. 

Time-lapse (4D) seismic data have been utilized in both oil/gas production and geological CO$_2$ storage settings. These data can provide information on time-varying pressure and saturation, and are thus very useful for reservoir monitoring \citep{lumley20104d,padhi2014efficient, tiwari20214d}. The use of 4D seismic for history matching has been considered in many studies; see \citet{oliver20214d} for a review of this area. \citet{dadashpour2008nonlinear}, for example, estimated saturation and pressure changes using a nonlinear Gauss-Newton method, in which amplitude differences were utilized to represent seismic data. \citet{lorentzen2024ensemble} proposed a workflow for ensemble-based history matching using 4D seismic data for a complex oil field. 
\citet{rossi2023data} applied ESMDA to history match a Brazilian deep water heavy-oil reservoir using both production data and 4D seismic data. Forward seismic modeling usually involves petro-elastic modeling (PEM), which is computationally expensive. For this reason, surrogates for forward seismic modeling have also been developed. See, e.g., \citet{macbeth2016fast}, \citet{geng2017seismic} and \citet{danaei2023all}.

In this study, we introduce a new deep learning-based framework for history matching of geological carbon storage with 4D seismic and monitoring well data. Our framework includes both new and existing components, though the integration of these various elements and their effective use for history matching with widely different data types is entirely new. 
In our procedure, rather than construct and train a single complex network to predict both types of data, we introduce two separate fit-for-purpose deep learning models. These models are each relatively straightforward to develop and train, which represents a major advantage of our approach. In particular, the total training time for both networks is about 2.25~hours. This represents a high level of efficiency when contrasted with training times of tens of hours or more for CNN-RNN or GNN procedures \citep{han2024surrogate, tang2024_ccs_graph}. Another new treatment introduced here is the scale change applied in the seismic surrogate model. Specifically, the 3D U-Net developed in this work takes as input the highly resolved geomodel, though the saturation output it provides is at a different scale; namely, the scale informed by the seismic data. Monitoring well data, which are highly resolved vertically but very local areally, are predicted using a 1D U-Net with multiple channels corresponding to time steps.


The two trained surrogates are incorporated within a hierarchical MCMC method \citep{han2024surrogate} for history matching. Higher degrees of prior uncertainty are considered here than in many previous history matching studies. In particular, our geomodels are uncertain both in their metaparameters (e.g., mean and variance of log-permeability, permeability anisotropy ratio) and in terms of detailed heterogeneous realizations. For these highly uncertain systems, we assess the impact of the seismic data on uncertainty reduction and on the predicted CO$_2$ plume using two different treatments. In one case, the seismic data are assumed to provide an estimate of the CO$_2$ saturation value (which may be possible in some practical settings), while in the other case the seismic data indicate only the presence or absence of CO$_2$. The data assimilation capabilities and detailed results presented here would not be feasible without the efficient and robust workflow developed in this study.

This paper proceeds as follows. In Section~\ref{sec:geomodel and seismic}, we describe the metaparameters and geomodels, the geomodel parameterization, and the approach for generating the interpreted seismic data. The deep learning models developed to predict seismic saturation and monitoring well data are presented in Section~\ref{sec:surr}. In Section~\ref{sec:mcmc} we discuss the hierarchical MCMC method and the overall history matching workflow. Results demonstrating the performance of the deep learning surrogates are presented in Section~\ref{sec:case}. History matching results illustrating the overall performance of our workflow and the impact of different data types on uncertainty reduction are provided in Section~\ref{sec:hm}. Conclusions and suggestions for future work appear in Section~\ref{sec:conclusion}.

\section{Model description and interpreted seismic data}
\label{sec:geomodel and seismic}

In this section, we describe the simulation setup and the synthetic geomodels considered in this study. We then discuss how the interpreted 4D seismic data used in our history matching framework are generated from high-fidelity models.

\subsection{Basic modeling setup} 
\label{sec:geomodel}

In practical CO$_2$ storage operations, the early-time dynamics (e.g., from a few months to 1-2~years) will be of great interest. This is because the aquifer response over this time frame can be used to assess whether the operation is proceeding as expected, to better understand formation properties, and to determine future injection and monitoring strategies. The geomodels considered in this study are designed to capture these early-time dynamics. The models are gridded such that the near-injector flow field is well resolved. The storage aquifer domain is relatively small since it is not our intention to use the model for long-time predictions.  

\begin{figure}[htbp] 
\centering 
\vspace{0.35cm} 
\setlength{\lineskip}{\medskipamount}
\subcaptionbox{Full domain including central storage aquifer and large surrounding region
\label{fig:full_domain}}
{\includegraphics[width=0.45\linewidth]{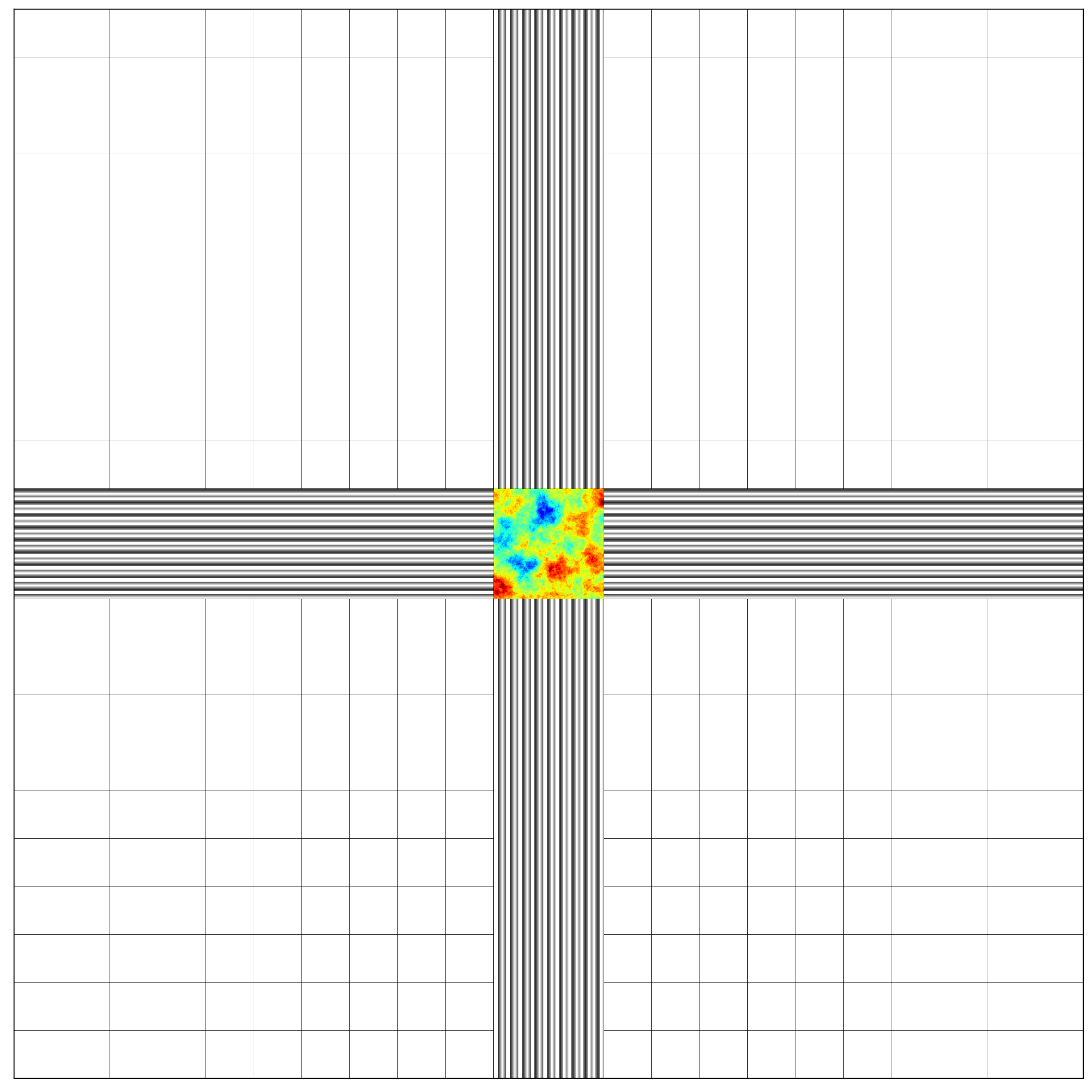}}
\subcaptionbox{Storage aquifer with injector and monitoring (observation) well
\label{fig:well_location}\hfill}
{\includegraphics[width=0.45\linewidth]{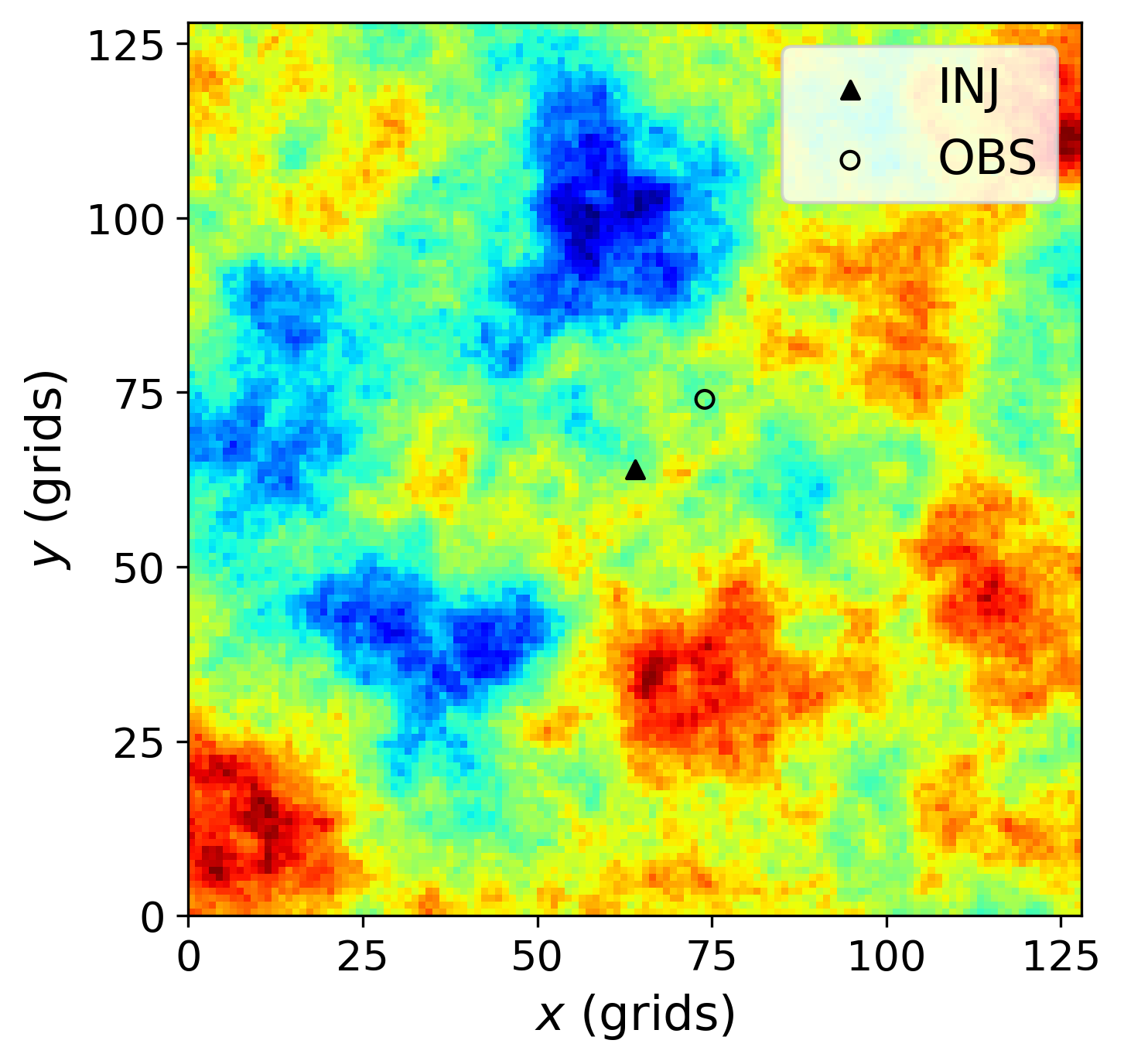}}
\caption{Model and wells used in all simulations.}
\label{fig:geomodel_well}
\end{figure}

A 3D geomodel is considered in this work. The physical domain consists of the storage aquifer and a large surrounding region (for pressure dissipation), as shown in Figure~\ref{fig:geomodel_well}(a). The storage aquifer, which is the target zone for CO$_2$ storage, is in the center of the domain. The size of the central storage aquifer is 896~m $\times$ 896~m $\times$ 70~m, and the full domain is 109~km $\times$ 109~km $\times$ 70~m. No-flow boundary conditions are applied on the outer boundaries of the surrounding region and at the top and bottom of the model. The full domain is discretized into 148 $\times$ 148 $\times$ 35 cells (total of 766,640 cells). High resolution is used in the storage aquifer domain (Figure~\ref{fig:geomodel_well}(a)), which is discretized into 128 $\times$ 128 $\times$ 35 cells (total of 573,440 cells). The cells in the storage aquifer are of dimensions 7~m $\times$ 7~m $\times$ 2~m. Large cells are used in the surrounding region, which is not an issue because very little (if any) CO$_2$ reaches this portion of the model.

Although our immediate interest is in predictions over a one-year time frame, the methodology introduced in this study is also directly applicable for longer-term predictions (years to decades). In such cases, the CO$_2$ plume will extend farther, which means that the high-resolution central storage aquifer domain will need to be larger (areally) than 896~m $\times$ 896~m. This in turn will require the use of more grid cells in this region, leading to higher computational costs for the full-order training simulations and network training.

A fully penetrating vertical injection well is placed near the center of the storage aquifer, at $i=65$, $j=65$ (where $(i,j)$ denotes areal location on the grid), as shown in Figure~\ref{fig:geomodel_well}(b). This well is specified to inject 0.5~Mt/year, for a period of one year. A monitoring well, shown in Figure~\ref{fig:geomodel_well}(b), is placed at $i=75$, $j=75$, which is about 100~m away from the injector. The monitoring well here is assumed to provide CO$_2$ saturation in each layer at a set of time steps. Pressure data, if available, could also be included in our framework.

\subsection{Metaparameters and geomodels} 
\label{sec:meta}

In this work, geomodels characterized by a set of scenario parameters, referred to as metaparameters, are considered. For any set of metaparameters, an infinite number of geomodel realizations can be generated, each with a different property distribution. 
These metaparameters, denoted by $\mathbf{h}$, are represented as
\begin{center}
\begin{equation}
\setlength{\abovedisplayskip}{-9pt}
\mathbf{h}=\left[\mu_{\log k}, \sigma_{\log k}, \log_{10}a_r, d, e \right],
\label{eq:metapara}
\end{equation}
\end{center}
where $\mu_{\log k}$ and $\sigma_{\log k}$ denote the mean and standard deviation of the log-permeability field, $a_r$ is the anisotropy ratio between the vertical and horizontal permeability, i.e., $a_r=k_v/k_h$, where $k_v=k_z$ is the vertical permeability and $k_h=k_x=k_y$ is the horizontal permeability, and $d$ and $e$ are parameters that relate porosity and log-permeability. We denote the storage aquifer porosity and permeability fields as $\boldsymbol \phi_s\in\mathbb{R}^{n_s\times 1}$ and $\mathbf{k}_s\in\mathbb{R}^{n_s\times 1}$, where $n_s$ is the total number of grid blocks in the storage aquifer. The relationship between porosity and permeability is given by
\begin{center}
\begin{equation}
\setlength{\abovedisplayskip}{-9pt}
(\phi_s)_i=d \cdot (\log k_s)_i + e, \quad i=1, \ldots, n_s,
\label{eq:phi_logk}
\end{equation}
\end{center}
where $(\phi_s)_i$ and $(\log k_s)_i$ are porosity and log-permeability for grid block $i$. 

The metaparameters associated with the storage aquifer are taken to be uncertain. These parameters and their (prior) ranges are shown in Table~\ref{table:meta_range}. The MCMC procedure applied in this work will act to reduce these uncertainty ranges.

\begin{table}[htbp]
\setlength{\belowcaptionskip}{0.2cm}
\centering
\caption{Metaparameters and ranges}
\makebox[\textwidth][c]{
\begin{tabular}{ccc}
\hline
\bf{Metaparameter}&\bf{Range}\\
\hline
Mean of log-permeability, $\mu_{\log k}$ &	$U$[2, 6] \\
Standard deviation of log-permeability, $\sigma_{\log k}$&	$U$[1.0, 2.5]\\
Log of permeability anisotropy ratio, $\log_{10}a_r$ &	$U$[-2, 0]\\
Parameter $d$  (in Eq.~\ref{eq:phi_logk}) & $U$[0.02, 0.05]\\
Parameter $e$  (in Eq.~\ref{eq:phi_logk}) & $U$[0.05, 0.12]\\
\hline
\end{tabular}
}
\label{table:meta_range}
\end{table}

Our procedure for constructing geological realizations of the storage aquifer follows the approach of \citet{han2024surrogate}. The first step involves the application of principal component analysis (PCA) to a set of prior realizations of standard Gaussian fields of prescribed correlation structure. These realizations are generated using the Stanford Geostatistical Modeling Software package, SGeMS \citep{remy2009applied}. A spherical variogram, with correlation lengths in $x$, $y$, and $z$ of $l_x=l_y=280$~m and $l_z=7$~m, is specified for all realizations. The use of PCA allows us to avoid running SGeMS during online computations. This accelerates the overall workflow since the overhead associated with SGeMS can be significant.

The PCA representation is constructed through application of singular value decomposition (SVD) to the matrix containing, as its columns, the centered realizations generated using SGeMS. With this representation, new (random) realizations that honor the prescribed correlation structure can be  generated using
\begin{center}
\begin{equation}
\setlength{\abovedisplayskip}{-9pt}
\mathbf{y}_{pca}= \Phi \boldsymbol \xi+\overline{\mathbf{y}}.
\label{eq:pca_expression}
\end{equation}
\end{center}
Here $\mathbf{y}_{pca}\in\mathbb{R}^{n_s\times 1}$ is a new (PCA-based) realization, $\Phi\in \mathbb{R}^{n_s\times n_d}$ denotes the basis matrix obtained from SVD, truncated based on the $n_d$ largest singular values and corresponding left singular vectors, $\boldsymbol \xi\in \mathbb{R}^{n_d\times 1}$ is a stochastic vector with each component sampled from $\mathcal{N}(0,1)$, and $\overline{\mathbf{y}}\in\mathbb{R}^{n_s\times 1}$ is the mean of the SGeMS realizations. With this representation, the high-dimensional Gaussian field is parameterized with a low-dimensional latent vector $\boldsymbol \xi$. 

\begin{figure}[H] 
\centering 
\vspace{0.35cm} 
\setlength{\lineskip}{\medskipamount}
\subcaptionbox{Original SGeMS field}{
\includegraphics[height=0.25\linewidth]{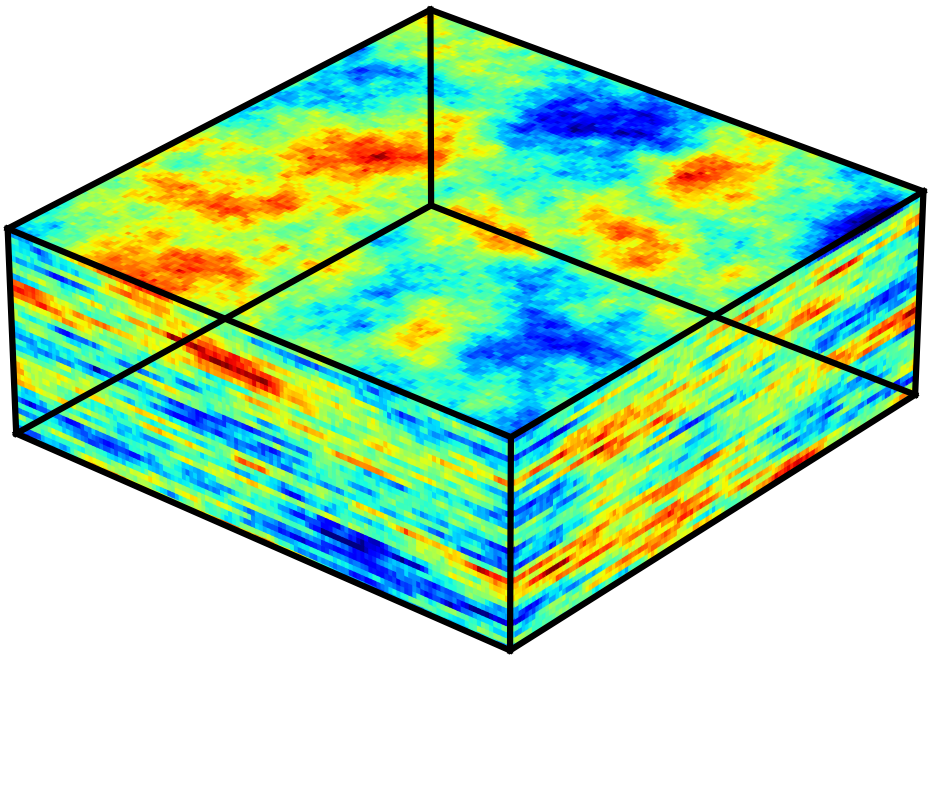}}
\hspace{0.35cm}
\subcaptionbox{PCA reconstruction with $n_d=800$}{
\includegraphics[height=0.25\linewidth]{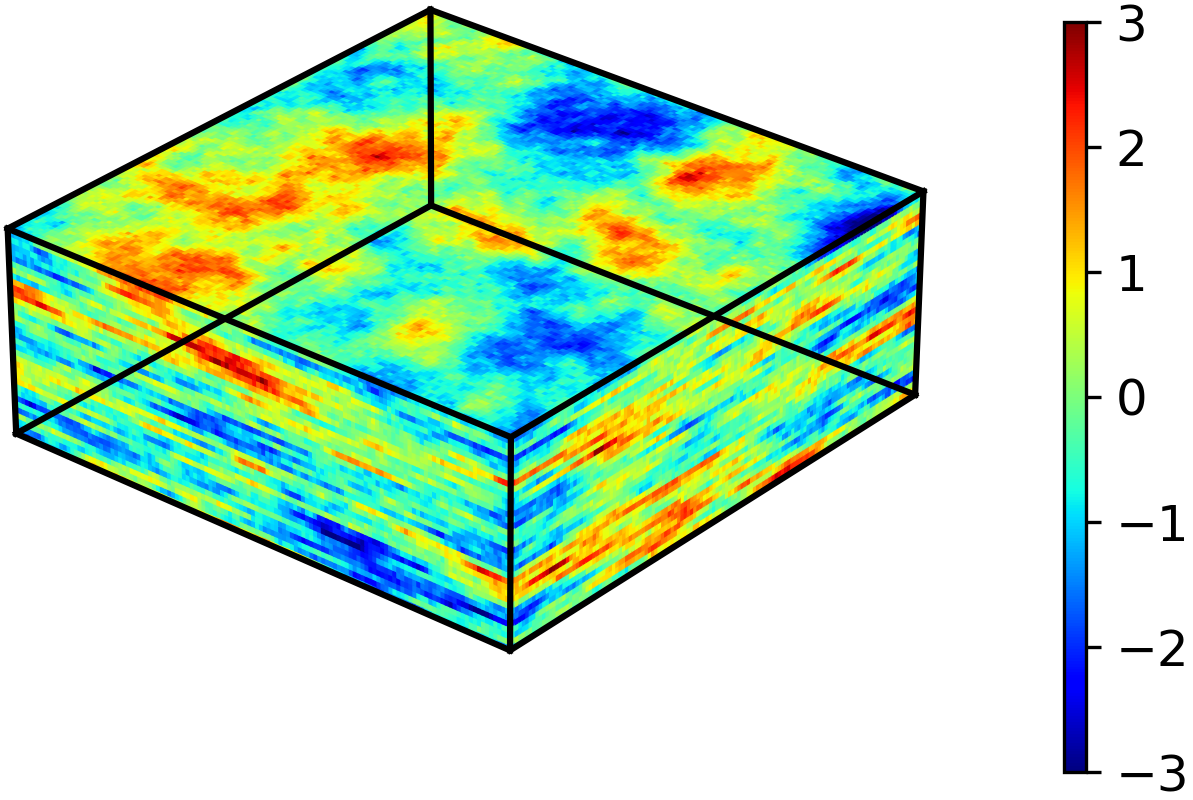}}

\caption{Original SGeMS and reconstructed PCA fields.}
\label{fig:logk_pca}
\end{figure}

In this work, the dimension of $\boldsymbol \xi$ ($n_d$) is set to 800, which preserves 89\% of the `energy' (sum of squared singular values) of the original realizations. The use of $n_d=800$ is sufficient to retain the key features and the important fine-scale detail in the SGeMS models. This is illustrated in Figure~\ref{fig:logk_pca}, where we display an SGeMS realization and the corresponding PCA reconstruction. The use of a lower value of $n_d$ could result in an overly smoothed PCA model, but this is not observed with $n_d=800$.

Given a set of metaparameters $\mathbf{h}$ and the PCA representation $\mathbf{y}_{pca}(\boldsymbol \xi)$, the full storage aquifer geomodel, which we denote as $\mathbf{m}=\mathbf{m}(\mathbf{h}, \boldsymbol \xi)$, is constructed as follows. Permeability (in $x$ and $y$) in each grid block is given by
\begin{center}
\begin{equation}
\setlength{\abovedisplayskip}{-9pt}
(\log k_s)_i=\sigma_{\log k} \cdot (y_{pca})_i+\mu_{\log k}, \quad i=1, \ldots, n_s,
\label{eq:mean_std}
\end{equation}
\end{center}
where $(y_{pca})_i$ is the component of $\mathbf{y}_{pca}$ corresponding to block $i$. The vertical permeability is given by $(k_z)_i=a_r \cdot (k_s)_i$, and porosity is computed from Eq.~\ref{eq:phi_logk}. We reiterate that, in our history matching, both $\mathbf{h}$ and  $\boldsymbol \xi$ are considered to be uncertain.

Cutoffs are applied to the PCA-generated permeability and porosity to avoid extreme (nonphysical) values. For permeability, the maximum and minimum values are $10^4$~mD and $10^{-4}$~mD, and for porosity they are 0.35 and 0.05.

\subsection{Generation of interpreted 4D seismic data} 
\label{sec:synth_seismic}

In practice, seismic interpretations are obtained by solving geophysical inverse problems. This involves the use of measured seismic data of various types, and the seismic inversion provides parameters such as seismic velocities, acoustic impedance, density, etc.~\citep{oliver20214d}. In the case of 4D seismic monitoring, data are collected at a sequence of times, and the data are inverted to obtain estimates of pressure and saturation throughout the domain.

In our framework, rather than use (and invert) `raw' 4D seismic data, we assume that we are provided with estimates of gas (supercritical CO$_2$) saturation or plume location (by plume location we mean the presence of gas above some threshold saturation). These interpreted estimates are considered to be at the level of resolution achieved by 4D seismic data. This is, we believe, a reasonable approach, since the time-lapse seismic data are often treated and interpreted in a workflow that is distinct from the flow-based data assimilation process. An analogous approach, in the context of history matching for oil reservoir simulation, was applied by \citet{bukshtynov2015comprehensive}.

We now describe how interpreted 4D seismic data are generated from flow simulation results for a highly resolved geomodel. The simulation provides CO$_2$ saturation fields (for the storage aquifer) on a grid of dimensions 128 $\times$ 128 $\times$ 35, with cells of size 7~m $\times$ 7~m $\times$ 2~m. The resolution of interpreted 4D seismic data depends on the seismic wavelength and other parameters, but a typical resolution level is about 20~m horizontally and 10~m vertically~\citep{souza20194d}. A resolution of 21~m $\times$ 21~m $\times$ 10~m corresponds to a 3 $\times$ 3 $\times$ 5 filtering of our fine-scale saturation fields. We now describe the details of this filtering.

\begin{figure}[htbp]
    \centering
    \includegraphics[width =0.9\textwidth]{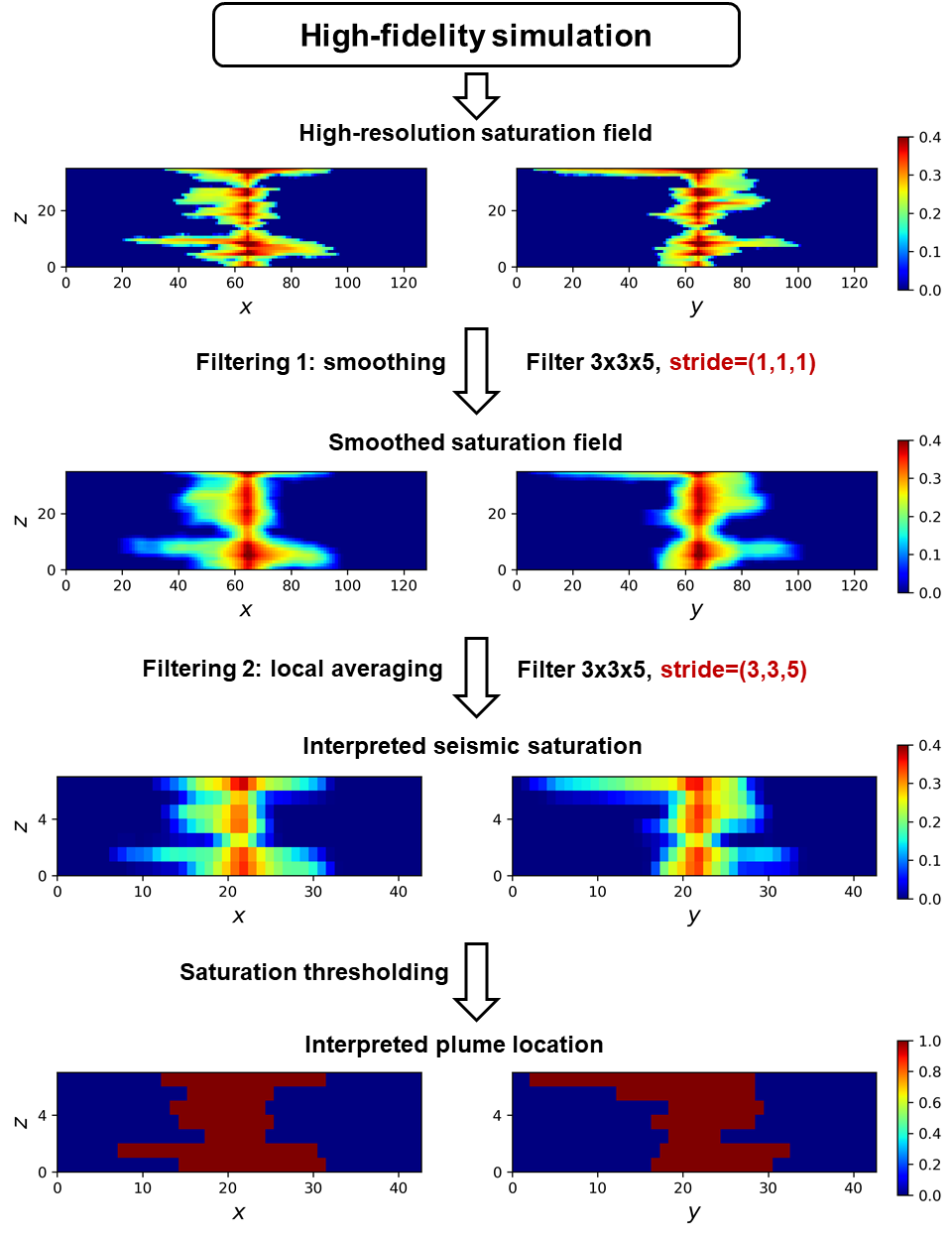}
    \caption{Schematic showing the workflow for generating interpreted seismic data from high-fidelity simulation results. Images on the left are $x$-$z$ cross sections from the full 3D model and those on the right are $y$-$z$ cross sections. Interpreted seismic data can provide estimates of either seismic-scale $S_g$ or plume location.} 
    \label{fig: generation}
\end{figure}

A two-step filtering, illustrated in Figure~\ref{fig: generation}, is applied to generate the interpreted seismic data. The first step acts to smooth the original simulation results, and the second step resamples the smoothed results onto a coarser grid. This two-step approach reduces the `blockiness' that could occur if a direct (one-step) averaging was used. In the first step, the filter is applied with a stride of 1, so the dimensions of the filtered results are 126 $\times$ 126 $\times$ 31. In the second step, we use a stride of 3 in the horizontal directions and 5 in the vertical direction, yielding a result of dimensions 42 $\times$ 42 $\times$ 7. Interpreted seismic data at these dimensions (42 $\times$ 42 $\times$ 7) will be used in our workflow.

In practice, seismic data have been widely used for monitoring CO$_2$ plume migration ~\citep{chadwick20054d,zhang2012cross}. Ideally, the seismic data can be inverted to provide an estimate of CO$_2$ saturation ($S_g$), using full-waveform inversion (FWI)~\citep{dupuy2017quantitative, queisser2013full} or even deep learning models~\citep{um2023real}. However, in some cases, the seismic data may be informative only in terms of the presence or absence of the plume. We refer to this as plume location identification. Plume location is represented in our modeling with a value of 1 where the plume is present (meaning $S_g$ exceeds a detection limit or threshold value), and a value of 0 otherwise. A binary image, as shown in Figure~\ref{fig: generation}, then indicates the presence or absence of the plume. In this work the threshold value is set to 0.05, which is a reasonable practical limit. The use of both types of interpreted seismic data, i.e., seismic-scale $S_g$ fields and binary plume location, will be considered separately in our history matching.

The treatments described above are straightforward and provide us with a means to generate interpreted seismic saturation data or plume location  data, at the appropriate scale, from a high-fidelity geomodel. It is important to note that our approaches for generating interpreted seismic data can be readily modified to include different smoothing, filtering or averaging procedures, etc. It is also straightforward to consider different threshold values in the plume location approach. In either case, the treatment should be consistent with the actual seismic data acquisition and processing workflow. Note also that noise, intended to represent data and model errors, is incorporated into the interpreted seismic data during history matching. Our specific treatment for this will be described in Section~\ref{sec:HM_setup}.

\section{Deep learning surrogate models}
\label{sec:surr}

The MCMC-based history matching procedure applied in this study requires a large number (e.g., $O(10^4 - 10^5)$) of flow simulation results for evaluation of the likelihood function. Because MCMC is essentially a sequential algorithm, the elapsed time required for history matching would be prohibitive if high-fidelity simulations were performed for all function evaluations. Deep learning surrogate models have been shown to be very effective for this type of application. As discussed in the Introduction, many such models have been developed for subsurface flow problems. 

In this study, we require predictions for the global saturation field at the time-lapse seismic-resolution scale, and for monitoring-well data at high vertical resolution but at one (or potentially a few) spatial locations. These data types are very different, and it would be inefficient to construct and train a single surrogate model to predict both sets of data. Therefore, we develop separate, fit-for-purpose surrogate models for each data type. With this approach, we can use simpler networks that do not require time-consuming training or need an excessive amount of training data. We now describe these models in turn.

\subsection{Surrogate model for seismic data}
\label{sec:surr-seis}

The seismic surrogate model accepts the detailed geomodel as input. It then provides interpreted saturation, at the seismic resolution scale and at a specified number of time steps, as output. This prediction process, denoted $f_{\text{seis}}$, can be expressed as follows:
\begin{center}
\begin{equation}
\setlength{\abovedisplayskip}{-9pt}
\hat{\textbf{S}}_{\text{seis}}=f_{\text{seis}}(\mathbf{k},\boldsymbol{\phi},a_r; \theta_{seis}).
\label{3d_unet_forward}
\end{equation}
\end{center}
Here $\hat{\textbf{S}}_{\text{seis}}\in\mathbb{R}^{n_x^{seis}\times n_y^{seis}\times n_z^{seis}\times n_t^s}$ denotes the interpreted seismic prediction, with $n_x^{seis}$, $n_y^{seis}$, and $n_z^{seis}$ the dimension of the interpreted saturation field and $n_t^s$ the number of (4D seismic) time steps, and $\mathbf{k}\in\mathbb{R}^{n_x\times n_y\times n_z}$ 
and $\boldsymbol{\phi}\in\mathbb{R}^{n_x\times n_y\times n_z}$ denote the high-resolution permeability and porosity fields ($n_x$, $n_y$, and $n_z$ are the number of grid blocks in the storage aquifer in the $x$, $y$, and $z$ directions). In this work $a_r\in\mathbb{R}$ is taken to be constant over the spatial domain for a particular geomodel, so a 3D field with constant elements is used for this input. The quantity $\theta_{seis}$ indicates the trainable network parameters.

\begin{figure}[htb]
    \centering
    \includegraphics[width =\textwidth]{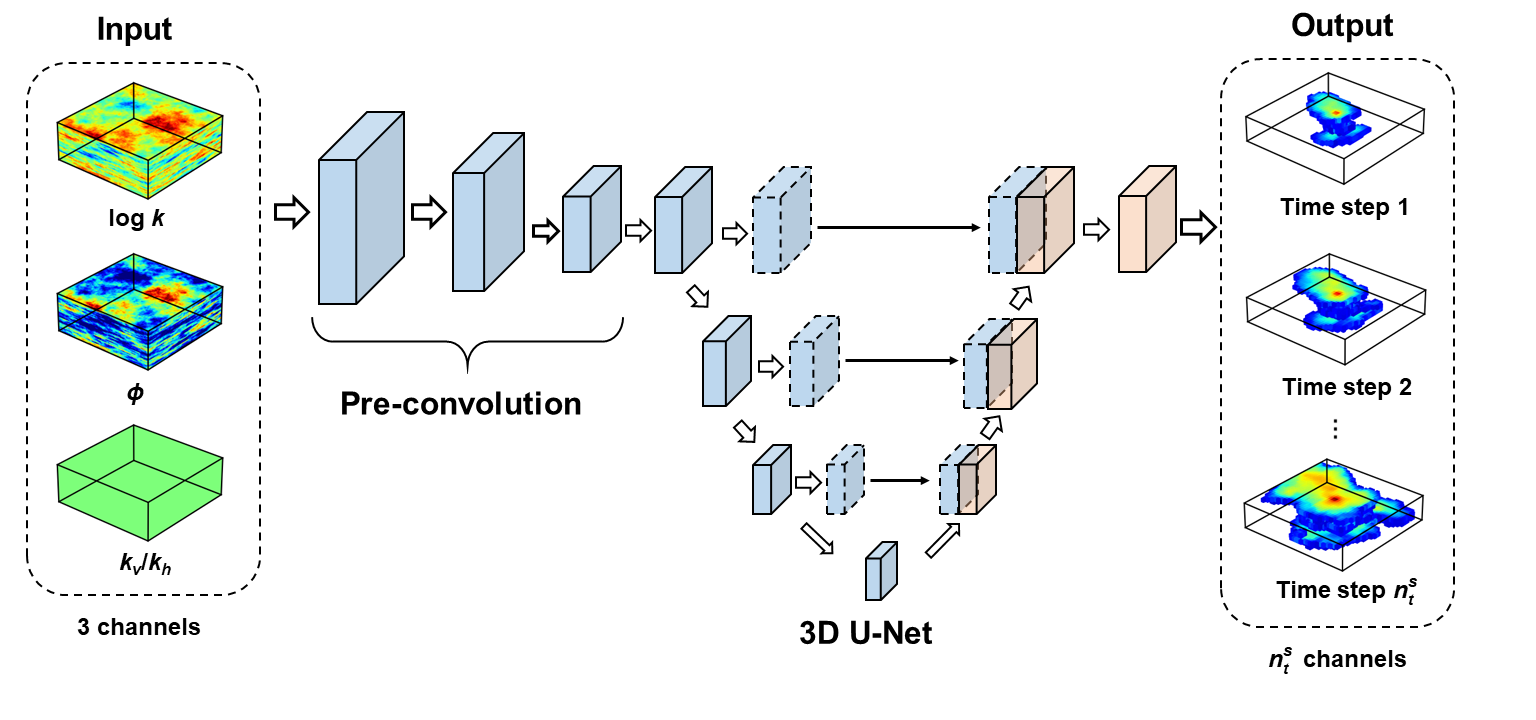}
    \caption{Schematic of U-Net surrogate model used to predict interpreted seismic saturation data.} 
    \label{fig:3d_unet}
\end{figure}

The U-Net architecture, shown in Figure~\ref{fig:3d_unet}, is used for the interpreted-seismic surrogate model. This network has three input channels, which define the high-resolution geomodel. There are multiple output channels, with each channel representing interpreted saturation at a single time step. We reiterate that the input and output properties are at different scales. The detailed architecture of the U-Net surrogate model is provided in Table~\ref{table:seis_surr}. Specifics on the downsampling and upsampling layers in the encoder and decoder portions of the U-Net are given in Tables~\ref{table:downsam} and \ref{table:upsam}.

This U-Net architecture shares similarities with the networks used by \citet{ronneberger2015u} and \citet{cciccek20163d}. Both of these are end-to-end learning models that were utilized for image segmentation. The task here -- approximating the relationship between geomodel properties and interpreted seismic saturation fields -- is also end-to-end, which makes this architecture appropriate for the seismic surrogate model. The skip-connections between the encoder and decoder are typical U-Net characteristics. These enable the network to effectively capture spatial features extracted at both high and low levels during the downsampling process.

\begin{table}[htbp]
\setlength{\belowcaptionskip}{0.2cm}
\centering
\caption{Detailed architecture of the U-Net surrogate model for (interpreted seismic) saturation prediction}
\makebox[\textwidth][c]{
\begin{tabular}{cp{7.5cm}<{\centering}c}\hline
\bf{Module}&\bf{Layers}&\bf{Output size}\\
\hline
\multirow{5}{*}{Preconvolution}&Input (3$\times$128$\times$128$\times$35)\\
&Conv, 64 filters of size 3$\times$3$\times$5, stride (1, 1, 1), padding (0, 0, 1) &	64$\times$126$\times$126$\times$33\\
&Activation (Swish) 	&64$\times$126$\times$126$\times$33\\
&Conv, 32 filters of size 3$\times$3$\times$5, stride (3, 3, 5), padding (0, 0, 1) 	&32$\times$42$\times$42$\times$7\\
&Activation (Swish) &		32$\times$42$\times$42$\times$7\\
\hline
\multirow{4}{*}{Encoder}&Downsample, 32 filters of size 3$\times$3$\times$3, stride (1, 1, 1), padding (0, 0, 1) &	32$\times$40$\times$40$\times$7\\
&Downsample, 64 filters of size 4$\times$4$\times$3, stride (2, 2, 1), padding (0, 0, 1) &	64$\times$19$\times$19$\times$7\\
&Downsample, 128 filters of size 3$\times$3$\times$3, stride (2, 2, 1), padding (0, 0, 0) &	128$\times$9$\times$9$\times$5\\
&Downsample, 256 filters of size 3$\times$3$\times$3, stride (2, 2, 1), padding (0, 0, 0) &	256$\times$4$\times$4$\times$3\\
\hline
\multirow{4}{*}{Decoder}&Upsample, 256 filters of size 3$\times$3$\times$3, stride (2, 2, 1), padding (0, 0, 0) &	256$\times$9$\times$9$\times$5\\
&Upsample, 128 filters of size 3$\times$3$\times$3, stride (2, 2, 1), padding (0, 0, 0) &	
128$\times$19$\times$19$\times$7\\
&Upsample, 64 filters of size 4$\times$4$\times$3, stride (2, 2, 1), padding (0, 0, 1) &
64$\times$40$\times$40$\times$7\\
&Upsample, 32 filters of size 3$\times$3$\times$3, stride (1, 1, 1), padding (0, 0, 1) &
32$\times$42$\times$42$\times$7\\

\hline
\multirow{3}{*}{Output}&Conv, 32 filters of size 3$\times$3$\times$3, stride (1, 1, 1), padding (1, 1, 1) &	32$\times$42$\times$42$\times$7\\
&Activation (Swish) 	&32$\times$42$\times$42$\times$7\\
&Conv, 6 filters of size 3$\times$3$\times$3, stride (1, 1, 1), padding (1, 1, 1) 	&6$\times$42$\times$42$\times$7\\
\hline
\end{tabular}
}
\label{table:seis_surr}
\end{table}

\begin{table}[htbp]
\setlength{\belowcaptionskip}{0.2cm}
\centering
\caption{Architecture of downsampling layers in the U-Net}
\makebox[\textwidth][c]{
\begin{tabular}{p{10cm}<{\centering}}
\hline
\bf{Layers}\\
\hline
Input \\
Conv, $n_{\text{channel}}$ (defined in Table~\ref{table:seis_surr} or \ref{table:mon_surr}) filters with size 3, stride 1, padding 1\\
Activation (ReLU)	\\
Conv, $n_{\text{channel}}$ (defined in Table~\ref{table:seis_surr} or \ref{table:mon_surr}) filters with size, stride, and padding defined in Table~\ref{table:seis_surr} or \ref{table:mon_surr}\\
Activation (ReLU)\\

\hline
\end{tabular}
}
\label{table:downsam}
\end{table}

\begin{table}[htbp]
\setlength{\belowcaptionskip}{0.2cm}
\centering
\caption{Architecture of upsampling layers in the U-Net}
\makebox[\textwidth][c]{
\begin{tabular}{p{10cm}<{\centering}}
\hline
\bf{Layers}\\
\hline
Input \\
ConvTranspose, $n_{\text{channel}}$ (defined in Table~\ref{table:seis_surr} or \ref{table:mon_surr})$\times 2$ filters with size 3, stride 1, padding 1\\
Activation (ReLU)	\\
ConvTranspose, $n_{\text{channel}}$ (defined in Table~\ref{table:seis_surr} or \ref{table:mon_surr}) filters with size, stride, and padding defined in Table~\ref{table:seis_surr} or \ref{table:mon_surr}\\
Activation (ReLU)\\

\hline
\end{tabular}
}
\label{table:upsam}
\end{table}

To train the surrogate model, the mismatch between the predicted and reference (interpreted seismic) saturation fields is minimized over a large set of geomodel realizations (the training set). These realizations are constructed by sampling the metaparameters from the ranges given in Table~\ref{table:meta_range} and the components of $\boldsymbol{\xi}$ from ${\mathcal{N}}(0,1)$, as described earlier. Flow simulation is performed using GEOS (more details on the setup will be provided in Section~\ref{sec:geos_model}). The filtered saturation fields provide the reference results. The loss function is given by
 \begin{center}
\begin{equation}
\setlength{\abovedisplayskip}{-9pt}
L_{\text{seis}}(\theta_{seis})=\frac{1}{n_x^{seis} n_y^{seis} n_z^{seis} n_t^s}\frac{1}{N}\sum_{i=1}^{N}\Vert \hat{{\mathbf S}}_{\text{seis},i}-{\mathbf S}_{\text{seis},i} \Vert_2^2,
\label{3d_unet_loss}
\end{equation}
\end{center}
where $N$ denotes the number of geomodel realizations used for training,  $\textbf{S}_{\text{seis},i}$ denotes the reference interpreted seismic saturation field for training sample $i$, and $\hat{{\mathbf S}}_{\text{seis},i}$ indicates the surrogate model prediction for sample $i$. The model is trained by adjusting the network parameters $\theta_{seis}$ to minimize $L_{\text{seis}}$. This is accomplished with the Adam optimization algorithm \citep{kingma2014adam} on a Nvidia A100 GPU. The initial learning rate is set to be $10^{-4}$, which is updated by multiplying by a factor of 0.2 when the performance does not improve for 10~epochs. The minimum learning rate is set to be $10^{-7}$. When plume location rather than saturation is needed, the predicted seismic saturation field is thresholded to provide the binary plume location field (i.e., the plume is considered to be present when $S_g>0.05$). Therefore, the same surrogate is used for both seismic saturation and plume location.

\subsection{Surrogate model for monitoring-well data}
\label{sec:surr-mon}

We now describe the surrogate model used to predict the CO$_2$ saturation profile (in $z$) at the monitoring well at a series of time steps. The surrogate model prediction of the monitoring data, $f_{\text{mon}}$, can be expressed as

\begin{center}
\begin{equation}
\setlength{\abovedisplayskip}{-9pt}
\hat{\textbf{S}}_{\text{mon}}=f_{\text{mon}}(\mathbf{k}_{\text{local}},\boldsymbol{\phi}_{\text{local}},a_r; \theta_{mon}).
\label{1d_unet_forward}
\end{equation}
\end{center}
Here $\hat{\textbf{S}}_{\text{mon}} \in\mathbb{R}^{n_z\times n_t^m}$ denotes the predicted saturation at the monitoring well, at all $n_z$ layers at $n_t^m$ time steps, $\mathbf{k}_{\text{local}}\in \mathbb{R}^{n_{x}^{l}\times n_{y}^{l}\times n_z}$ and $\boldsymbol{\phi}_{\text{local}}\in \mathbb{R}^{n_{x}^{l}\times n_{y}^{l}\times n_z}$ denote the permeability and porosity in a set of cells in a local $x$-$y$ region (at all values of $z$) encompassing the injection well and monitoring well, with $n_{x}^{l}$ and $n_{y}^{l}$ the number of local cells in the $x$ and $y$ directions. The trainable parameters for this network are denoted $\theta_{mon}$.

\begin{figure}[htb]
    \centering
    \includegraphics[width =\textwidth]{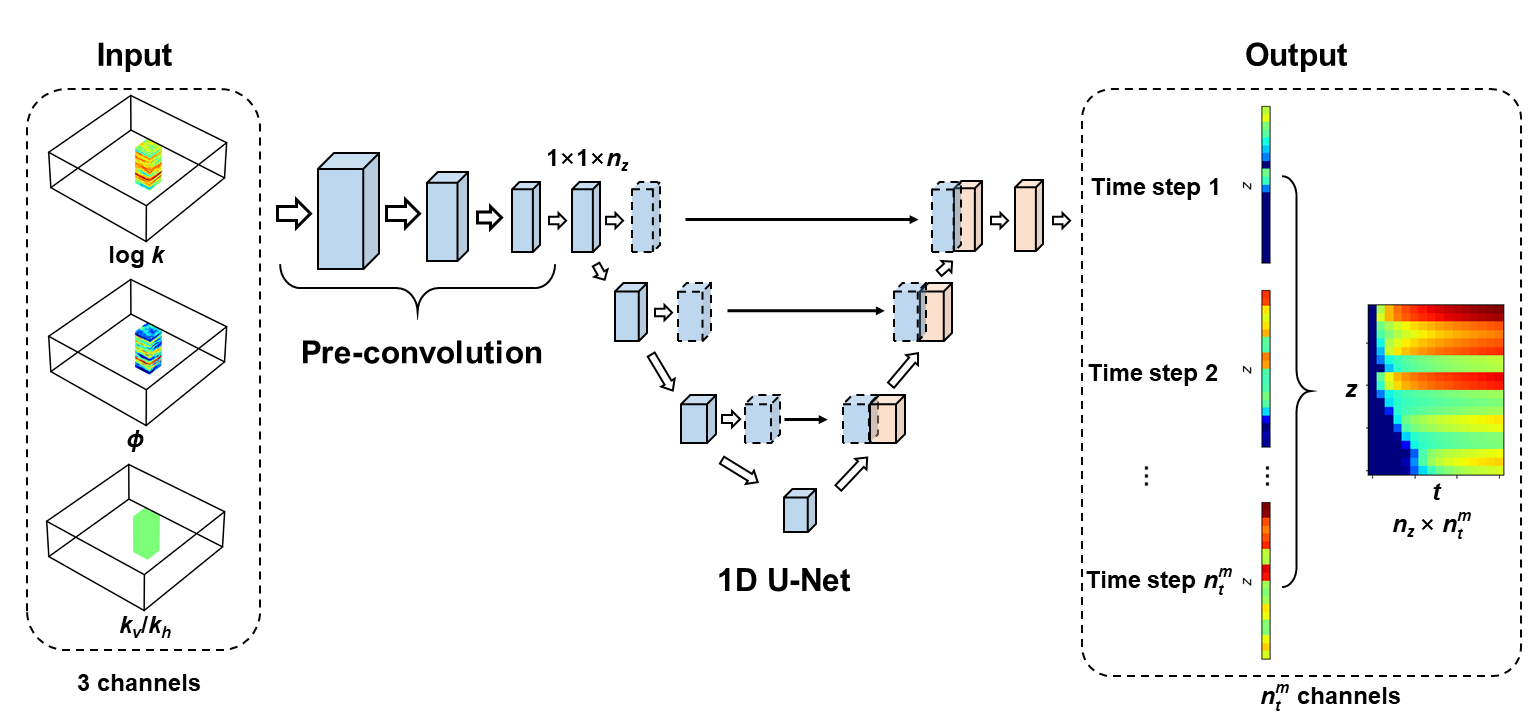}
    \caption{Architecture of 1D U-Net surrogate for monitoring data.} 
    \label{1d_unet}
\end{figure}

We use a 1D U-Net architecture, shown in Figure~\ref{1d_unet}, for the monitoring data surrogate model. The surrogate model outputs a series of 1D vectors -- the elements of these vectors are the predicted saturation values for all $n_z$ cells at the monitoring location at a particular time. There are multiple output channels, with each channel providing the saturation profile at a single time step. The saturation vectors at different time steps can then be combined to form a 2D map of dimensions $n_z\times n_t^m$, as shown in Figure~\ref{1d_unet}. This map displays the evolution of the vertical saturation profile through time.

The monitoring well saturation data are most impacted by storage aquifer properties in the vicinity of the injection and monitoring wells. As noted in Section~\ref{sec:geomodel}, the injection well is located at areal coordinates $(i,j)=(65,65)$, and the monitoring well is located at $(i,j)=(75,75)$. The local geomodel information provided in the network inputs $\mathbf{k}_{\text{local}}$ and $\boldsymbol{\phi}_{\text{local}}$ is over the region $i=63, \ldots, 81$, $j=63, \ldots, 81$, $k=1, \ldots, 35$ (here $k$ is the index in the $z$ direction). Thus the local region includes all layers within a domain that incorporates (with some padding) the injection well and monitoring well. These ranges for the local region correspond to $n_{x}^{l}=n_{y}^{l}=19$.

We apply the 1D U-Net architecture for this surrogate model because the data we seek to represent are themselves a series of 1D vectors. Specifically, the 1D convolutional layers can capture features along the vectors, which correspond to the saturation as a function of depth. Similar network architectures were also used in previous studies predicting 1D signals \citep{hou2021one, chen2023post}. The detailed architecture of the 1D U-Net surrogate model is provided in Table~\ref{table:mon_surr}. Specifications for the residual blocks utilized in the 1D U-Net are given in Table~\ref{table:resblock}.

The loss function to be minimized in the training process for the 1D U-Net surrogate is given by
 \begin{center}
\begin{equation}
\setlength{\abovedisplayskip}{-7pt}
L_{\text{mon}}(\theta_{mon})=\frac{1}{n_z n_t^m}\frac{1}{N}\sum_{i=1}^{N}\left| \hat{\textbf{S}}_{\text{mon},i}-\textbf{S}_{\text{mon},i}\right|,
\label{1d_unet_loss}
\end{equation}
\end{center}
where $\textbf{S}_{\text{mon},i}\in\mathbb{R}^{n_z\times n_t^m}$ denotes the reference result for the saturation profile at the monitoring well location at $n_t^m$ time steps for training sample $i$, and $\hat{\textbf{S}}_{\text{mon},i}$ is the corresponding surrogate model prediction. The RAdam algorithm is adopted to train the network \citep{liu2019variance}. The initial learning rate is 0.001, which decays with a factor of 0.2 when the performance does not improve for 5 epochs. The minimum learning rate is again $10^{-7}$. The training process is implemented on a Nvidia A100 GPU.

The two trained surrogate models can now be used for history matching CO$_2$ storage operations where both 4D seismic and monitoring well data are available. Our framework is flexible, and could be easily extended to handle, e.g., pressure data measured at the monitoring well (though this was not considered here). Either surrogate model could also be modified or enhanced as necessary to treat more complicated geomodels. We now describe the MCMC history matching procedure.

\begin{table}[htbp]
\setlength{\belowcaptionskip}{0.2cm}
\centering
\caption{Detailed architecture of the 1D U-Net surrogate model for prediction of monitoring well data}
\scalebox{0.85}{
\makebox[\textwidth][c]{
\begin{tabular}{cp{9cm}<{\centering}c}\hline
\bf{Module}&\bf{Layers}&\bf{Output size}\\
\hline
\multirow{9}{*}{Preconvolution}&Input 	(3$\times$19$\times$19$\times$35)\\
&Conv, 16 filters of size 3$\times$3$\times$3, stride (1, 1, 1), padding (0, 0, 1)&	16$\times$17$\times$17$\times$35\\
&Activation (Swish)	&16$\times$17$\times$17$\times$35\\
&Conv, 32 filters of size 3$\times$3$\times$3, stride (2, 2, 1), padding (1, 1, 1)	&32$\times$9$\times$9$\times$35\\
&Activation (Swish)&		32$\times$9$\times$9$\times$35\\
&Conv, 64 filters of size 5$\times$5$\times$3, stride (2, 2, 1), padding (0, 0, 1)	&64$\times$3$\times$3$\times$35\\
&Activation (Swish)&		64$\times$3$\times$3$\times$35\\
&Conv, 128 filters of size 3$\times$3$\times$3, stride (1, 1, 1), padding (0, 0, 1)	&128$\times$1$\times$1$\times$35\\
&Activation (Swish)&		128$\times$1$\times$1$\times$35\\
\hline

\multirow{4}{*}{Encoder}&Downsample, 64 filters of size 3, stride 1, padding 0&	64$\times$33\\
&Downsample, 128 filters of size 3, stride 1, padding 0&	128$\times$31\\
&Downsample, 256 filters of size 3, stride 2, padding 0&	256$\times$15\\
&Downsample, 512 filters of size 3, stride 2, padding 0&	512$\times$7\\
\hline

\multirow{4}{*}{Residual}&Residual block, 512 filters of size 3, stride 1, padding 1&	512$\times$7\\
&Residual block, 512 filters of size 3, stride 1, padding 1&	512$\times$7\\
&Residual block, 512 filters of size 3, stride 1, padding 1&	512$\times$7\\
&Residual block, 512 filters of size 3, stride 1, padding 1&	512$\times$7\\
\hline

\multirow{4}{*}{Decoder}&Upsample, 512 filters of size 3, stride 2, padding 0&	512$\times$15\\
&Upsample, 256 filters of size 3, stride 2, padding 0& 256$\times$31\\
&Upsample, 128 filters of size 3, stride 1, padding 0& 128$\times$33\\
&Upsample, 64 filters of size 3, stride 1, padding 0& 64$\times$35\\

\hline
\multirow{3}{*}{Output part}&Conv, 64 filters of size 3, stride 1, padding 1&	64$\times$35\\
&Activation (Swish)	&64$\times$35\\
&Conv, 16 filters of size 3, stride 1, padding 1	&16$\times$35\\
\hline
\end{tabular}
}
}
\label{table:mon_surr}
\end{table}

\begin{table}[htbp]
\setlength{\belowcaptionskip}{0.2cm}
\centering
\caption{Architecture of residual block in the 1D U-Net}
\makebox[\textwidth][c]{
\begin{tabular}{p{10cm}<{\centering}}
\hline
\bf{Layers}\\
\hline
Input \\
Conv, $n_{\text{channel}}$ (defined in Table~\ref{table:mon_surr}) filters with size 3, stride 1, padding 1\\
Activation (ReLU)	\\
Conv, $n_{\text{channel}}$ (defined in Table~\ref{table:mon_surr}) filters with size 3, stride 1, padding 1\\
Addition (with input)\\
Activation (ReLU)\\

\hline
\end{tabular}
}
\label{table:resblock}
\end{table}

\section{Hierarchical MCMC method and overall workflow}
\label{sec:mcmc}

Our data assimilation problem is similar to that considered by \citet{han2024surrogate} in that both metaparameters and associated geomodel realizations must be determined (the previous study did not consider 4D seismic data so the setup is otherwise quite different). Thus, we apply the hierarchical Markov Chain Monte Carlo (MCMC) method used in that work. MCMC is a widely used history matching technique in which geomodels are sampled iteratively and the likelihood, which involves the data mismatch, is evaluated for each sample. After a large number of iterations, the accepted samples can approximate the posterior distributions. In hierarchical MCMC, two levels of sampling are applied -- one for the metaparameters $\mathbf{h}$, and one for the PCA latent variables $\boldsymbol \xi$. Upon convergence (termination), we have an estimate of the posterior probability density function (PDF) $p(\boldsymbol \xi, \mathbf{h} \mid  \mathbf{d}_\text{obs} )$, where $\mathbf{d}_\text{obs}$ denotes the observed data. Because MCMC methods require a large number of sequential function evaluations (e.g., $O(10^4-10^5)$, the use of surrogate models is essential in cases with expensive function evaluations, such as ours.

We now describe the hierarchical procedure. Our description here follows \citet{han2024surrogate}, and that work should be consulted for details. In the first level, the PCA latent variables are sampled. The initial sample is obtained by sampling each PCA variable from ${\mathcal N}(0,1)$. Subsequently, a stochastic perturbation is applied to the previous samples to obtain the new sample. Specifically, the new sample $\boldsymbol \xi'\in\mathbb{R}^{n_d\times 1}$ at iteration $k$ is given by
\begin{center}
\begin{equation}
\setlength{\abovedisplayskip}{-9pt}
\boldsymbol \xi'= (1-\beta^2)\boldsymbol \xi^{k-1}+\beta \boldsymbol{\eta},
\label{eq:pca_sample}
\end{equation}
\end{center}
where $\boldsymbol \xi^{k-1} \in \mathbb{R}^{n_d\times 1}$ denotes the sample from the previous iteration, $\boldsymbol{\eta} \in \mathbb{R}^{n_d\times 1}$ denotes a stochastic vector with each element sampled from ${\mathcal N}(0,1)$, and $\beta\in \mathbb{R}$ is a user-defined hyperparameter that controls the size of the sample update and impacts the MCMC acceptance rate. 

The acceptance or rejection of $\boldsymbol \xi'$ depends on its likelihood. Here we use the Metropolis–Hastings \citep{hastings1970monte} criterion for this determination. With this approach, the acceptance probability for $\boldsymbol \xi'$ is taken as
\begin{center}
\begin{equation}
\setlength{\abovedisplayskip}{-7pt}
\alpha (\boldsymbol \xi^{k-1},\boldsymbol \xi')=\min\left(1, \frac{p( \mathbf{d}_\text{obs} \mid\boldsymbol \xi',\mathbf{h}^{k-1})}{p( \mathbf{d}_\text{obs} \mid\boldsymbol \xi^{k-1},\mathbf{h}^{k-1})}\right),
\label{eq:pca_ap}
\end{equation}
\end{center}
where $p( \mathbf{d}_\text{obs} \mid\boldsymbol \xi',\mathbf{h}^{k-1})$ and $p( \mathbf{d}_\text{obs} \mid\boldsymbol \xi^{k-1},\mathbf{h}^{k-1})$ are the likelihood of the current sample $\boldsymbol \xi'$ and the previous sample $\boldsymbol \xi^{k-1}$, both conditioned to metaparameter sample $\mathbf{h}^{k-1}$ from the previous iteration. The detailed likelihood function will be given in Section~\ref{sec:HM_setup}. A random variable $u_{\text{pca}}$ is then sampled from a uniform distribution $U[0,1]$. If $\alpha (\boldsymbol \xi^{k-1},\boldsymbol \xi')>u_{\text{pca}}$, the new sample $\boldsymbol \xi'$ is accepted, meaning $\boldsymbol \xi^k=\boldsymbol \xi'$. Otherwise, the new sample is rejected, in which case $\boldsymbol \xi^k=\boldsymbol \xi^{k-1}$.

In the second level of hierarchical MCMC the metaparameters are sampled. The initial sample is obtained from the prior distribution. In later iterations, new metaparameters are proposed from a multivariate Gaussian distribution centered on the previously accepted sample. The standard deviations for these distributions are user-defined and problem specific (further details will be provided in Section~\ref{sec:hm}). The acceptance probability for the newly sampled set of metaparameters $\mathbf{h}'$ is analogous to that used for $\boldsymbol \xi'$, i.e.,
\begin{center}
\begin{equation}
\setlength{\abovedisplayskip}{-7pt}
\alpha (\mathbf{h}^{k-1},\mathbf{h}')=\min\left(1, \frac{p(\mathbf{h}')p( \mathbf{d}_\text{obs} \mid\boldsymbol \xi^k,\mathbf{h}')}{p(\mathbf{h}^{k-1})p( \mathbf{d}_\text{obs} \mid\boldsymbol \xi^{k},\mathbf{h}^{k-1})}\right).
\label{eq:meta_ap}
\end{equation}
\end{center}
Here $p(\mathbf{h}')$ and $p(\mathbf{h}^{k-1})$ are prior probabilities of the proposed metaparameters $\mathbf{h}'$ and previously accepted metaparameters $\mathbf{h}^{k-1}$, $p( \mathbf{d}_\text{obs} \mid\boldsymbol \xi^k,\mathbf{h}')$ and $p( \mathbf{d}_\text{obs} \mid\boldsymbol \xi^{k},\mathbf{h}^{k-1})$ denote the likelihood of $\mathbf{h}'$ and $\mathbf{h}^{k-1}$, both conditioned to PCA variables in the current iteration ($\boldsymbol \xi^{k}$). As in the first step, we sample $u_{\text{meta}}$ from $U[0,1]$; if $\alpha (\mathbf{h}^{k-1},\mathbf{h}')>u_{\text{meta}}$, the proposed metaparameters $\mathbf{h}'$ are accepted, otherwise they are rejected. 

The hierarchical MCMC procedure continues until a termination criterion is reached. Here, as in \citet{nicolaidou2022stochastic} and \citet{han2024surrogate}, the procedure is terminated when the relative change in the posterior histogram for the metaparameters falls below a specified threshold. At that point, the accepted samples are taken to provide estimates of their corresponding posterior probability densities.

\begin{figure}[htb]
    \centering
    \includegraphics[width =\textwidth]{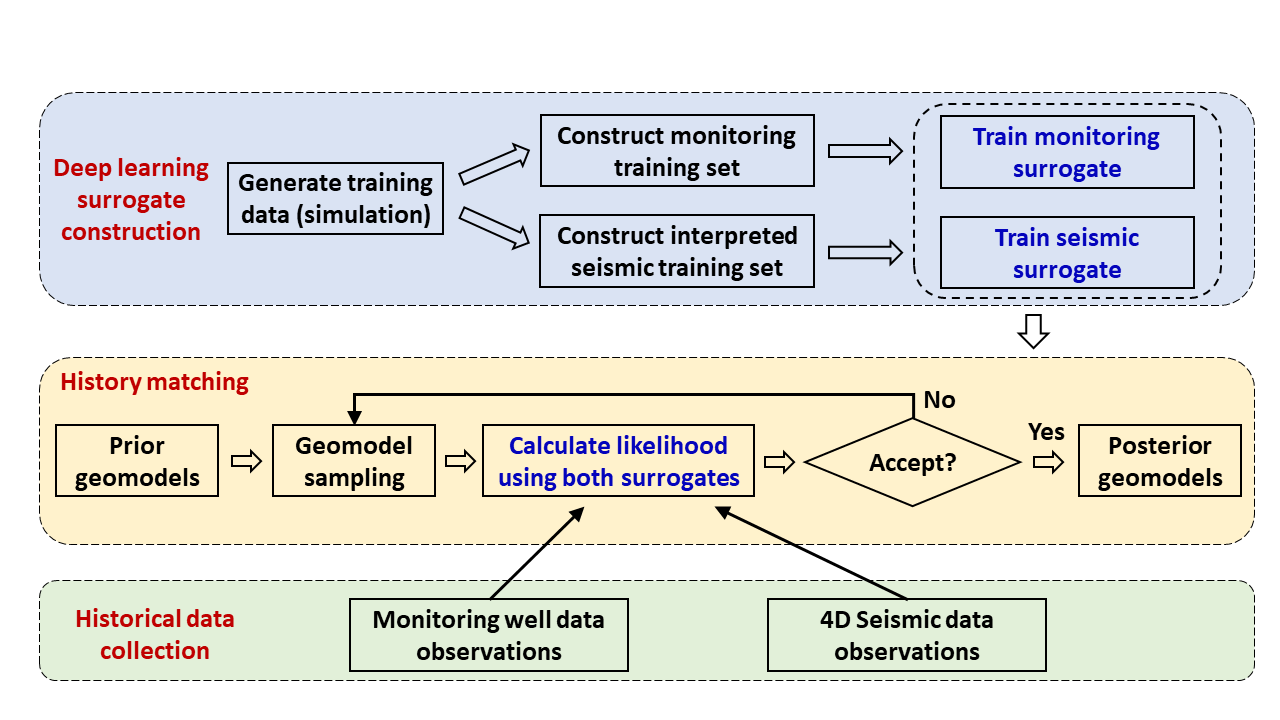}
    \caption{Deep-learning-based workflow for history matching using both monitoring well data and 4D seismic data.}
    \label{fig:hm_framework}
\end{figure}

We now describe the overall workflow involving the use of our deep learning surrogate models in the hierarchical MCMC history matching framework. Figure~\ref{fig:hm_framework} illustrates the two key components of the framework -- surrogate model construction, accomplished in a preprocessing (offline) step, and (online) history matching. For surrogate model construction, the training data are generated by simulating a large number of geomodels. Each geomodel is  generated by randomly sampling the metaparameters and PCA components. The data used for training the monitoring well surrogate are extracted directly from the high-fidelity simulation results, and the data for the seismic surrogate are obtained through use of the filtering procedure. In a real application, the actual monitoring well saturation and interpreted 4D seismic data would be used for history matching. In our examples these data are synthetic, i.e., generated by simulating a specific `true' model and then adding noise. From the observed data (real or synthetic), the mismatch and likelihood are computed, with all function evaluations performed using the surrogate models. Following termination of the hierarchical MCMC procedure, a set of posterior geomodels and associated predictions are attained.

Before presenting numerical results, it is useful to discuss some important differences between hierarchical MCMC and other history matching methods. In practical settings, ensemble-based methods are commonly applied. ESMDA \citep{emerick2013ensemble}, for example, was used by \citet{crain2024integrated} to history match a CO$_2$ storage project in the Illinois Storage Corridor in the US. Ensemble-based methods, however, typically involve several limiting assumptions that are not compatible with our setup. Specifically, these methods are generally used for cases with fixed metaparameters. Even if they are applied hierarchically to consider metaparameter uncertainty, these methods still require Gaussian priors~\citep{oliver2022hybrid}. The setup here is more general in that we consider a range of metaparameters that can have any prior distribution (uniform rather than Gaussian priors are used in our examples). Thus, standard implementations of ensemble-based methods are not applicable, so direct numerical comparisons with such approaches are not straightforward.

Standard (rather than hierarchical) MCMC methods are also difficult to compare against because the dimension of the geomodels is very high, even using a dimension reduction procedure such as PCA. Thus, it may be difficult for the Markov chain to converge in a standard approach. Our hierarchical MCMC procedure is `dimension robust' \citep{chen2018dimension}, meaning the sampling efficiency does not degrade as the dimension of the PCA representation increases. This allows our method to achieve high-dimensional sampling more efficiently than in a standard MCMC approach.

\section{Surrogate model evaluation}
\label{sec:case}

In this section, we describe the simulation setup and evaluate the performance of the surrogate models for a wide range of geomodel realizations. 

\subsection{Simulation setup}
\label{sec:geos_model}

The model domain is illustrated in Figure~\ref{fig:geomodel_well}, and aspects of the problem setup have been discussed in previous sections. We consider heterogeneous, multi-Gaussian log-permeability and porosity fields in the storage aquifer. The permeability and porosity in the surrounding region are set to constant values of 20~mD and 0.2.

\begin{figure}[htbp] 
\centering 
\vspace{0.35cm} 
\setlength{\lineskip}{\medskipamount}
\subcaptionbox{Relative permeability
\label{fig:relperm}}
{\includegraphics[width=0.45\linewidth]{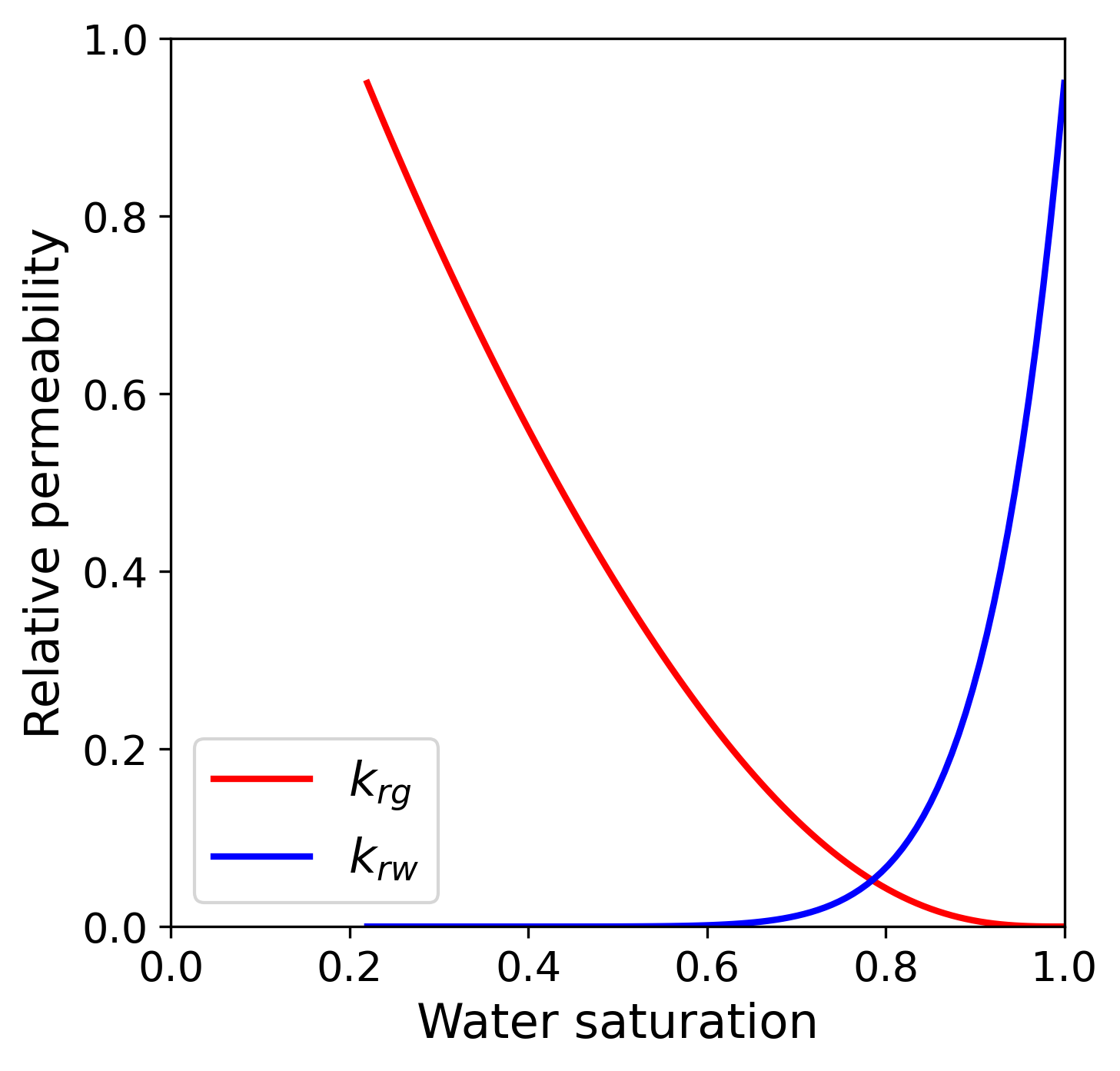}}
\subcaptionbox{Capillary pressure ($\phi=0.2$, $k=20$~md)
\label{fig:cappre}}
{\includegraphics[width=0.45\linewidth]{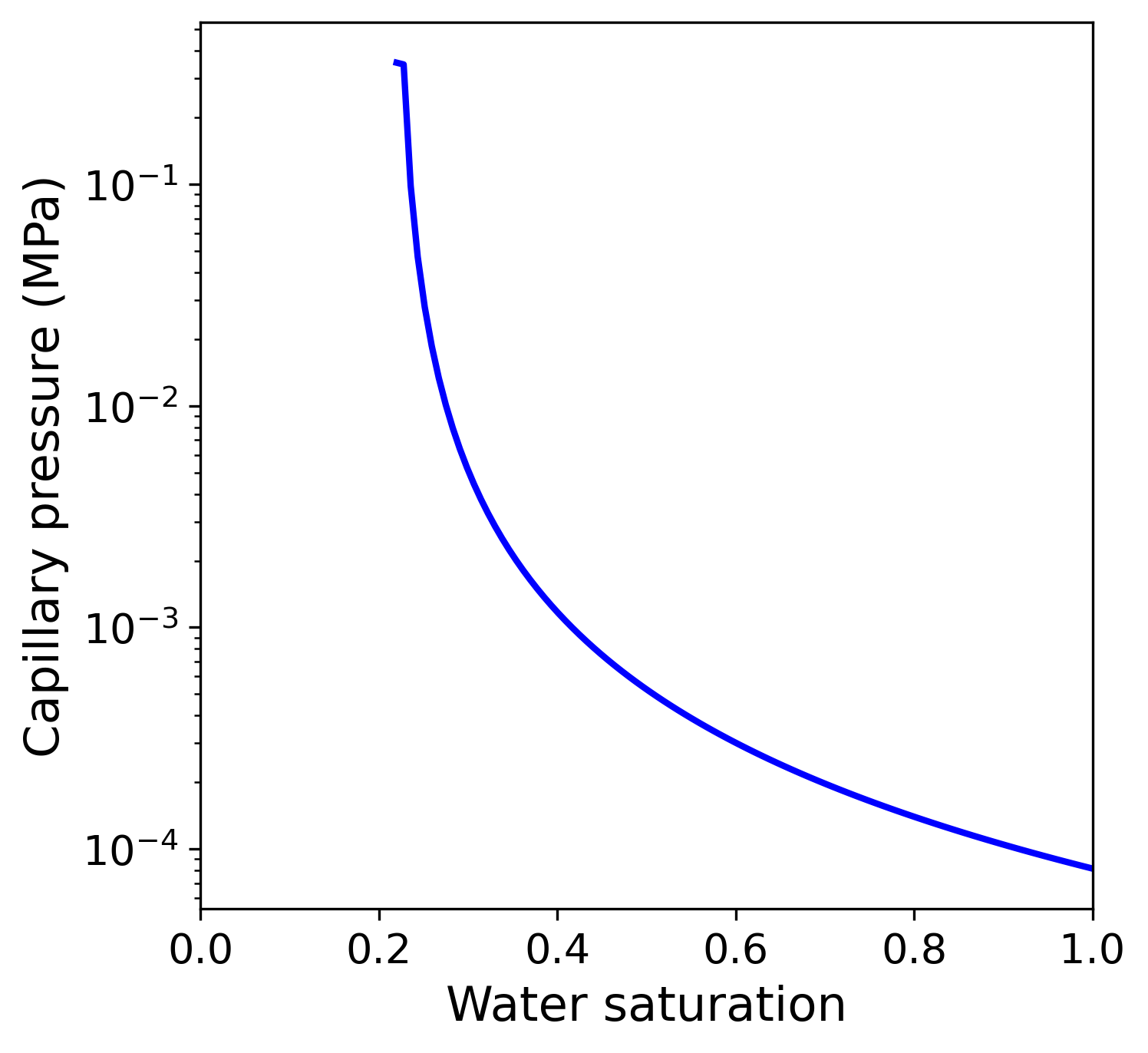}}
\caption{Relative permeability and capillary pressure curves for \texorpdfstring{CO$_2$}{}--brine system.}
\label{fig:relperm_and_cappre}
\end{figure}

The relative permeability and capillary pressure functions are generated from the Brooks-Corey model. The coefficients used here, shown in Table~\ref{table:relperm_cp}, derive from data reported for the Mt.~Simon sandstone by \citet{krevor2012relative}. The capillary pressure curve is calculated using the Leverett J-function. Relative permeability curves for the CO$_2$--water system are presented in Figure~\ref{fig:relperm_and_cappre}(a), and the capillary pressure curve calculated for porosity of 0.2 and permeability of 20~md appear in Figure~\ref{fig:relperm_and_cappre}(b). The initial pressure (at a depth of 1955~m) is 20~MPa. Temperature at 1955~m depth is 50.3~$^\circ$C. The viscosity and density of CO$_2$ and brine/water are computed (as a function of pressure and temperature) in the simulator. All simulations are performed using GEOS, an open-source, multiphysics simulator designed for carbon storage modeling \citep{bui2021multigrid}.

\begin{table}[htbp]
\setlength{\belowcaptionskip}{0.2cm}
\centering
\caption{Brooks-Corey model coefficients for relative permeability and capillary pressure functions}
\makebox[\textwidth][c]{
\begin{tabular}{ccc}
\hline
\bf{Coefficient}&\bf{Value}\\
\hline
Irreducible water saturation, $S_{wi}$ &	0.22 \\
Residual CO$_2$ saturation, $S_{gr}$&	0\\
Water exponent for Corey model, $n_w$ &	9\\
CO$_2$ exponent for Corey model, $n_g$ & 4\\
Relative permeability of CO$_2$ at $S_{wi}$, $k_{rg}(S_{wi})$ & 0.95\\
Capillary pressure exponent, $\lambda$ & 0.55\\
\hline
\end{tabular}
}
\label{table:relperm_cp}
\end{table}

Since our interest is in early time behavior and history matching, the simulation time frame is restricted to 1~year, which is divided into 30~time steps. A total of 4000 geomodel realizations are generated and simulated for surrogate model training and evaluation. Of these models, 3500 comprise the training set and 500 comprise the test set. For each realization, the metaparameters are sampled from the ranges shown in Table~\ref{table:meta_range}, the PCA latent variables are sampled from ${\mathcal N}(0,1)$, and the procedure described in Section~\ref{sec:meta} is applied to construct the full geomodel. Following the flow simulation, the saturation fields are saved at a set of time steps. From these solutions the monitoring data are extracted and the interpreted seismic data are constructed.

\subsection{Surrogate performance evaluation}
\label{sec:surr_evl}
Before assessing the accuracy of the surrogate models, we quantify the runtimes for high-fidelity simulation and surrogate evaluations. The GEOS runs require an average of about 38~minutes. This timing can vary from about 15~minutes to 200~minutes depending on time-step cutting, Newton iteration convergence, etc., which in turn depend on the geomodel. The surrogate models require less than 0.1~second to provide a function evaluation. It is this very high degree of speedup that enables the use of sophisticated history matching methods such as hierarchical MCMC. We now evaluate the performance of the two deep learning surrogate models.   

\subsubsection{4D seismic surrogate evaluation}
\label{sec:seis_surr_evl}

The surrogate for 4D seismic data was described in Section~\ref{sec:surr-seis}. The model is trained to predict interpreted seismic data at six time steps, specifically time steps~5, 10, 15, 20, 25, and 30, which correspond to times of 2, 4, 6, 8, 10 and 12~months. The network architecture has six output channels -- one for each of the six time steps. In the training process, the batch size of the training dataset is set to be 10, and the model is trained for 150 epochs with an initial learning rate of $10^{-4}$. The training process takes approximately 2~hours on a Nvidia A100 GPU. 

\begin{figure}[htbp] 
\centering 
\vspace{0.35cm} 
\setlength{\lineskip}{\medskipamount}
\subcaptionbox{Realization 1 (sim)}
{\includegraphics[height=0.195\linewidth]{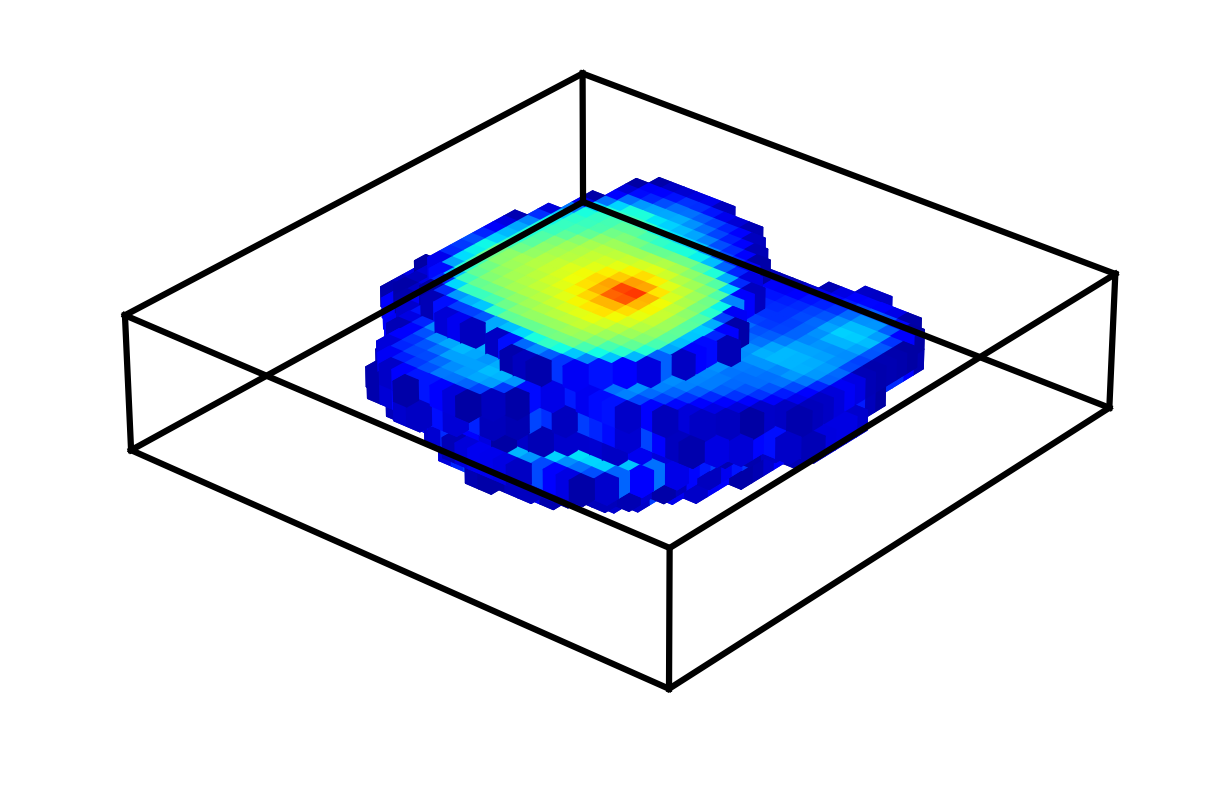}}
\subcaptionbox{Realization 2 (sim)}{
\includegraphics[height=0.195\linewidth]{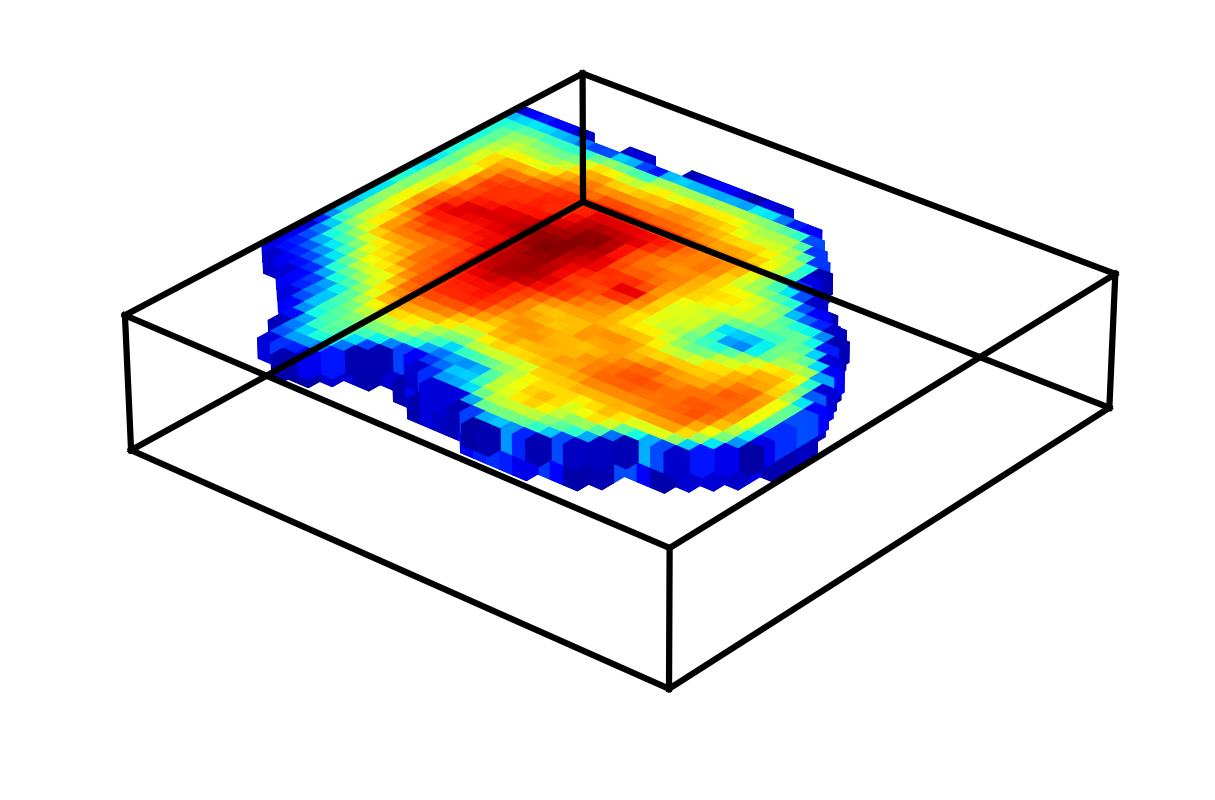}}
\subcaptionbox{Realization 3 (sim)}{
\includegraphics[height=0.195\linewidth]{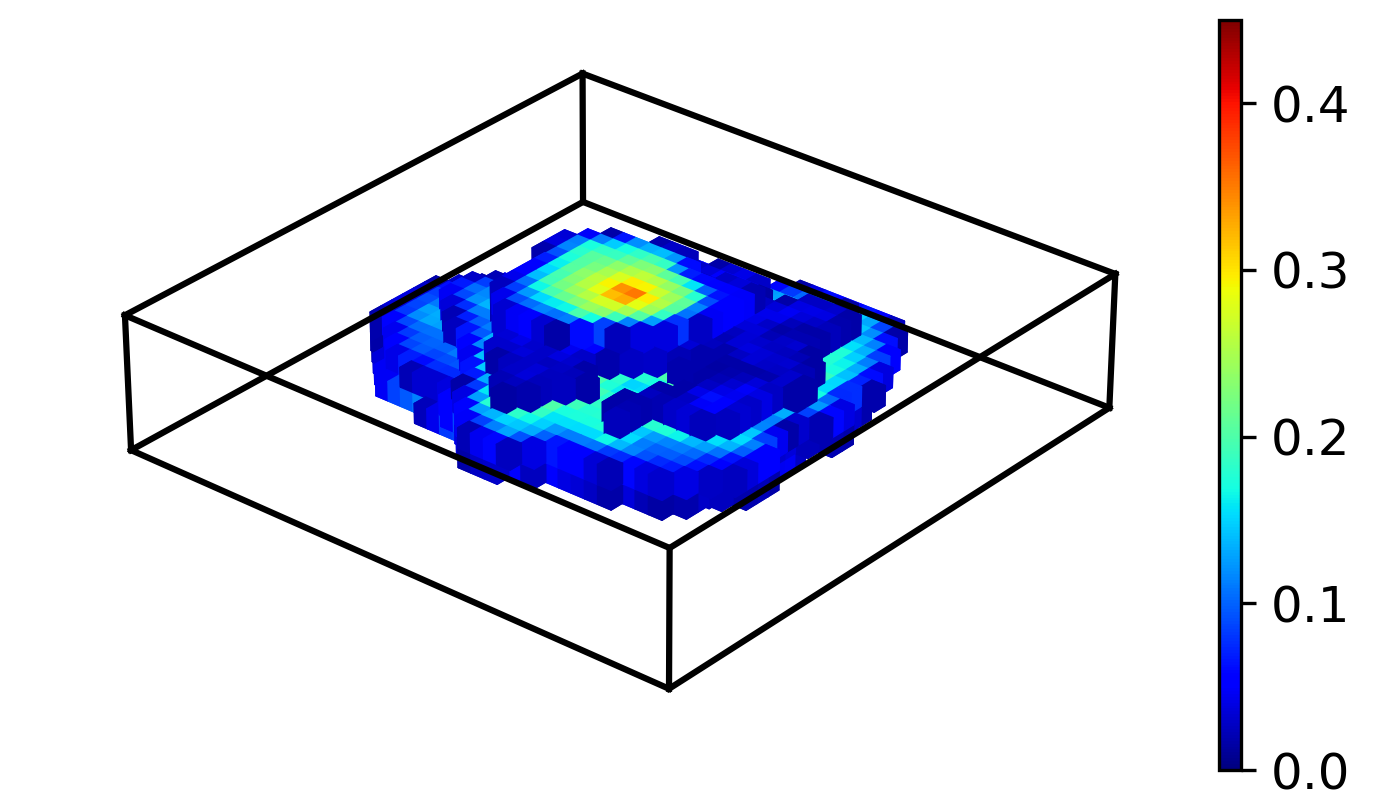}}

\subcaptionbox{Realization 1 (surr)}{
\includegraphics[height=0.195\linewidth]{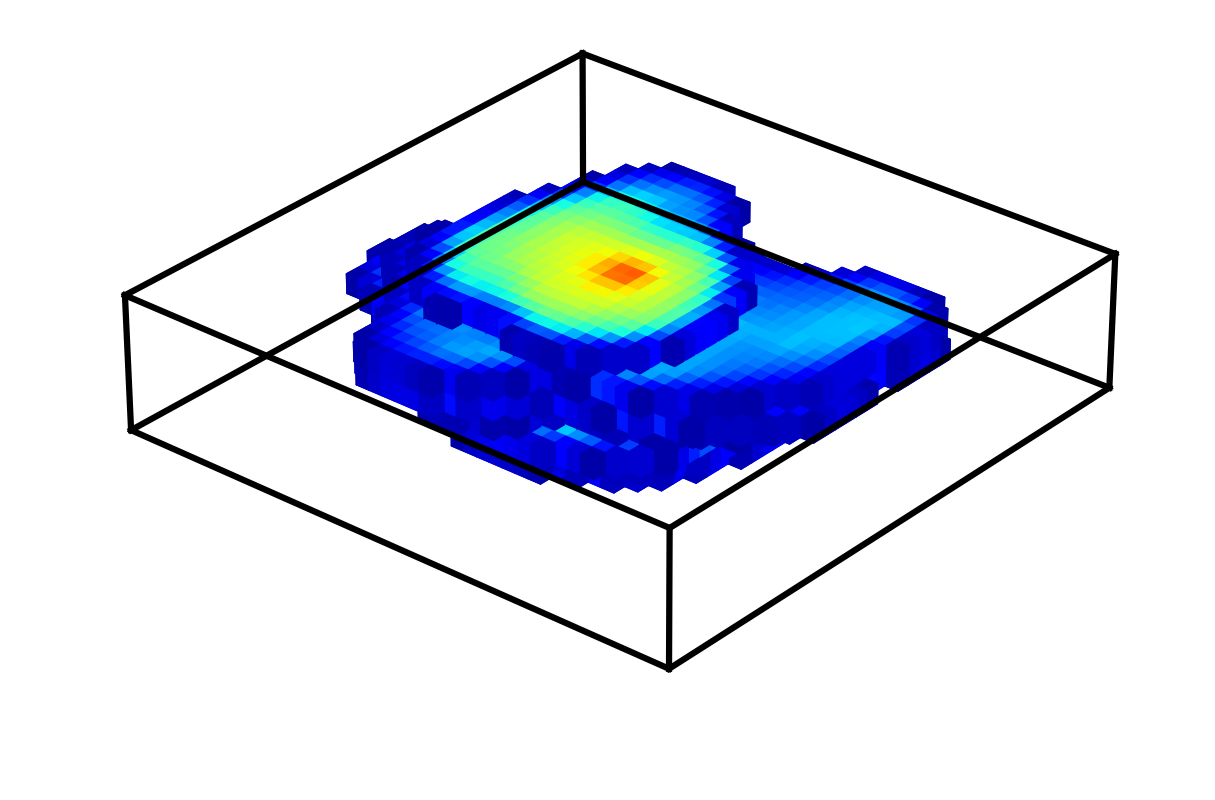}}
\subcaptionbox{Realization 2 (surr)}{
\includegraphics[height=0.195\linewidth]{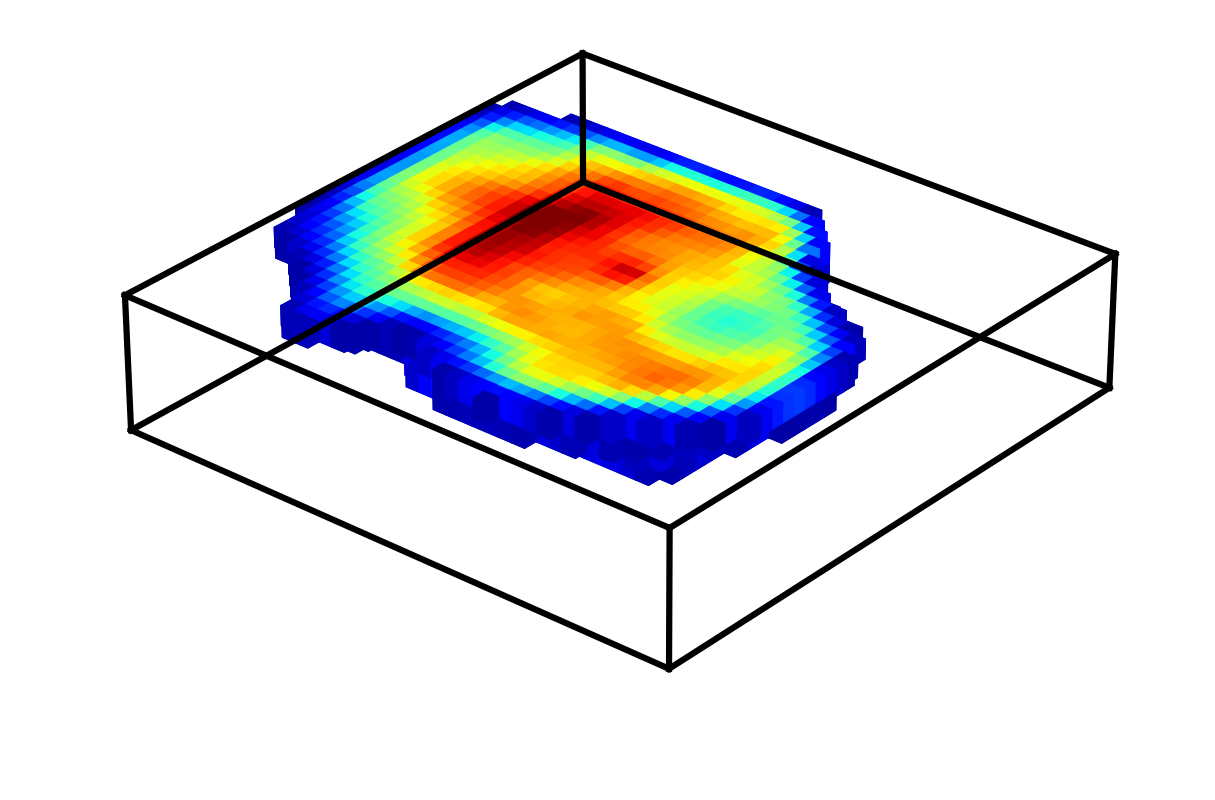}}
\subcaptionbox{Realization 3 (surr)}{
\includegraphics[height=0.195\linewidth]{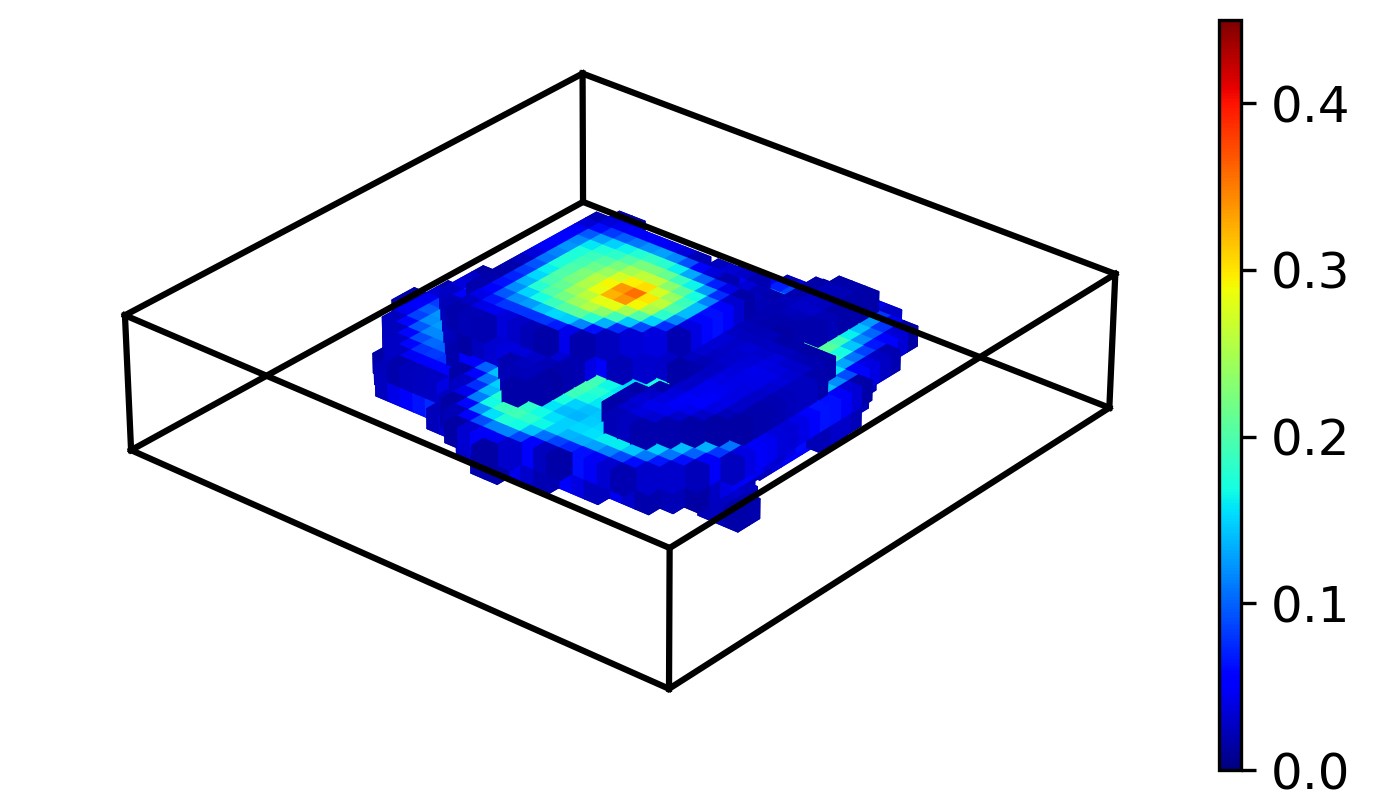}}
\caption{Comparison of surrogate model predictions (surr) and reference simulation results (sim) for interpreted seismic saturation fields for three realizations at 1~year.}
\label{fig:seis_pred_comparison}
\end{figure}

We now show results for test-set models. Comparison between predicted seismic data and reference results (generated by simulating the test-case model and then filtering the saturation field), for three randomly selected realizations, are shown in Figure~\ref{fig:seis_pred_comparison}. These results are at a time of 1~year (last time step). It is evident that the predicted plume shapes (lower row) for the different realizations closely resemble the reference results (upper row). We also see that the saturation fields differ significantly between realizations. 

\begin{figure}[htbp] 
\centering 
\vspace{0.35cm} 
\setlength{\lineskip}{\medskipamount}
\subcaptionbox{Realization 1 (sim)}{
\includegraphics[height=0.15\linewidth]{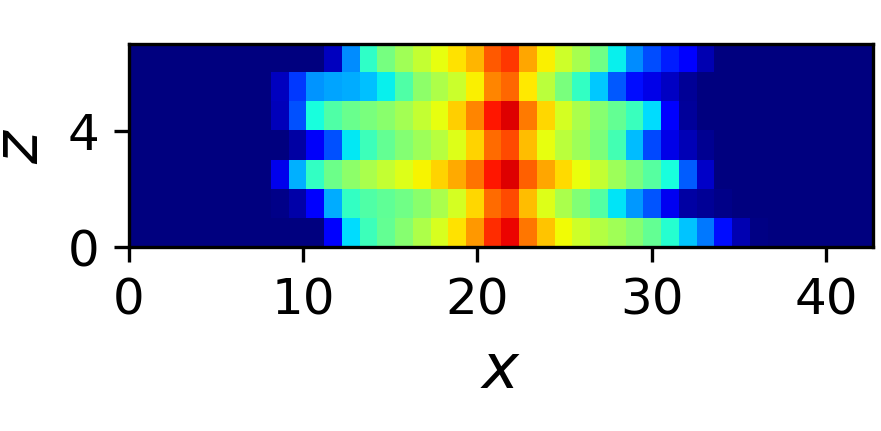}}
\subcaptionbox{Realization 2 (sim)}{
\includegraphics[height=0.15\linewidth]{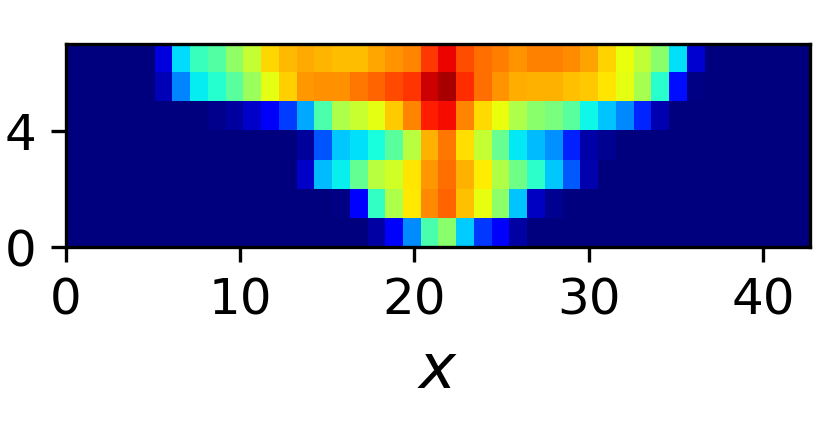}}
\subcaptionbox{Realization 3 (sim)}{
\includegraphics[height=0.15\linewidth]{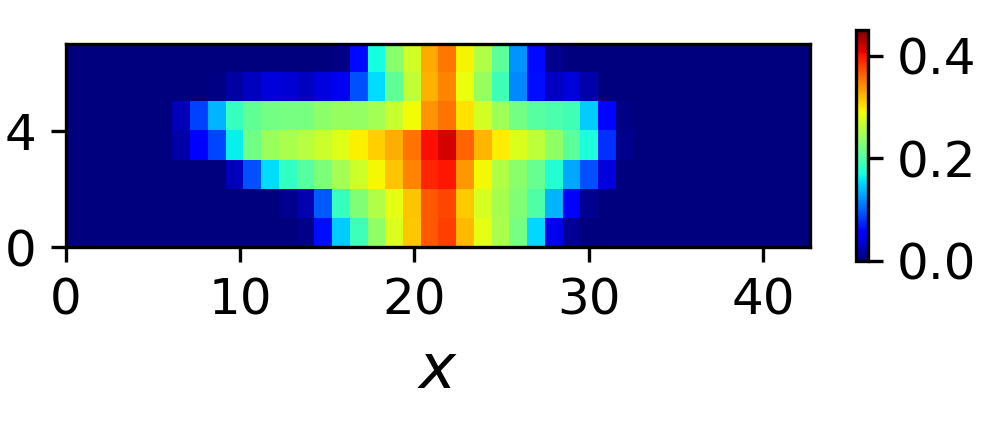}}

\subcaptionbox{Realization 1 (surr)}{
\includegraphics[height=0.15\linewidth]{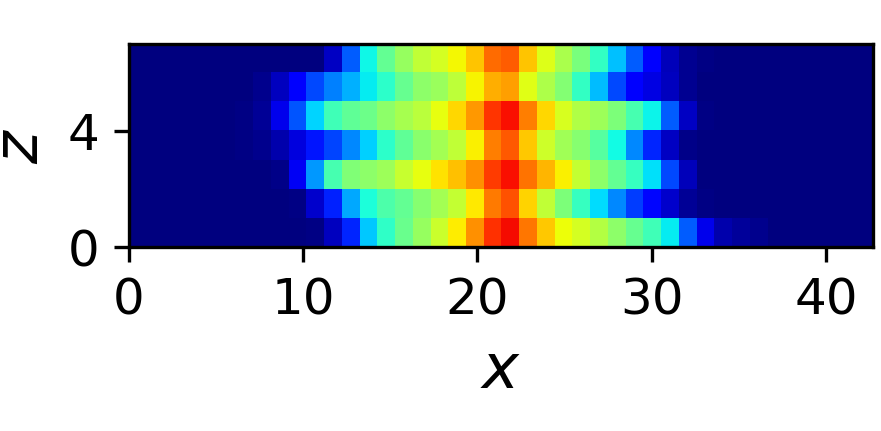}}
\subcaptionbox{Realization 2 (surr)}{
\includegraphics[height=0.15\linewidth]{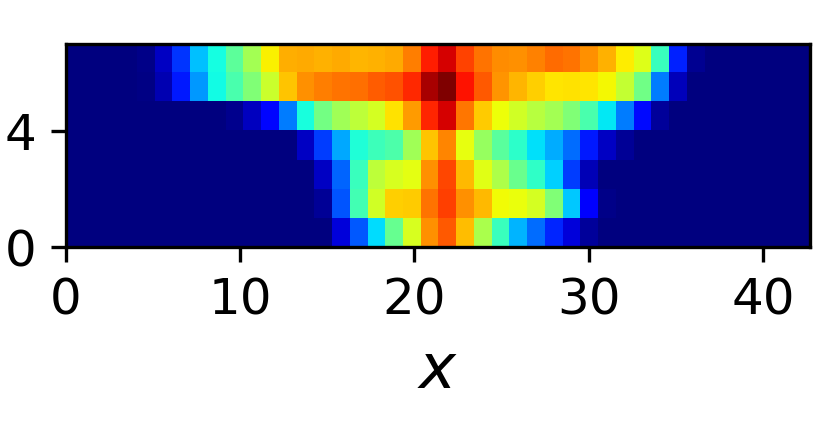}}
\subcaptionbox{Realization 3 (surr)}{
\includegraphics[height=0.15\linewidth]{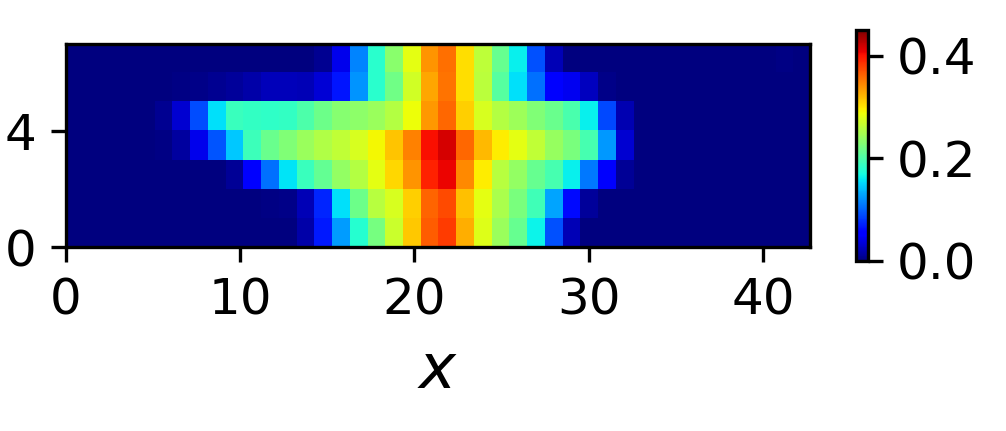}}
\caption{Comparison of surrogate model predictions (surr) and reference simulation results (sim) for interpreted seismic saturation fields, in $x$-$z$ cross sections through the injection well, for three realizations at 1~year.}
\label{fig:seis_pred_cross}
\end{figure}

Cross sections through the plumes for the three realizations, again at 1~year, are presented in Figure~\ref{fig:seis_pred_cross}. These results are for the $x$-$z$ cross section through the injection well. Predictions from the surrogate model are in reasonable agreement with reference results, both for relatively uniform plumes (Figure~\ref{fig:seis_pred_cross}(a)) and for irregular shaped plumes (Figure~\ref{fig:seis_pred_cross}(b) and (c)).

\begin{figure}[htbp] 
\centering 
\vspace{0.35cm} 
\setlength{\lineskip}{\medskipamount}
\subcaptionbox{2 months (sim)}{
\includegraphics[height=0.135\linewidth]{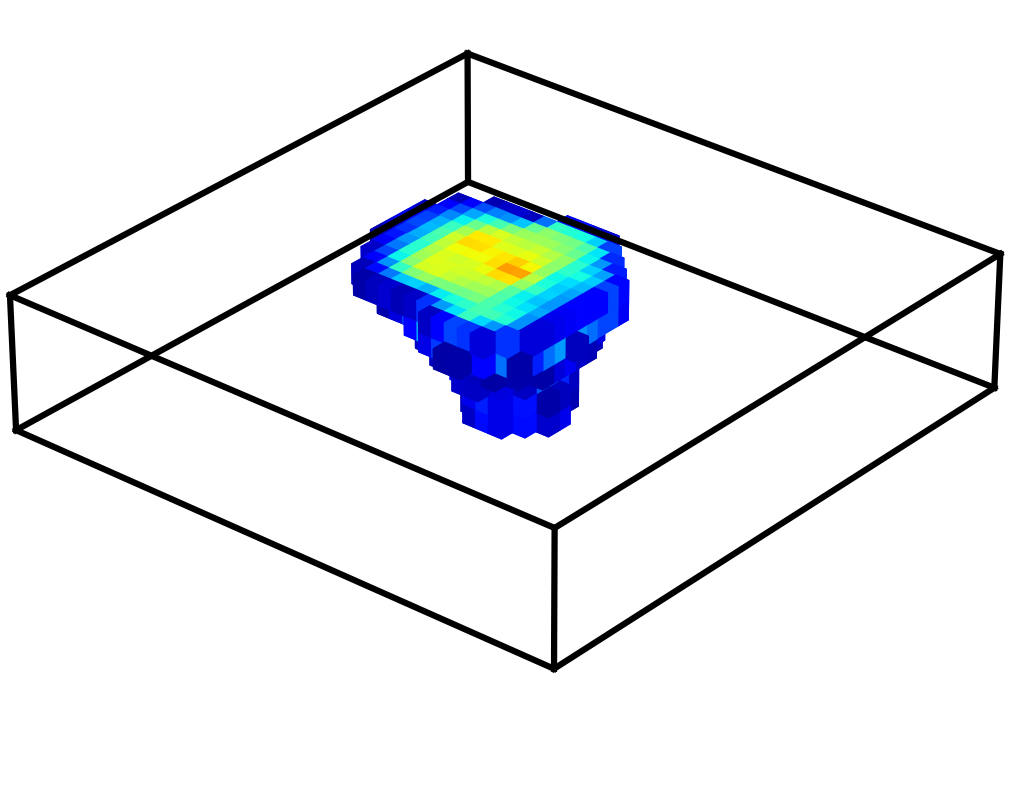}}
\subcaptionbox{4 months (sim)}{
\includegraphics[height=0.135\linewidth]{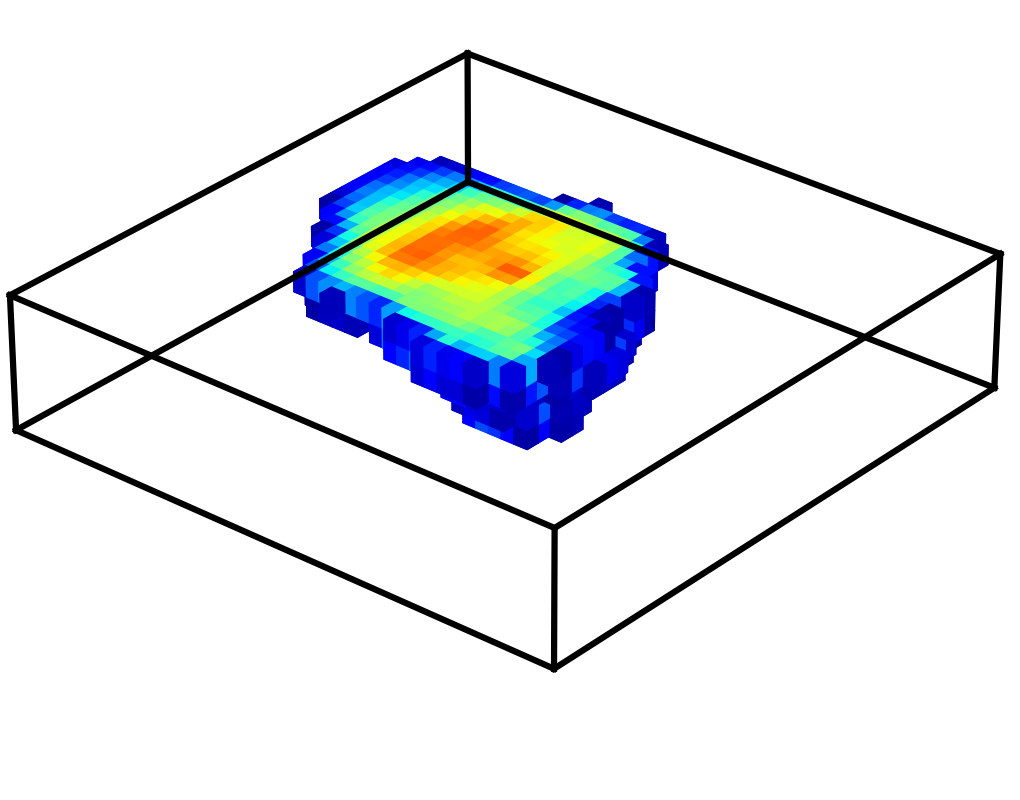}}
\subcaptionbox{6 months (sim)}{
\includegraphics[height=0.135\linewidth]{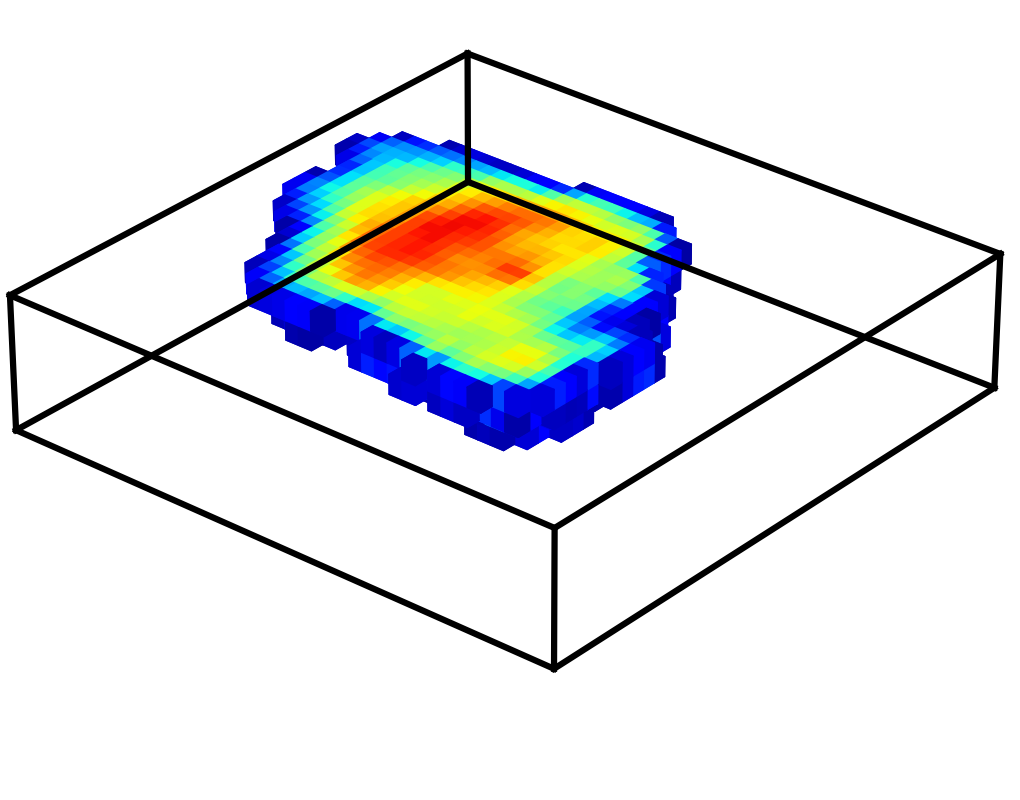}}
\subcaptionbox{8 months (sim)}{
\includegraphics[height=0.135\linewidth]{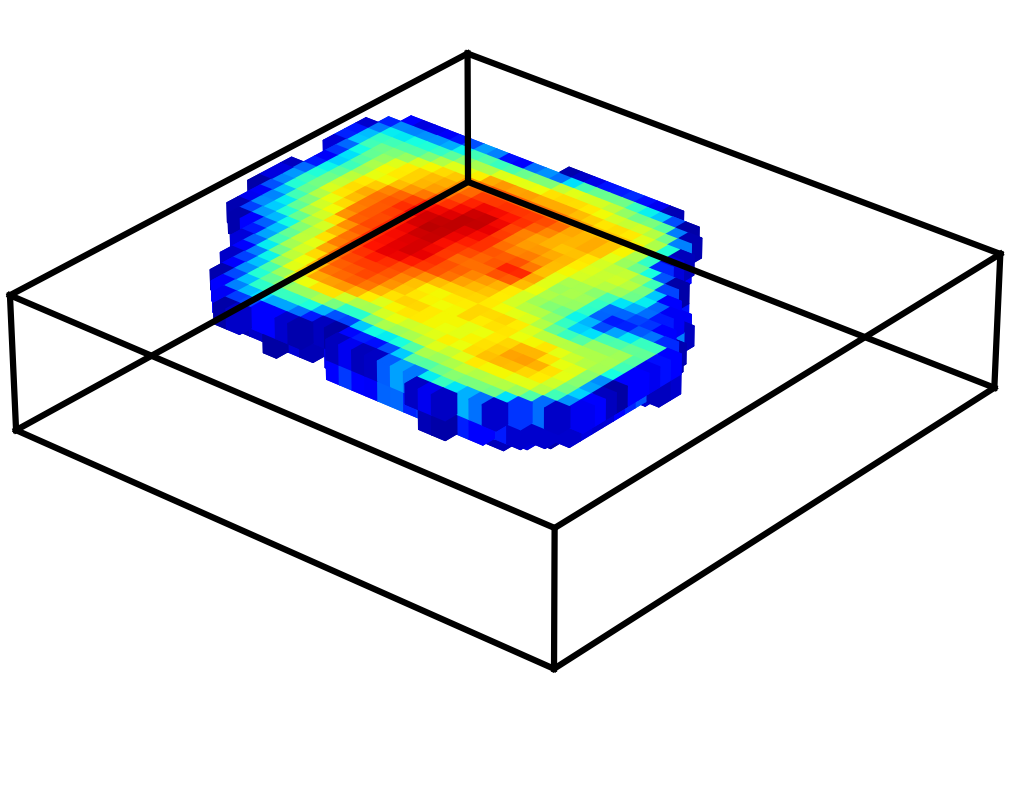}}
\subcaptionbox{10 months (sim)}{
\includegraphics[height=0.135\linewidth]{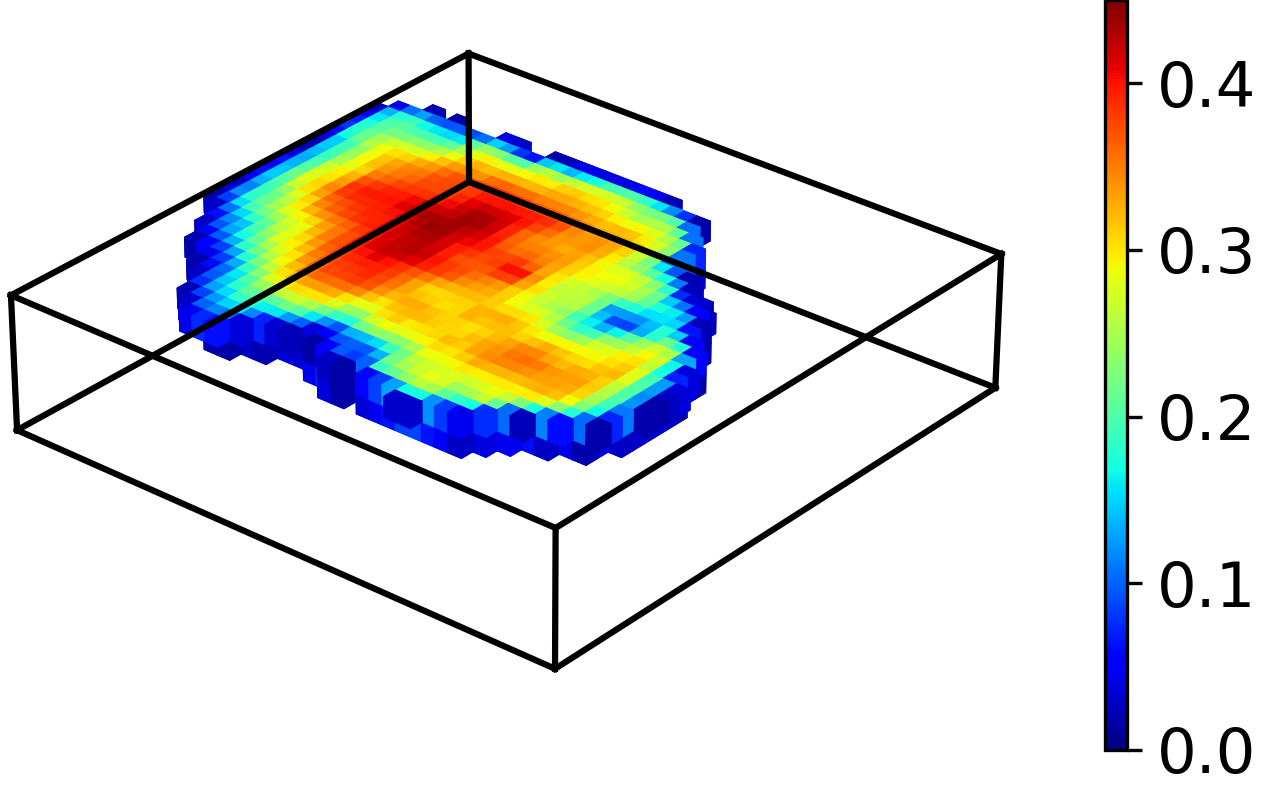}}

\subcaptionbox{2 months (surr)}{
\includegraphics[height=0.135\linewidth]{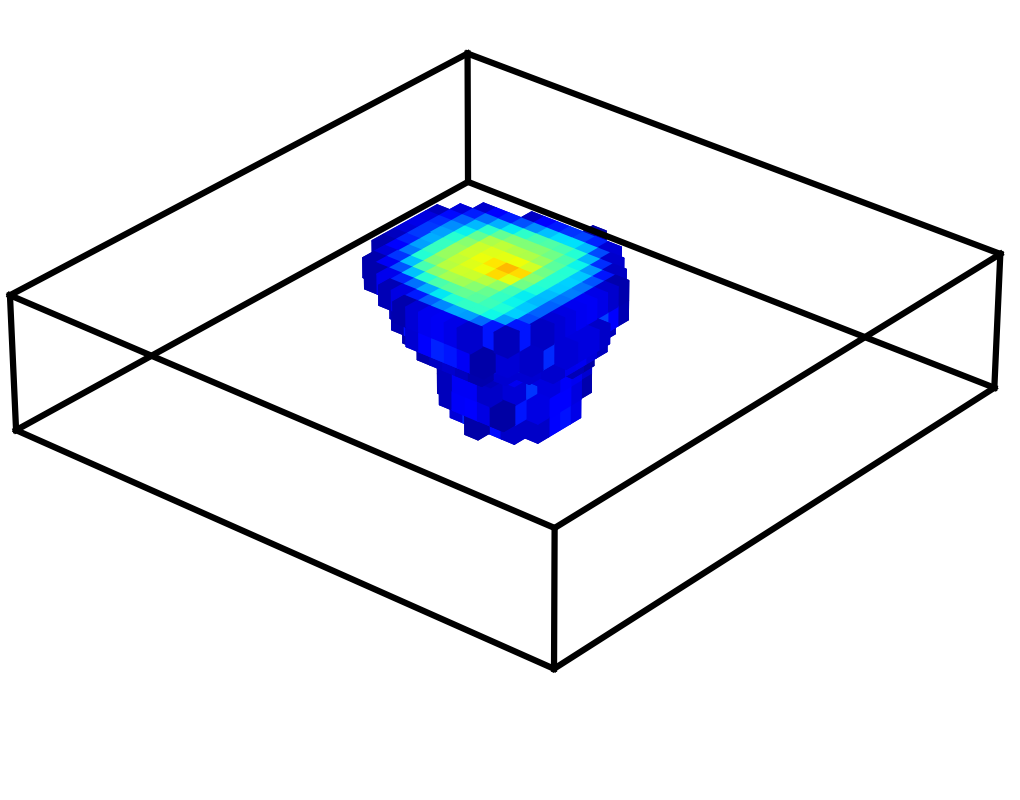}}
\subcaptionbox{4 months (surr)}{
\includegraphics[height=0.135\linewidth]{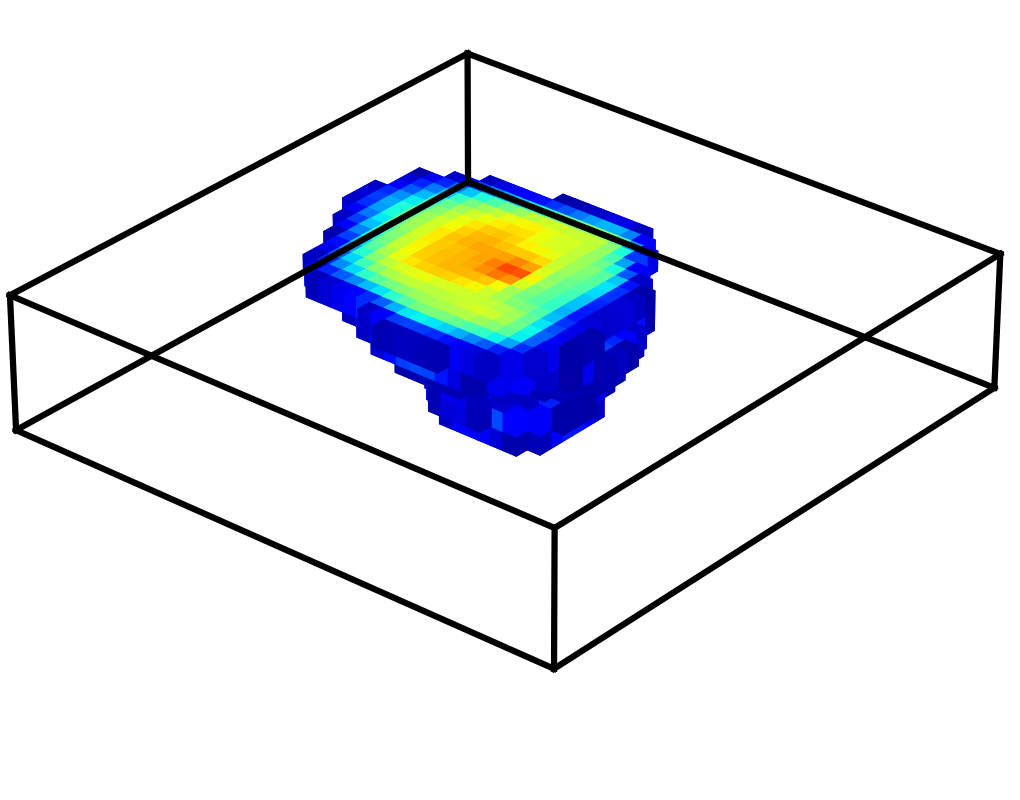}}
\subcaptionbox{6 months (surr)}{
\includegraphics[height=0.135\linewidth]{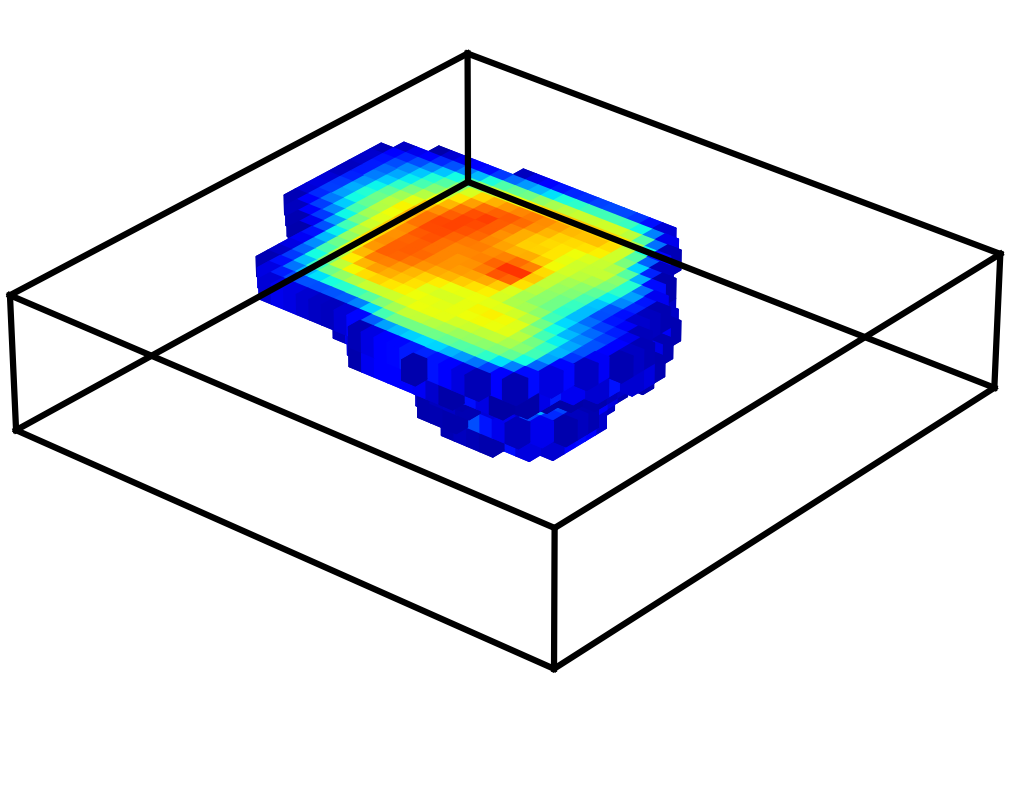}}
\subcaptionbox{8 months (surr)}{
\includegraphics[height=0.135\linewidth]{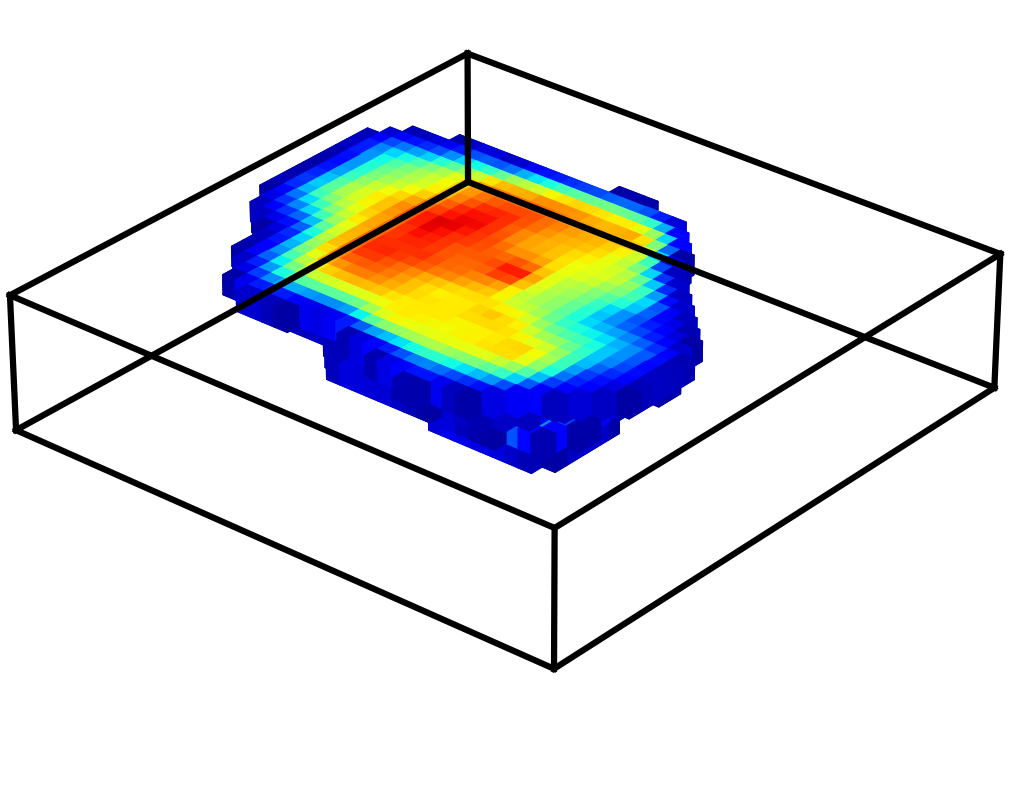}}
\subcaptionbox{10 months (surr)}{
\includegraphics[height=0.135\linewidth]{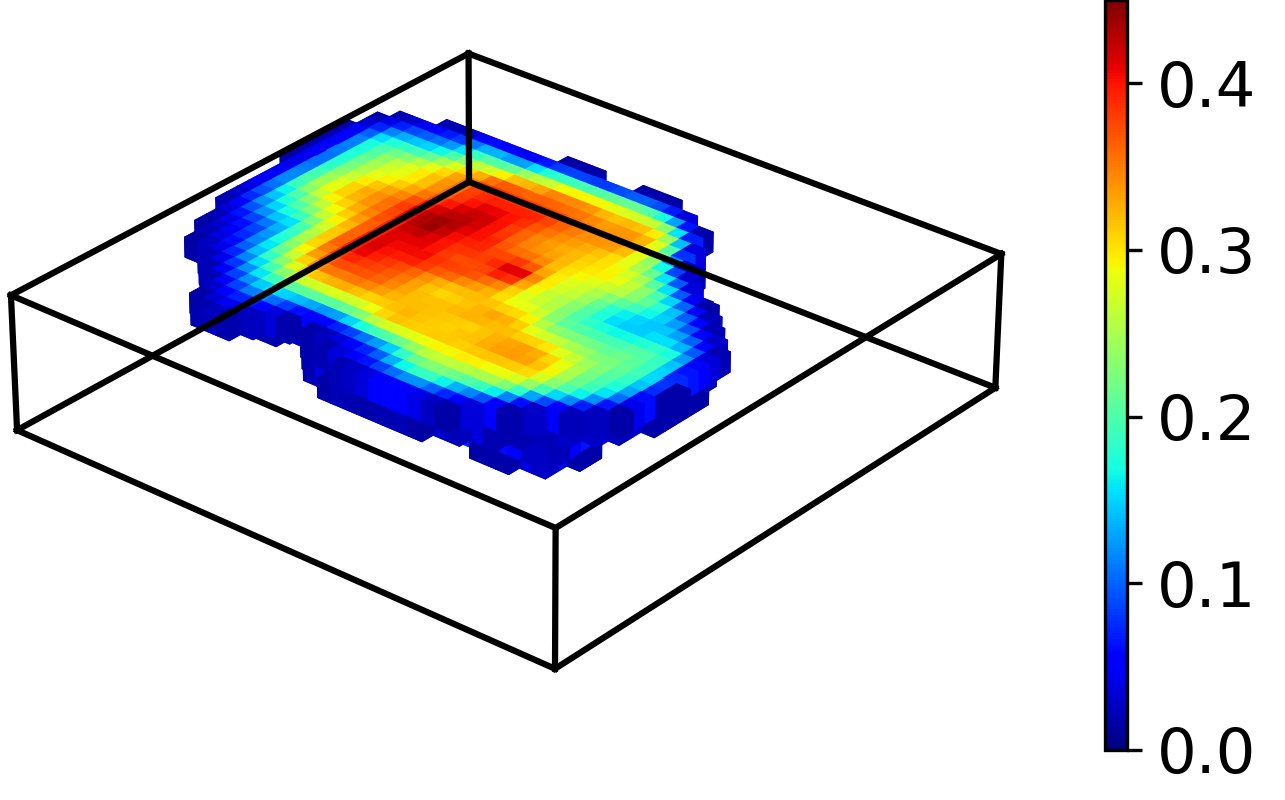}}

\caption{Comparison of time evolution between surrogate model predictions (surr) and reference simulation results (sim) for interpreted seismic saturation fields (Realization~2).}
    \label{fig:time_lapse_pred}
\end{figure}

We next present time-lapse results for a specific geomodel (Realization~2 in Figures~\ref{fig:seis_pred_comparison} and \ref{fig:seis_pred_cross}). Interpreted saturation fields at five time steps are displayed in Figure~\ref{fig:time_lapse_pred}. Although slight differences are evident at some steps, it is clear that the general time evolution is captured accurately by the surrogate model.

\subsubsection{Monitoring well data surrogate evaluation}
\label{sec:mon_surr_evl}

The surrogate model for monitoring well data, described in Section~\ref{sec:surr-mon}, is trained to predict vertically resolved CO$_2$ saturation at 16~time steps (time steps~0, 2, 4, 6, $\ldots$, 30). The first time step corresponds to the initial condition. The network contains 16~output channels, with each channel predicting saturation at a particular time step. For training, the batch size is set to 10, the model is trained for 150~epochs, and the initial learning rate is 0.001. This model requires only about 15~minutes for training on a Nvidia A100 GPU. Training is very fast in this case because this surrogate uses only local geomodel properties as inputs rather than the entire field, and it predicts saturations only for a single column of grid blocks (at 16~time steps). 

\begin{figure}[htbp] 
\centering 
\vspace{0.35cm} 
\setlength{\lineskip}{\medskipamount}
\subcaptionbox{Realization 4 (sim)}{
\includegraphics[height=0.35\linewidth]{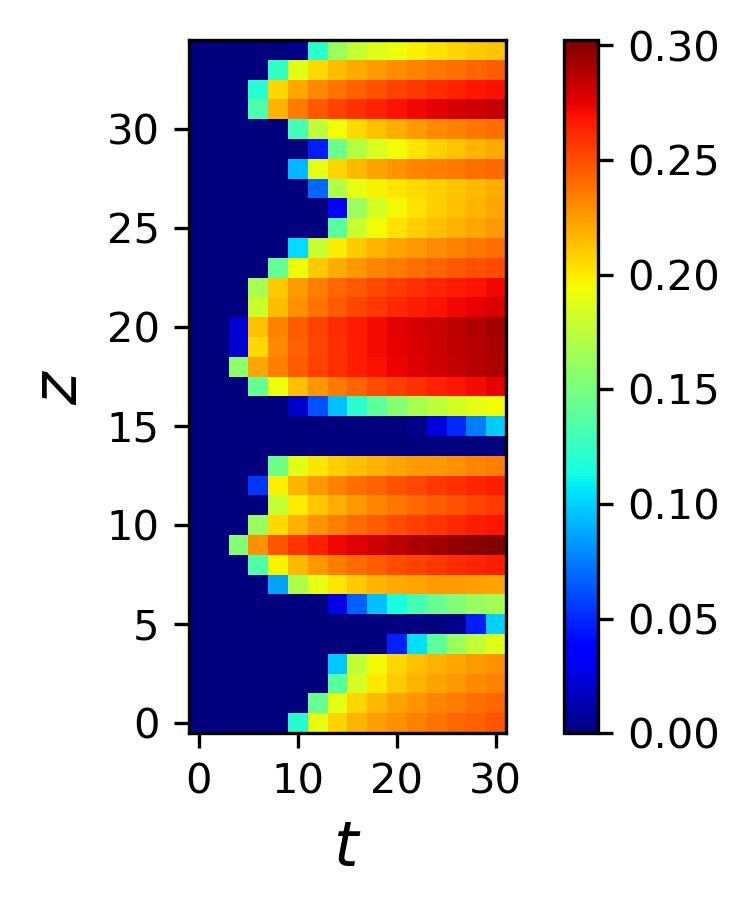}}
\subcaptionbox{Realization 5 (sim)}{
\includegraphics[height=0.35\linewidth]{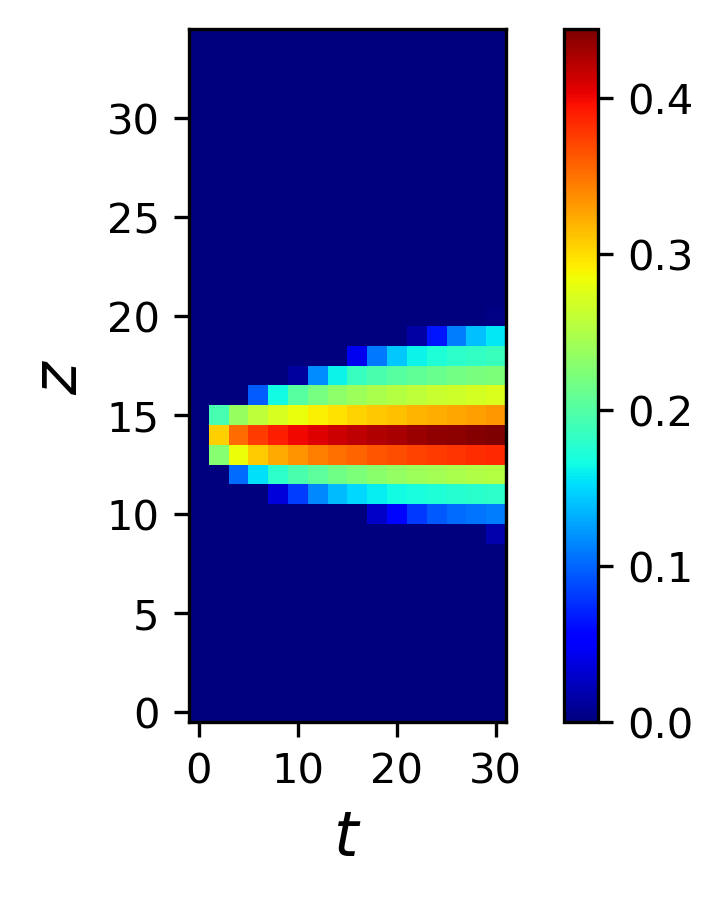}}
\subcaptionbox{Realization 6 (sim)}{
\includegraphics[height=0.35\linewidth]{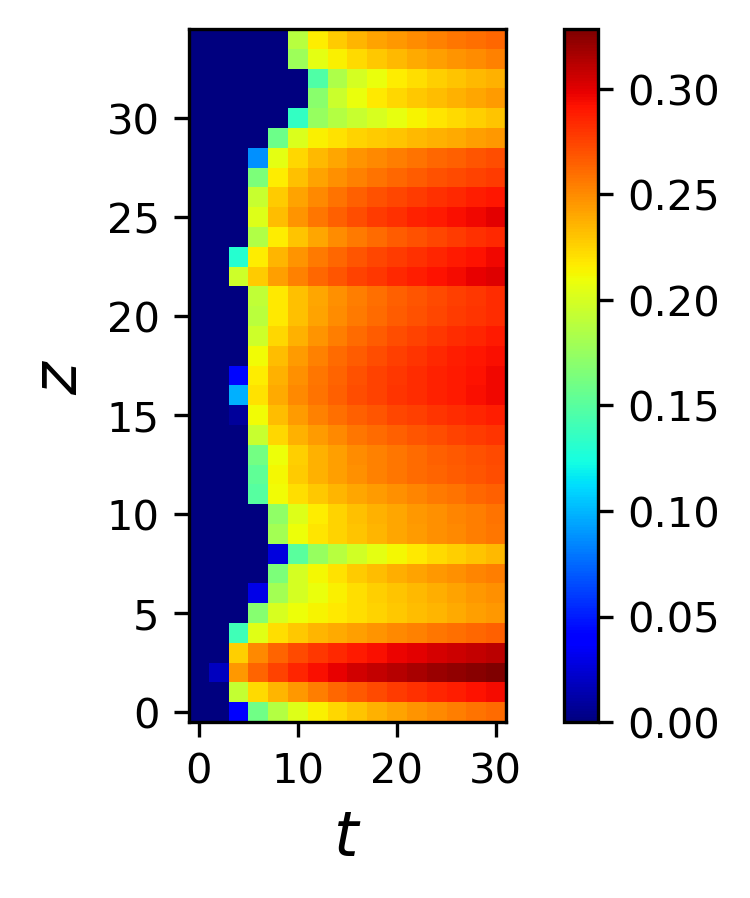}}

\subcaptionbox{Realization 4 (surr)}{
\includegraphics[height=0.35\linewidth]{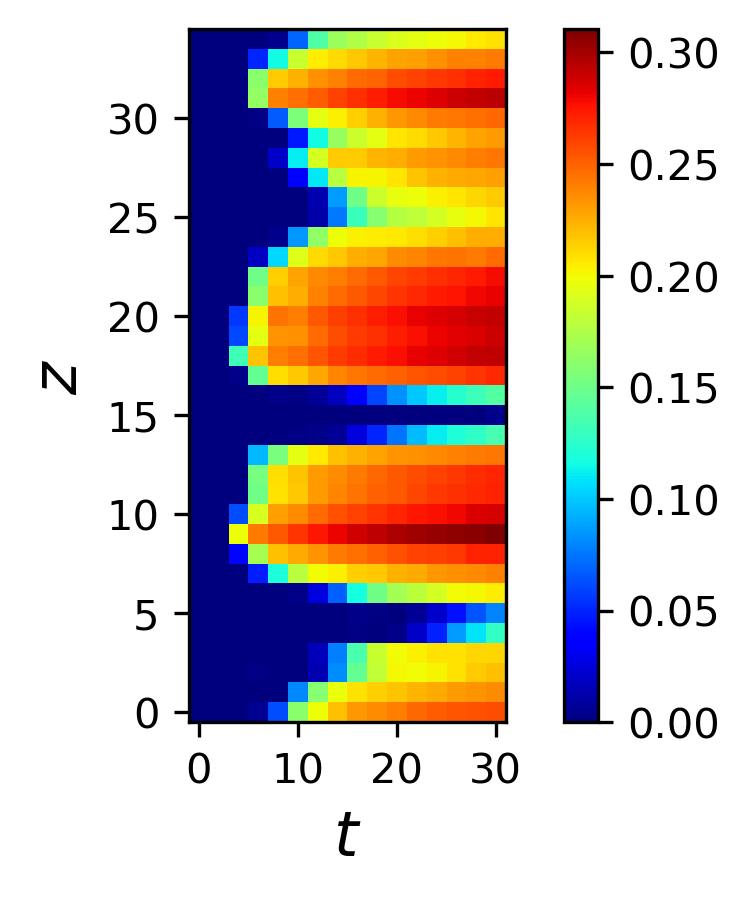}}
\subcaptionbox{Realization 5 (surr)}{
\includegraphics[height=0.35\linewidth]{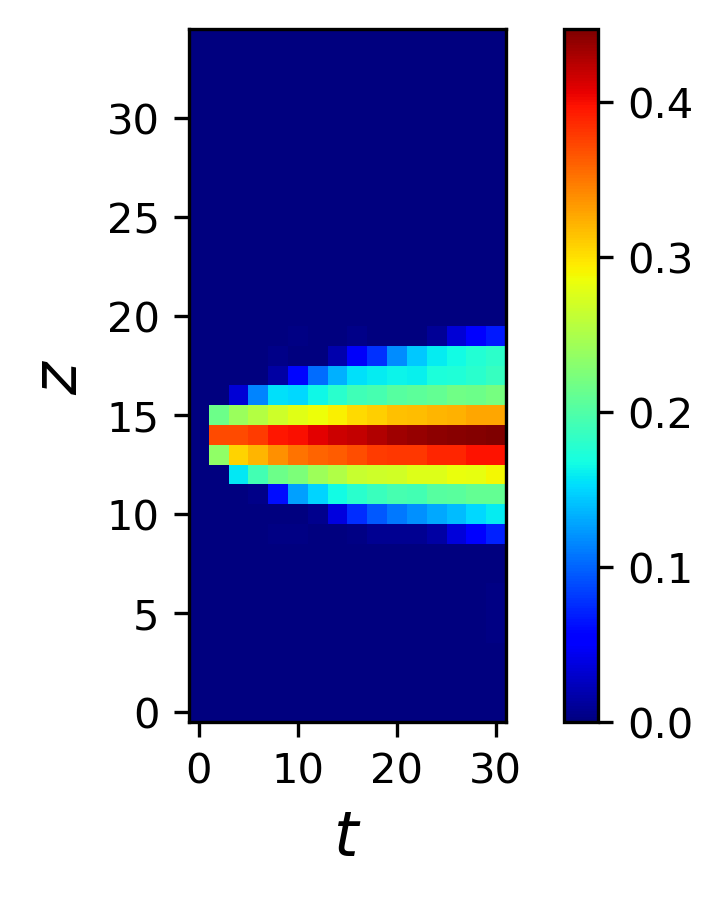}}
\subcaptionbox{Realization 6 (surr)}{
\includegraphics[height=0.35\linewidth]{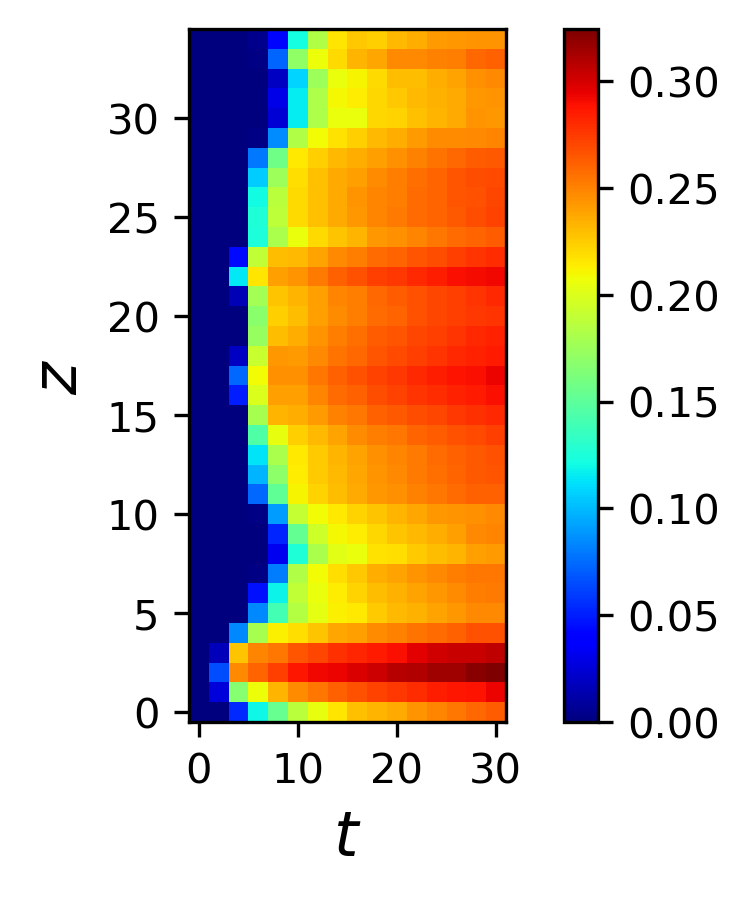}}
\caption{Comparison of time evolution between monitoring well surrogate saturation predictions (surr) and reference simulation results (sim) for three realizations. Each column (in each subfigure) corresponds to one of the 16~time steps considered.}
\label{fig:mon_pred_compare}
\end{figure}

Figure~\ref{fig:mon_pred_compare} shows the comparison between the predicted and true CO$_2$ saturation profiles at the monitoring well over time for three new randomly selected test-case realizations. The columns in each figure represent a single time step, and results are shown for all 35~layers in the model. We see very different behaviors in the three realizations -- fairly uniform arrival times and saturation distributions in Realization~6, somewhat less uniformity in Realization~4, and a very localized saturation peak in Realization~5. For Realization~5, CO$_2$ has arrived at the monitoring well in only a few layers. Despite the high degree of variability between realizations, the surrogate model provides predictions in close visual agreement with reference simulation results in all cases.

It is useful to quantify the level of agreement between reference results and surrogate model predictions. To accomplish this, we compute the saturation mean absolute error (MAE, denoted as $\delta_s$) between the two sets of results as follows: 
\begin{center}
\begin{equation}
\setlength{\abovedisplayskip}{-9pt}
\delta_s= \frac{1}{n}\sum_{i=1}^{n}\lvert S_i-\hat{S}_i \rvert, \ \ {\rm for} \ S_i > \varepsilon  \ {\rm or} \ {\hat S}_i > \varepsilon,
\label{eq:relative_error}
\end{equation}
\end{center}
where $S_i$ and ${\hat S}_i$ denote the reference and surrogate model prediction, respectively. Here the contribution to $\delta_s$ is calculated only for grid cells where either simulated or predicted saturation is higher than a threshold $\varepsilon$. Consistent with this, $n$ is the total number of cells where $S_i > \varepsilon$ or ${\hat S}_i > \varepsilon$. Thus, the value of $n$ will be different for different realizations. A value for $\delta_s$ is computed for each test case for both surrogate models.

\begin{figure}[htbp]
    \centering
    \vspace{0.35cm} 
    \includegraphics[width =0.6\textwidth]{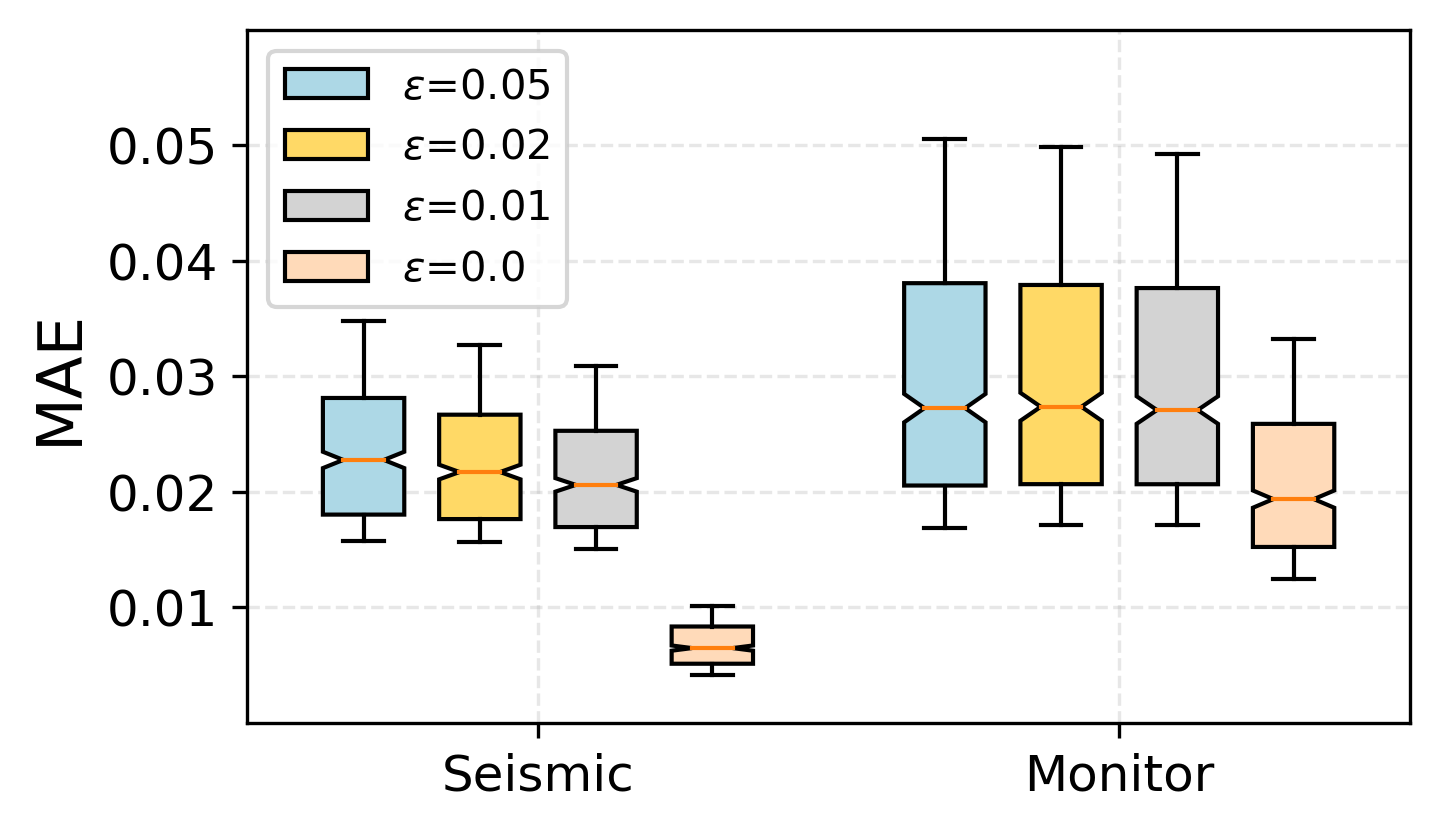}
    \caption{Box plots of mean absolute saturation errors with different thresholds ($\varepsilon$) for interpreted seismic surrogate (Seismic) and monitoring well surrogate (Monitor) over 500 test-case models.}
    \label{fig:quant_evaluate}
\end{figure}

The MAEs for both the interpreted seismic surrogate and the monitoring well surrogate, over the 500 test-case realizations, are shown as box plots in Figure~\ref{fig:quant_evaluate}. The various boxes for each case correspond to different thresholds ($\varepsilon=0$, 0.01, 0.02, 0.05).  In the box plots, the top and bottom of the box represent the 75th and 25th percentile (P$_{75}$ and P$_{25}$) errors and the lines above and below the boxes represent the 90th and 10th percentile errors. The orange lines inside the boxes correspond to the median error. The median MAEs range from 0.006 to 0.023 for interpreted seismic data, and from 0.019 to 0.028 for monitoring well data. Error is lowest for $\varepsilon=0$ because many cells with near-zero saturation (and thus near-zero error) contribute to the MAE calculation. The errors remain fairly consistent and relatively low for different nonzero threshold values. As we will see, this level of accuracy is more than sufficient to enable the use of the surrogate models for history matching.

\subsection{Robustness of surrogates towards noise}
\label{sec:robustness}

\begin{figure}[htbp] 
\centering 
\vspace{0.35cm} 
\setlength{\lineskip}{\medskipamount}
\subcaptionbox{Realization 1 (original)}{
\includegraphics[height=0.15\linewidth]{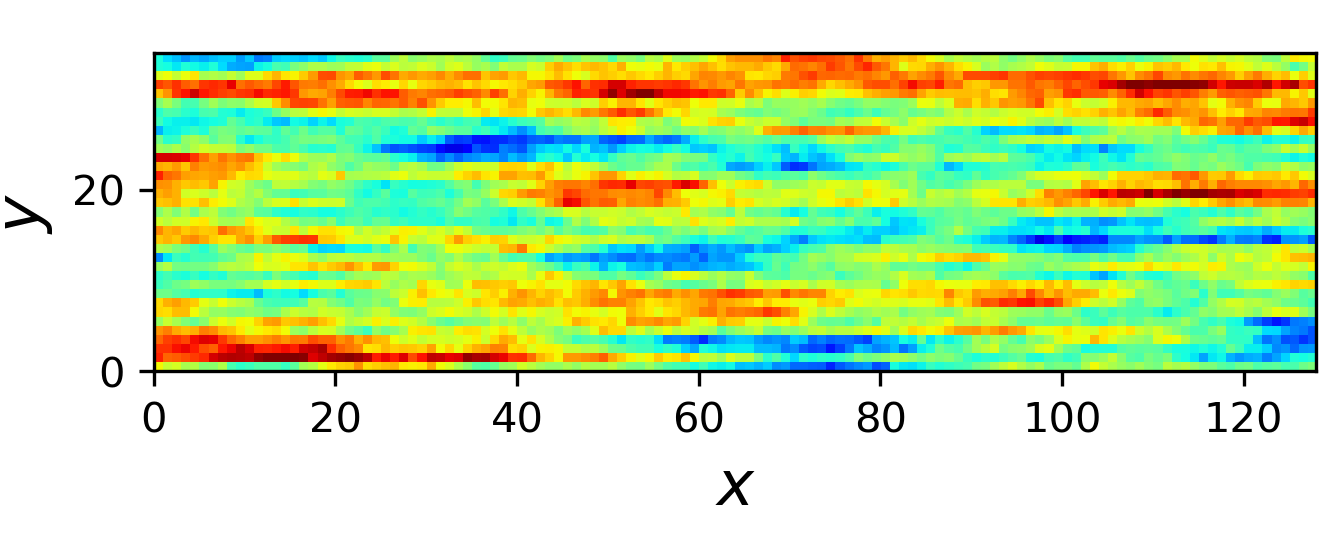}}
\subcaptionbox{Realization 1 (perturbed)}{
\includegraphics[height=0.15\linewidth]{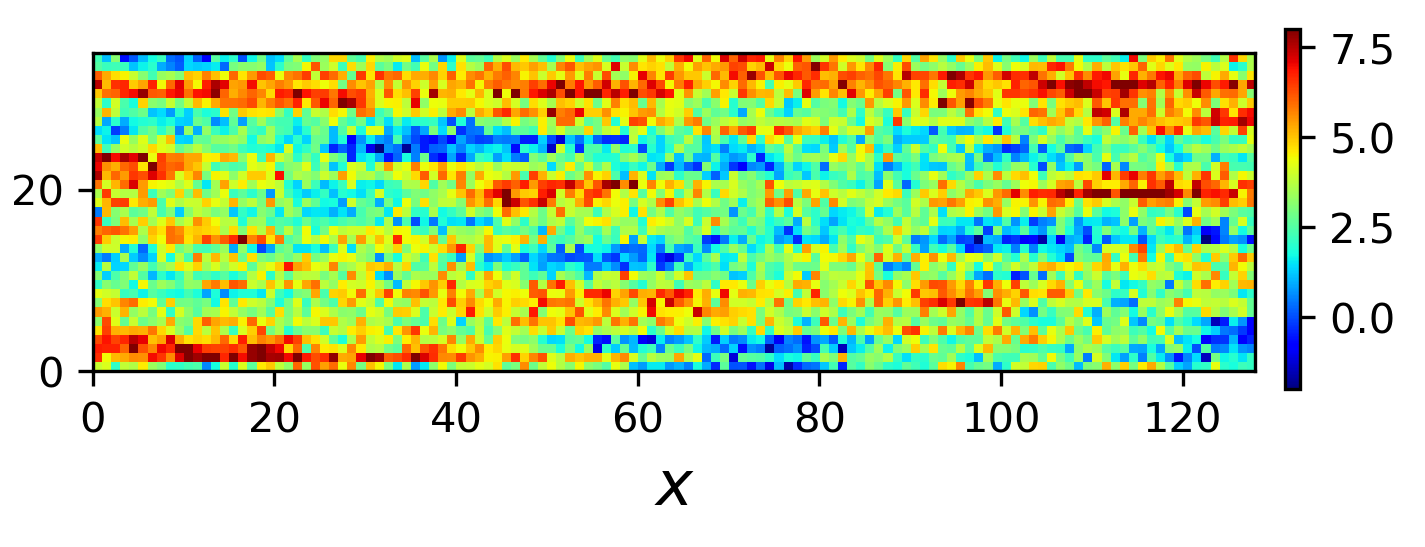}}

\subcaptionbox{Realization 2 (original)}{
\includegraphics[height=0.15\linewidth]{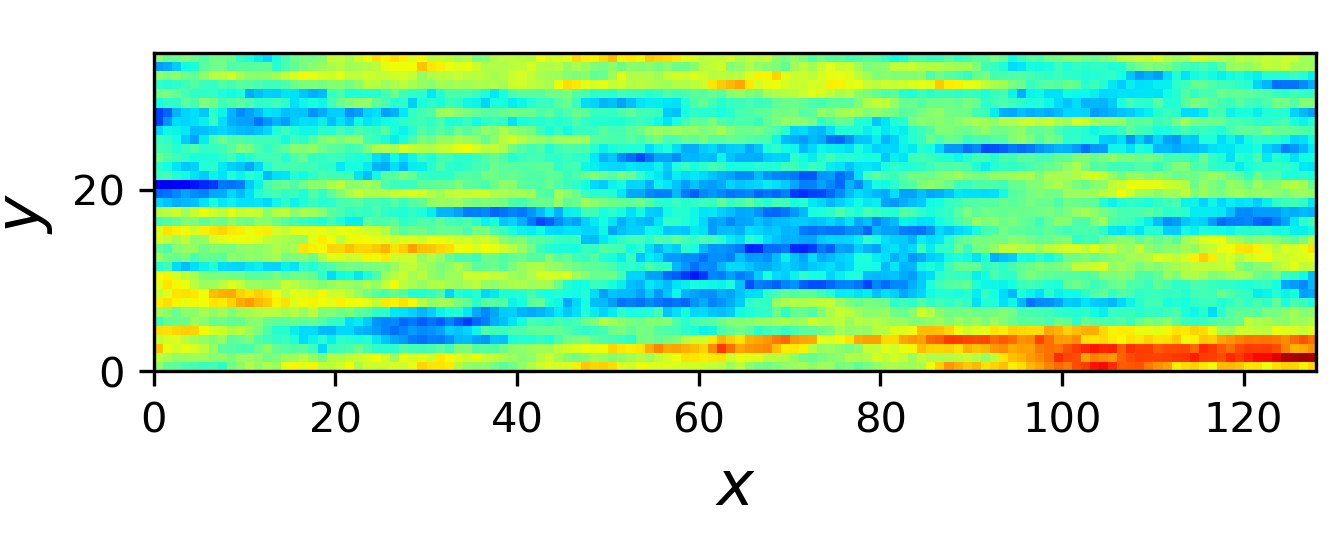}}
\subcaptionbox{Realization 2 (perturbed)}{
\includegraphics[height=0.15\linewidth]{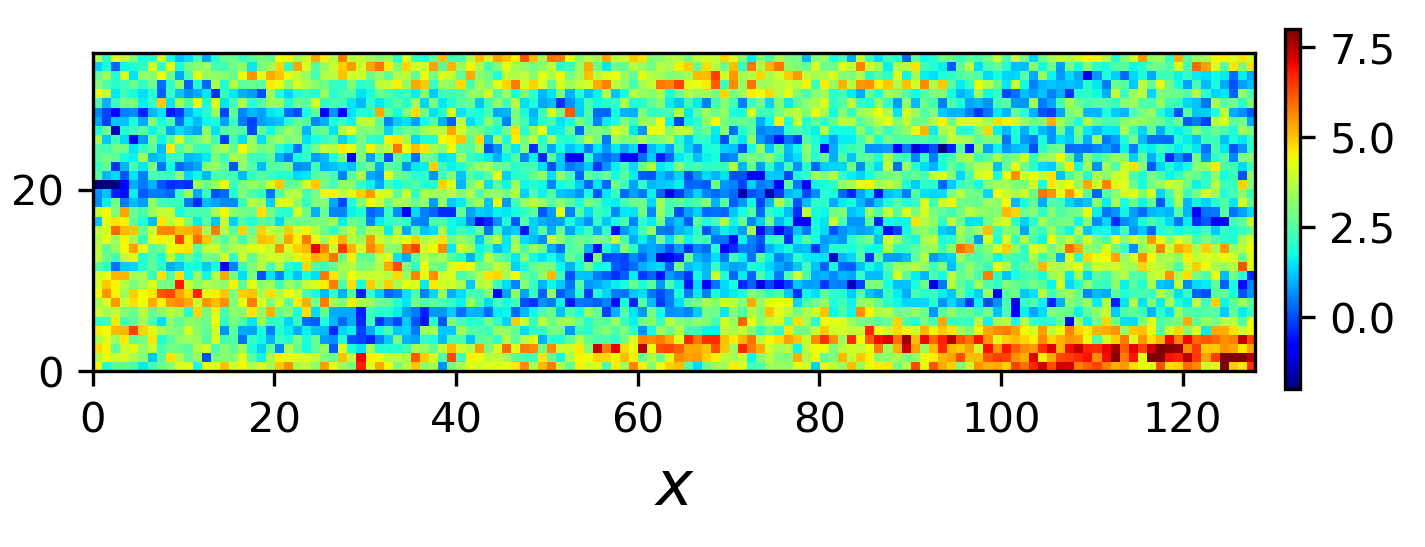}}

\caption{Original and perturbed geomodels. The quantity displayed is $\log k$ for the $x$-$z$ cross section through the injection well.}
\label{fig:noisy_logk}
\end{figure}

In the previous subsection, we evaluated the performance of our surrogate models and showed that they provide a high degree of accuracy relative to the underlying simulation results. It is also useful to evaluate surrogate model robustness to `noise' in the input geomodel. In such cases the geomodel will not exactly honor the assumed parameterization $\mathbf{m}=\mathbf{m}(\mathbf{h}, \boldsymbol \xi)$. For this assessment, we add Gaussian noise with zero mean and a standard deviation of 10\% of the maximum value to the $\log k$, $\phi$, and $\log_{10}a_r$ values in all cells in a set of test-case realizations. Two sets of realizations (original, using $\mathbf{m}=\mathbf{m}(\mathbf{h}, \boldsymbol \xi)$, and perturbed) are shown in Figure~\ref{fig:noisy_logk}. The quantity displayed is $\log k$ in an $x$-$z$ cross-section of the model. Fine-scale noise is clearly evident in the perturbed realizations.

\begin{figure}[htbp] 
\centering 
\vspace{0.35cm} 
\setlength{\lineskip}{\medskipamount}
\subcaptionbox{Real.~1 (sim, original~$\mathbf{m}$)}{
\includegraphics[height=0.135\linewidth]{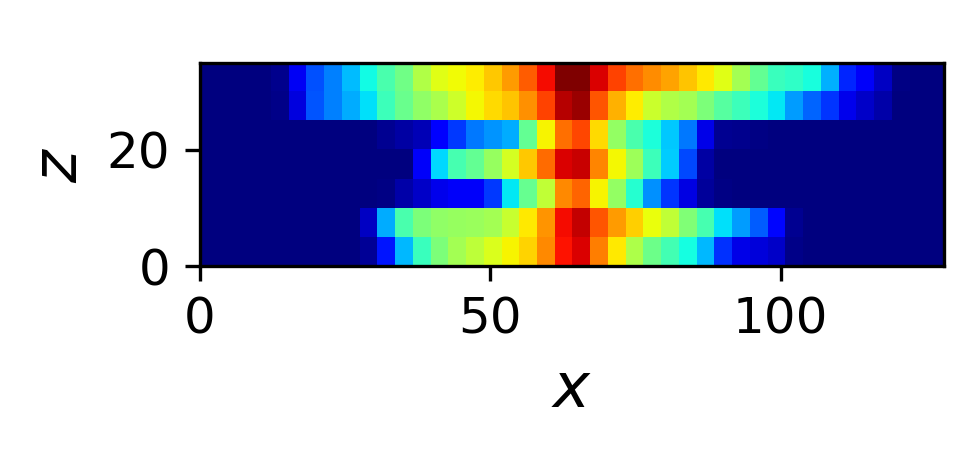}}
\subcaptionbox{Real.~1 (surr, original~$\mathbf{m}$)}{
\includegraphics[height=0.135\linewidth]{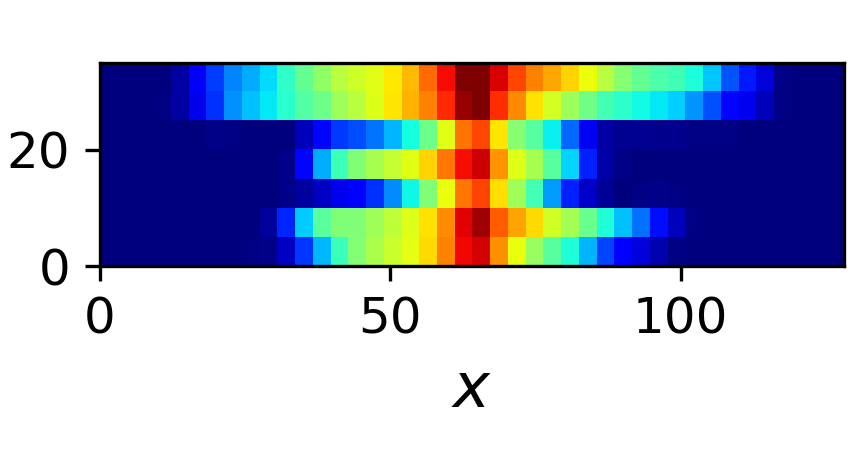}}
\subcaptionbox{Real.~1 (surr, perturbed~$\mathbf{m}$)}{
\includegraphics[height=0.135\linewidth]{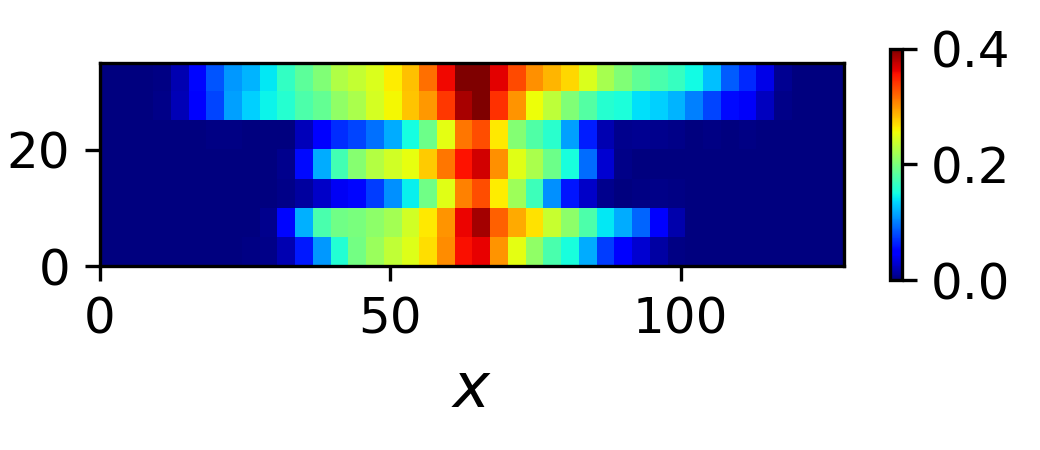}}

\subcaptionbox{ Real.~2 (sim, original $\mathbf{m}$)}{
\includegraphics[height=0.135\linewidth]{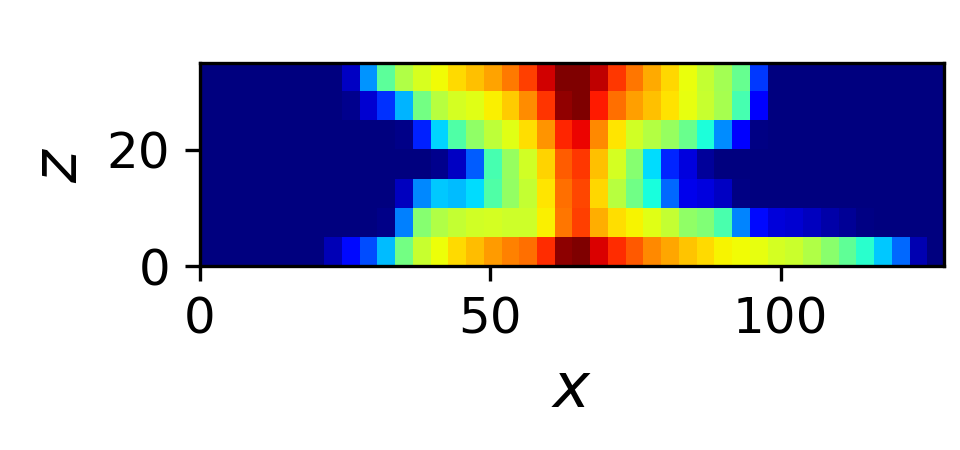}}
\subcaptionbox{Real.~2 (surr, original $\mathbf{m}$)}{
\includegraphics[height=0.135\linewidth]{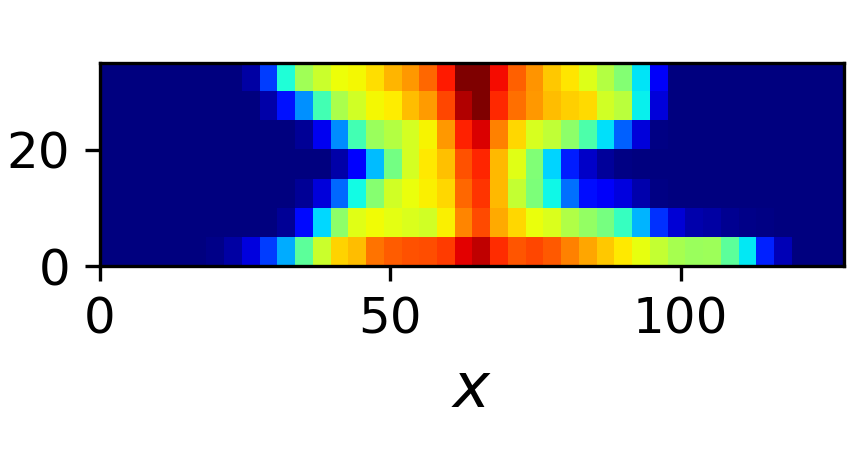}}
\subcaptionbox{Real.~2 (surr, perturbed $\mathbf{m}$)}{
\includegraphics[height=0.135\linewidth]{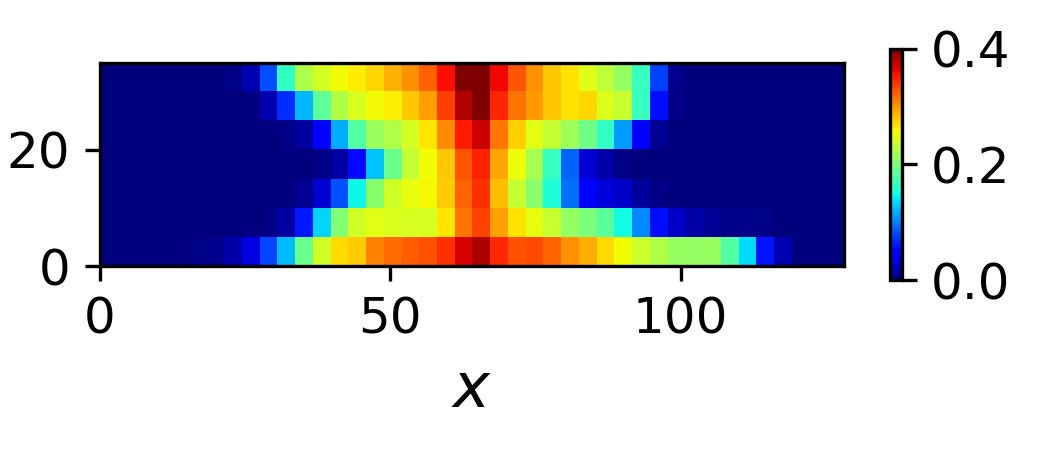}}

\caption{Comparison of reference simulation results (sim, original $\mathbf{m}$) and surrogate predictions with original geomodel (surr, original $\mathbf{m}$) and perturbed geomodel (surr, perturbed $\mathbf{m}$). Results are for interpreted seismic saturation fields in the $x$-$z$ cross section through the injection well at 1~year for two different realizations.}
\label{fig:noise_impact}
\end{figure}

Next, using the surrogate trained on realizations satisfying $\mathbf{m}=\mathbf{m}(\mathbf{h}, \boldsymbol \xi)$, we predict seismic saturation fields for both the original and perturbed models. These results are displayed in Figure~\ref{fig:noise_impact}, along with (filtered) GEOS simulation results for the original model. We see that the predictions for the perturbed models are almost identical to those for the original models, which demonstrates the robustness of the surrogate to noise (and small variations) in the input geomodel. Similar results are also observed for the monitoring well surrogate, though for brevity these are not presented here.

\section{Use of surrogate models for history matching}
\label{sec:hm}

Here we first describe the problem setup, our treatment of the various error components appearing in the history matching formulation, and the representations used for the likelihood function. History matching results using the hierarchical MCMC procedure, with the deep learning surrogates applied for all forward simulations, are then presented. 

\subsection{Setup, errors and likelihood function}
\label{sec:HM_setup}

\begin{figure}[htbp] 
\centering 
\vspace{0.35cm} 
\setlength{\lineskip}{\medskipamount}
\subcaptionbox{High-fidelity $y$-$z$ cross section}{
\includegraphics[width=0.48\linewidth]{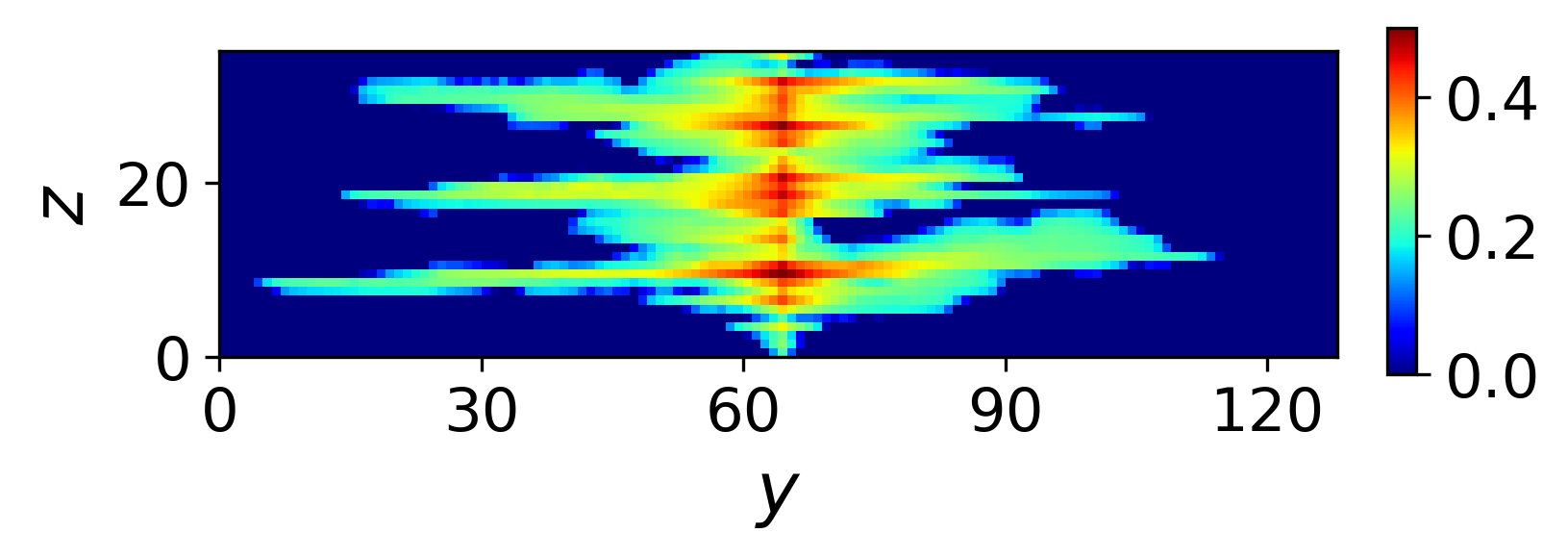}}
\subcaptionbox{High-fidelity $x$-$z$ cross section}{
\includegraphics[width=0.48\linewidth]{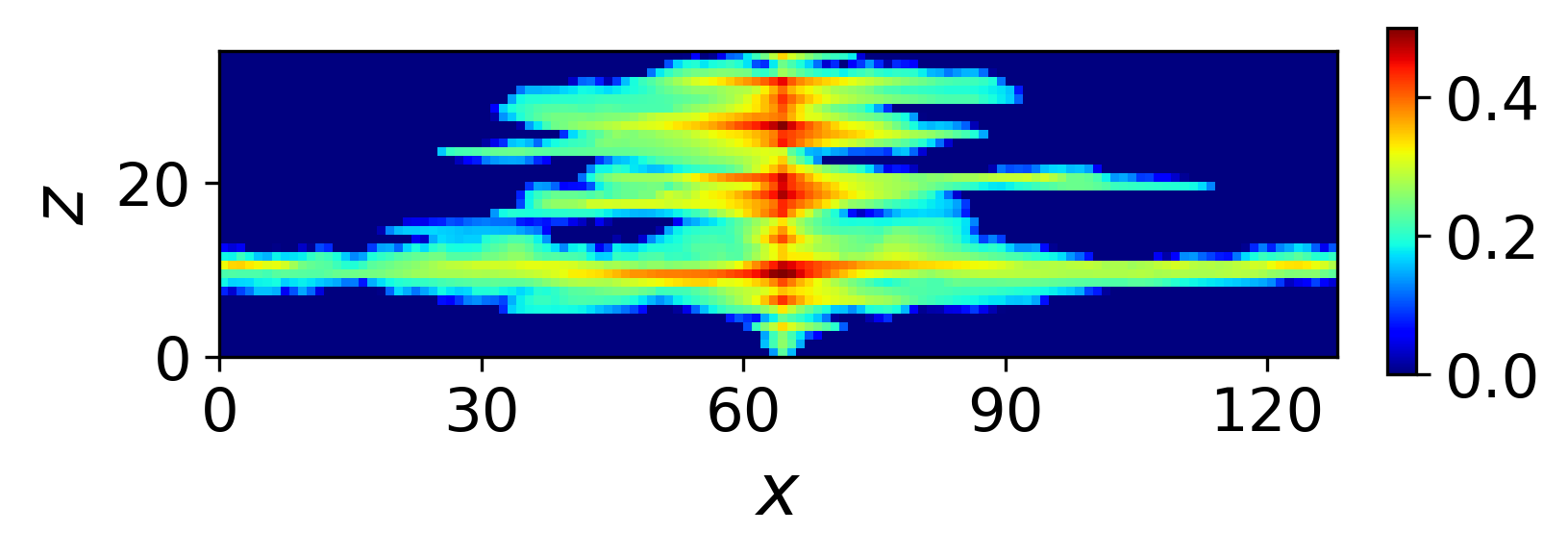}}

\subcaptionbox{Filtered $y$-$z$ cross section}{
\includegraphics[width=0.47\linewidth]{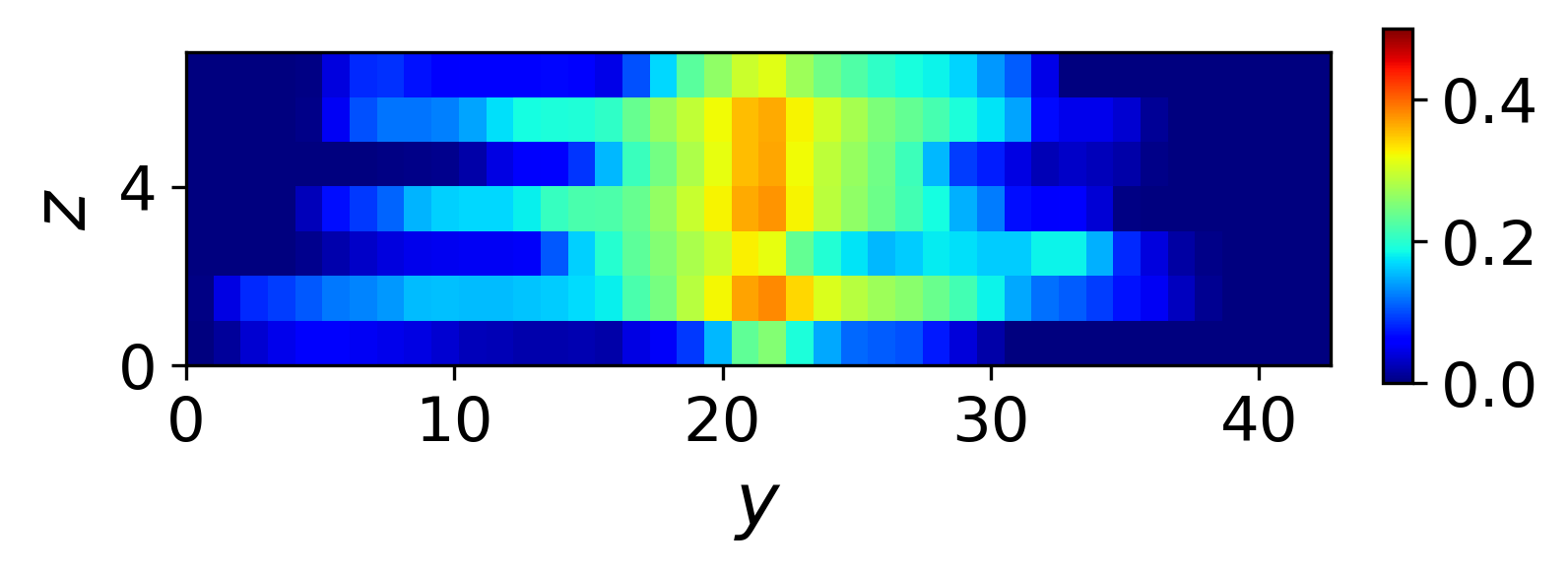}}
\hspace{0.0001\linewidth}
\subcaptionbox{Filtered $x$-$z$ cross section}{
\includegraphics[width=0.47\linewidth]{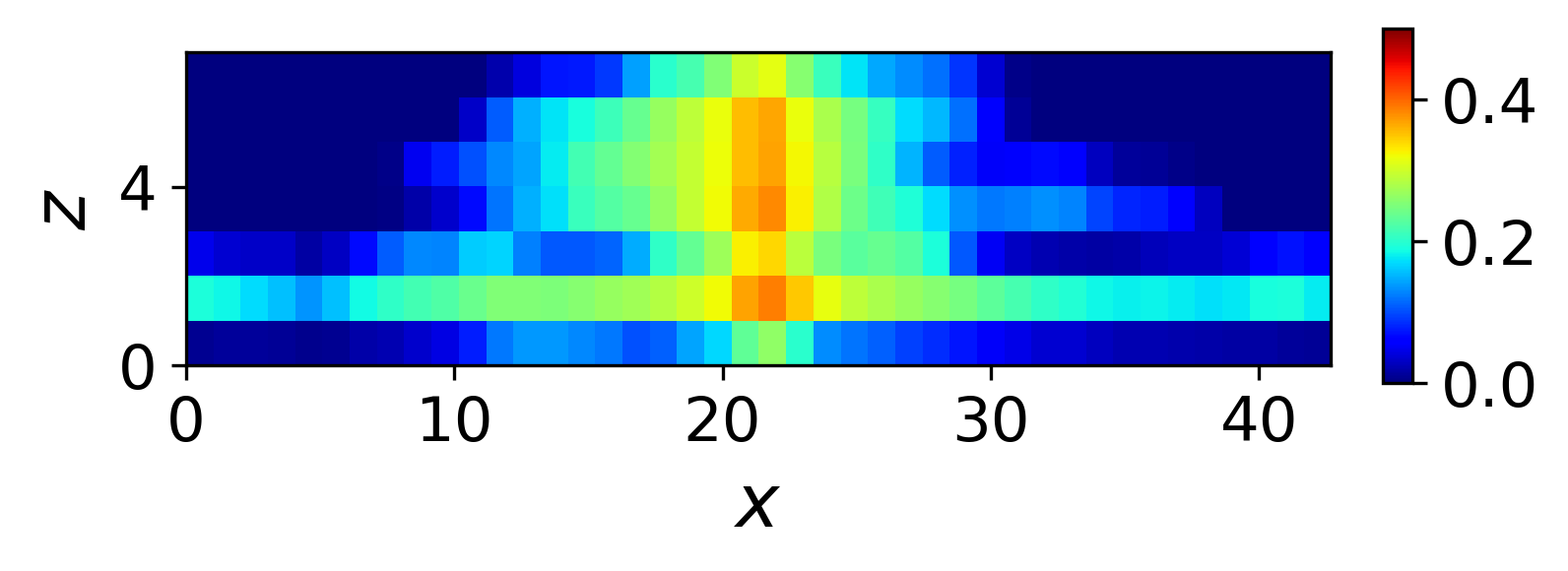}}

\subcaptionbox{Plume location $y$-$z$ cross section}{
\includegraphics[width=0.47\linewidth]{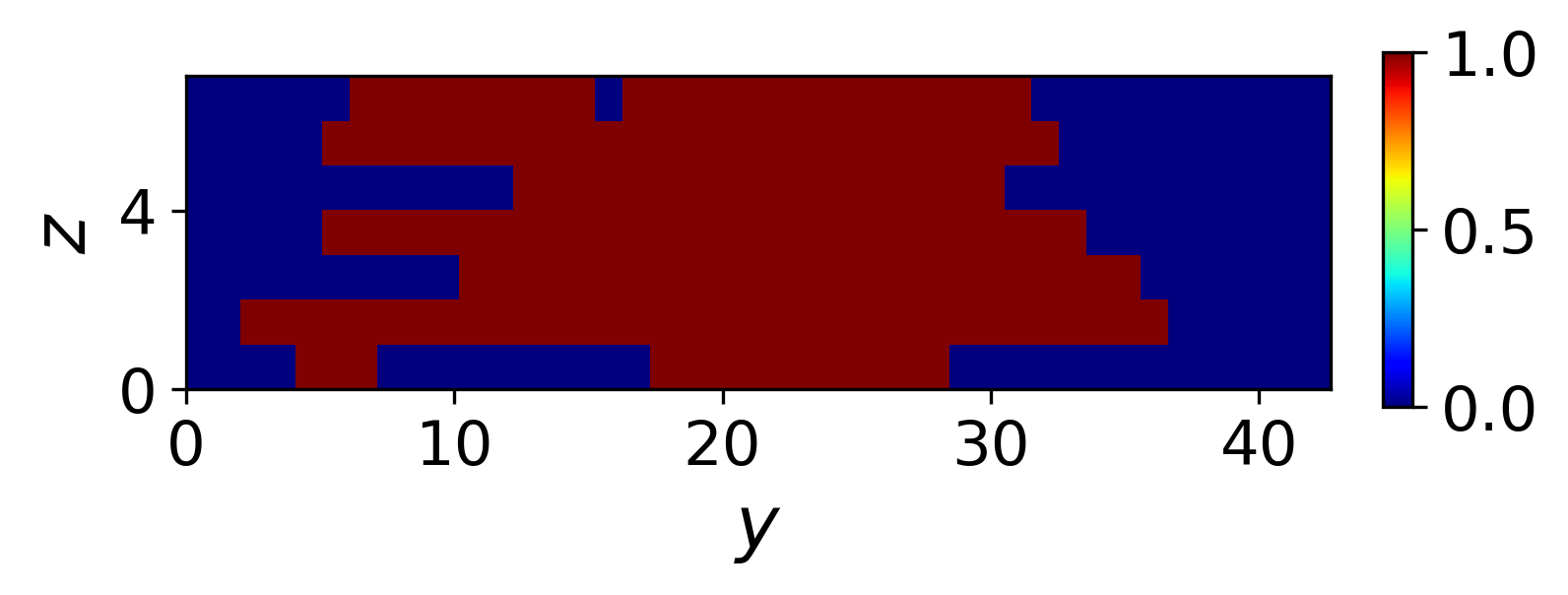}}
\hspace{0.0001\linewidth}
\subcaptionbox{Plume location $x$-$z$ cross section}{
\includegraphics[width=0.47\linewidth]{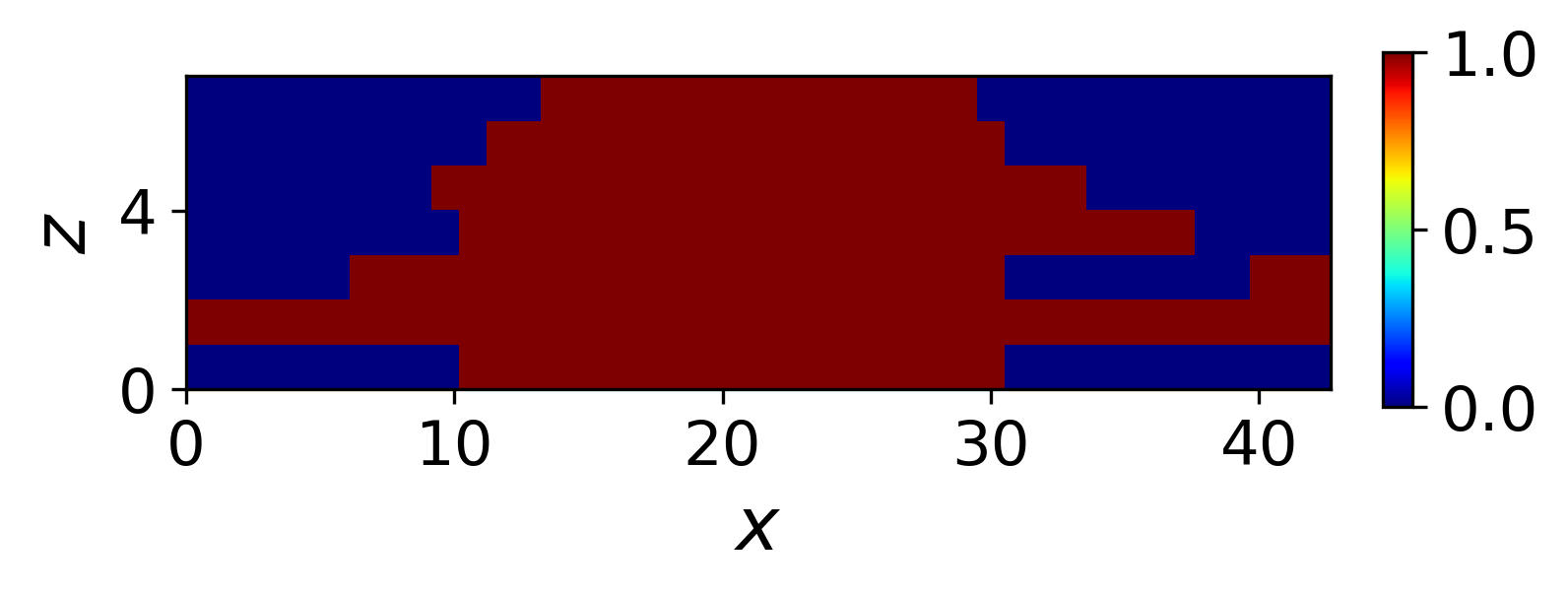}}

\caption{True-model saturation fields at 1~year. Top row shows high-fidelity results for cross sections through the injection well, second row shows corresponding filtered (interpreted saturation seismic) results, and bottom row shows plume location using a saturation threshold of 0.05.}
    \label{fig:plume_true}
\end{figure}

The true (synthetic) geomodel, used to provide observed data, is a realization characterized by the following metaparameter values: $\mu_{\log k}=3.0$, $\sigma_{\log k}=2.3$, $\log_{10}a_r=-1.8$ ($a_r=0.0158$), $d=0.03$ and $e=0.055$. The true-model response for CO$_2$ saturation, along $y$-$z$ and $x$-$z$ cross sections through the injection well at a time of 1~year, is presented in Figure~\ref{fig:plume_true}(a) and (b). The corresponding filtered seismic interpretations are displayed in Figure~\ref{fig:plume_true}(c) and (d). The filtered response loses some of the detail, but the heterogeneity-driven channeling features still appear. The plume locations for this case are presented in Figure~\ref{fig:plume_true}(e) and (f). Note that the channeling is somewhat preserved, though the level of detail is clearly less than that in Figure~\ref{fig:plume_true}(c) and (d).

Some amount of random measurement error (noise) will be present in the observed seismic and monitoring data. To represent this measurement error, as well as other types of error such as model resolution error, we add a noise vector $\boldsymbol\epsilon$ to the true data used in history matching. The observed data vector is thus expressed as:
\begin{center}
\begin{equation}
\setlength{\abovedisplayskip}{-9pt}
\mathbf{d}_\text{obs}=\mathbf{d}_\text{true}+\boldsymbol\epsilon.
\label{eq:obs}
\end{equation}
\end{center}
Here we take $\boldsymbol\epsilon$ to be uncorrelated and unbiased, i.e., of zero mean and diagonal covariance matrix. The measurement error and model resolution error are assigned directly in this work. For the monitoring well data, the standard deviation of the combined measurement and model resolution errors is set to be 5\% of the maximum of the collected monitoring data over all available time steps. In our case this is 0.0162, which is close to the value (0.02) used by \citet{sun2019data}. For the interpreted seismic data, the standard deviation of the error is set to be 10\% of the maximum of the observed data over all available time steps. This is similar to the approach used by \citet{gervais2010integration}, who set the standard deviation of error to be 10\% of the seismic data mean. The noise for the (synthetic) seismic data is added to the `true' high-fidelity simulation results before the filtering procedure, consistent with the approach used by \citet{bukshtynov2015comprehensive}. Other treatments, e.g., error varying as a function of saturation, could be readily incorporated into our framework.

Error in the surrogate models relative to full-physics simulation also enters the history matching formulation. This surrogate-model approximation error, denoted $\boldsymbol\epsilon_\text{surr}$, is estimated during the test-set evaluation. Specifically,
\begin{center}
\begin{equation}
\setlength{\abovedisplayskip}{-9pt}
\boldsymbol\epsilon_\text{surr}=  \mathbf{d}_\text{pred}-\mathbf{d}_\text{true},
\label{eq:surr_error}
\end{equation}
\end{center}
where $\mathbf{d}_\text{pred}$ and $\mathbf{d}_\text{true}$ denote data predicted by the surrogate and true models, respectively. These errors can be computed at each time step used in the history matching. In our case, the mean error (over all time steps) in the monitoring well data surrogate is 0.00124. For the seismic surrogate, the mean error for interpreted saturation seismic and plume location seismic are -0.0001 and -0.0004, respectively. Thus we see that the average errors are near zero, consistent with the assumption of unbiased errors.

The standard deviations ($\sigma_\text{surr}$) of the errors are larger than the mean errors. Specifically, the observed $\sigma_\text{surr}$ values are around 0.03 for monitoring well data, 0.01 for interpreted seismic data, and 0.1 for plume location data. It is these quantities that enter into the history matching computations. The total standard deviation ($\sigma_\text{total}$), which combines the standard deviation of $\boldsymbol\epsilon_\text{surr}$ ($\sigma_\text{surr}$) and the measurement and model resolution errors (we denote the standard deviation of these two errors as $\sigma_\text{d}$), is obtained via $\sigma_\text{total}=\left(\sigma_\text{surr}^2+\sigma_\text{d}^2\right)^{1/2}$. In our history matching computations, the total standard deviation is calculated separately for each history matching time step (based on $\sigma_\text{surr}$ at that time step).

We will apply the hierarchical MCMC method introduced in Section~\ref{sec:mcmc} for three cases -- monitoring well data only, monitoring well data plus interpreted seismic plume location, and monitoring well data plus interpreted seismic saturation. This will allow us to assess the impact of different types and combinations of data on uncertainty reduction. In cases with seismic data, the likelihood appearing in Eq.~\ref{eq:pca_ap} and Eq.~\ref{eq:meta_ap} is given by

\begin{center}
\begin{equation}
\setlength{\abovedisplayskip}{-7pt}
\begin{split}
p( \mathbf{d}_\text{obs} \mid\boldsymbol \xi,\mathbf{h})=  c\exp\left(\frac{1}{T_\text{mon}}\left(-\frac{1}{2}(\mathbf{d}_\text{obs}^\text{mon}-f_\text{mon}(\boldsymbol \xi,\mathbf{h}))^T C_\text{mon}^{-1}(\mathbf{d}_\text{obs}^\text{mon}-f_\text{mon}(\boldsymbol \xi,\mathbf{h}))\right.\right)\\
+\frac{1}{T_\text{seis}}\left(\left.-\frac{1}{2}(\mathbf{d}_\text{obs}^\text{seis}-f_\text{seis}(\boldsymbol \xi,\mathbf{h}))^T C_\text{seis}^{-1}(\mathbf{d}_\text{obs}^\text{seis}-f_\text{seis}(\boldsymbol \xi,\mathbf{h}))\right)\right),
\label{eq:likelihood}
\end{split}
\end{equation}
\end{center}
where $c$ is a normalization constant, $\mathbf{d}_\text{obs}^\text{mon}$ and $\mathbf{d}_\text{obs}^\text{seis}$ denote observed monitoring well data and interpreted seismic data, respectively, and $C_\text{mon}$ and $C_\text{seis}$ are the total error covariance matrices for monitoring well and seismic data, respectively. The parameters $T_\text{mon}$ and $T_\text{seis}$, referred to as simulated temperatures, control the impact of the different data types and the MCMC acceptance rate~\citep{li2012momcmc}. If only monitoring data are used for history matching, the seismic data portion of the likelihood does not appear, resulting in
\begin{center}
\begin{equation}
\setlength{\abovedisplayskip}{-7pt}
\begin{split}
p( \mathbf{d}_\text{obs} \mid\boldsymbol \xi,\mathbf{h})=  c\exp\left(\frac{1}{T_\text{mon}}\left(-\frac{1}{2}(\mathbf{d}_\text{obs}^\text{mon}-f_\text{mon}(\boldsymbol \xi,\mathbf{h}))^T C_\text{mon}^{-1}(\mathbf{d}_\text{obs}^\text{mon}-f_\text{mon}(\boldsymbol \xi,\mathbf{h}))\right)\right).
\label{eq:likelihood_mon}
\end{split}
\end{equation}
\end{center}

Based on numerical experimentation, we specify $T_\text{mon}=0.8$ when only monitoring data are used, $T_\text{mon}=1.5$ and $T_\text{seis}=200$ when monitoring well data and plume location seismic data are used, and $T_\text{mon}=1$ and $T_\text{seis}=6$ when monitoring well data and interpreted saturation seismic data are used. These settings result in MCMC acceptance rates in the appropriate range, i.e., 10\% to 40\% \citep{gelman1996efficient,han2024surrogate}, and balance the impact of the different terms in the likelihood function.

Additional parameter specifications are as follows. We set $\beta=0.05$, where $\beta$ (appearing in Eq.~\ref{eq:pca_sample}) scales the magnitude of sample updates. The standard deviation of the Gaussian proposal distribution for the metaparameters is set to be 0.05 of the parameter range. In the results below, hierarchical MCMC using only monitoring well data requires 57,819 function evaluations (17,750 sets of metaparameters and their associated PCA latent variables are accepted). Results using monitoring well data and plume location seismic data require 83,353 function evaluations (17,250 sets of metaparameters and associated PCA variables accepted). Results using monitoring well data and saturation seismic data require 61,592 function evaluations (15,000 sets of metaparameters and associated PCA variables accepted). 

Recall that, to train the surrogate models, we required 4000 simulation runs and about 2.25~hours of GPU-based training. Once these tasks are completed, the MCMC history matching for each case (involving $\sim$58,000 or more function evaluations plus associated geomodel construction computations) can be completed relatively quickly, in about 9~hours. Without these deep learning-based surrogate models, the history matching framework applied in this study would be intractable.

Prior to performing history matching, it is useful to conduct a sensitivity analysis to quantify the impact of the various parameters on the data types used in data assimilation. We performed such a sensitivity analysis, of the type presented by \citet{han2024surrogate}, for our case. The findings are that, with either monitoring well or seismic saturation data, the most sensitive parameters are the PCA variables, $\boldsymbol \xi$. Monitoring well data are then most sensitive to $\mu_{\log k}$, $\sigma_{\log k}$ and $a_r$; very little sensitivity to $d$ and $e$ is observed. Seismic data show some sensitivity to $\mu_{\log k}$ and $\sigma_{\log k}$, but they are less sensitive to the other parameters. Thus, we expect the data to be informative in terms of $\boldsymbol \xi$ (i.e., particular realizations) and somewhat informative for the metaparameters other than $d$ and $e$. These general expectations will be seen to be consistent with our history matching results.

\subsection{History matching results}
\label{sec:HM_results}
For the history matching, we use monitoring well data at times of 24, 49, 73, 97 and 122~days and, in the case when it is considered, interpreted seismic data at 2 and 4~months. Results for later times are used to evaluate the predictions of the history matched models. Recall that the simulations proceed until a final time of 1~year.

\begin{figure}[htbp] 
\centering 
\vspace{0.35cm} 
\setlength{\lineskip}{\medskipamount}
\subcaptionbox{$\mu_{\log k}$ without seismic}{
\includegraphics[height=0.275\linewidth]{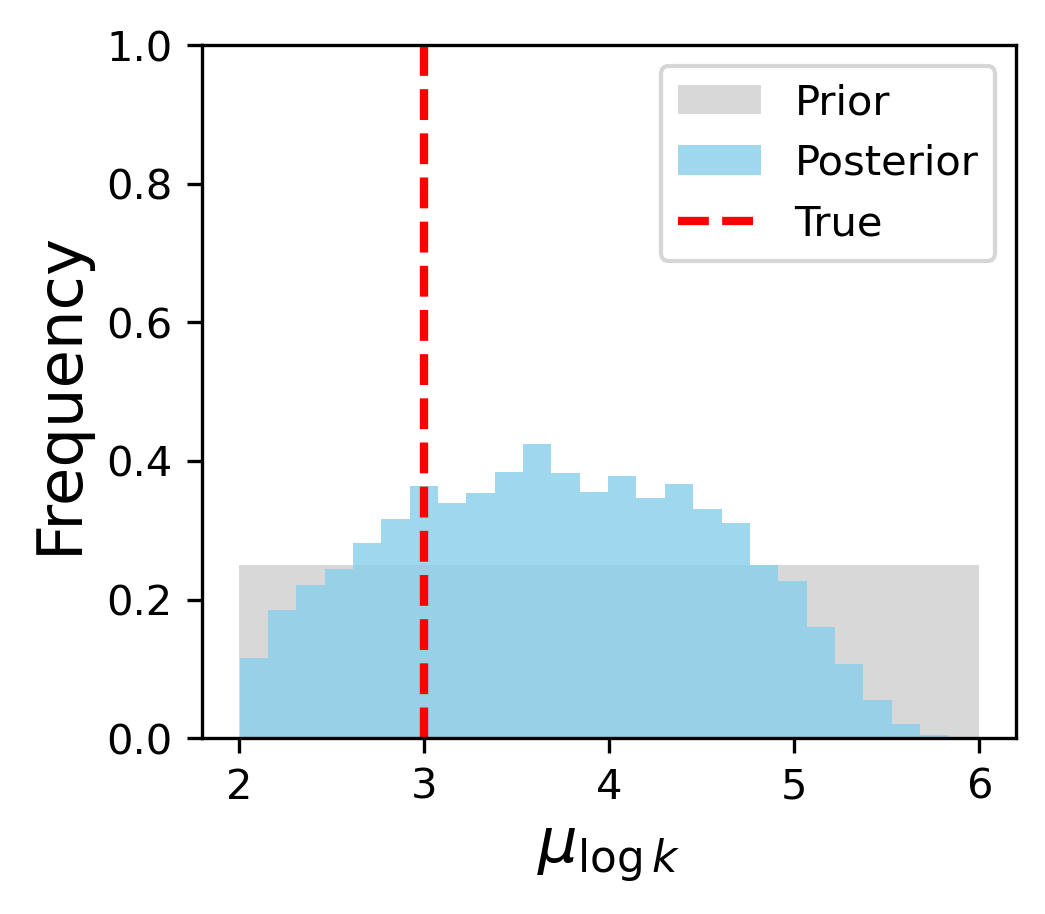}}
\subcaptionbox{$\mu_{\log k}$ with plume location}{
\includegraphics[height=0.275\linewidth]{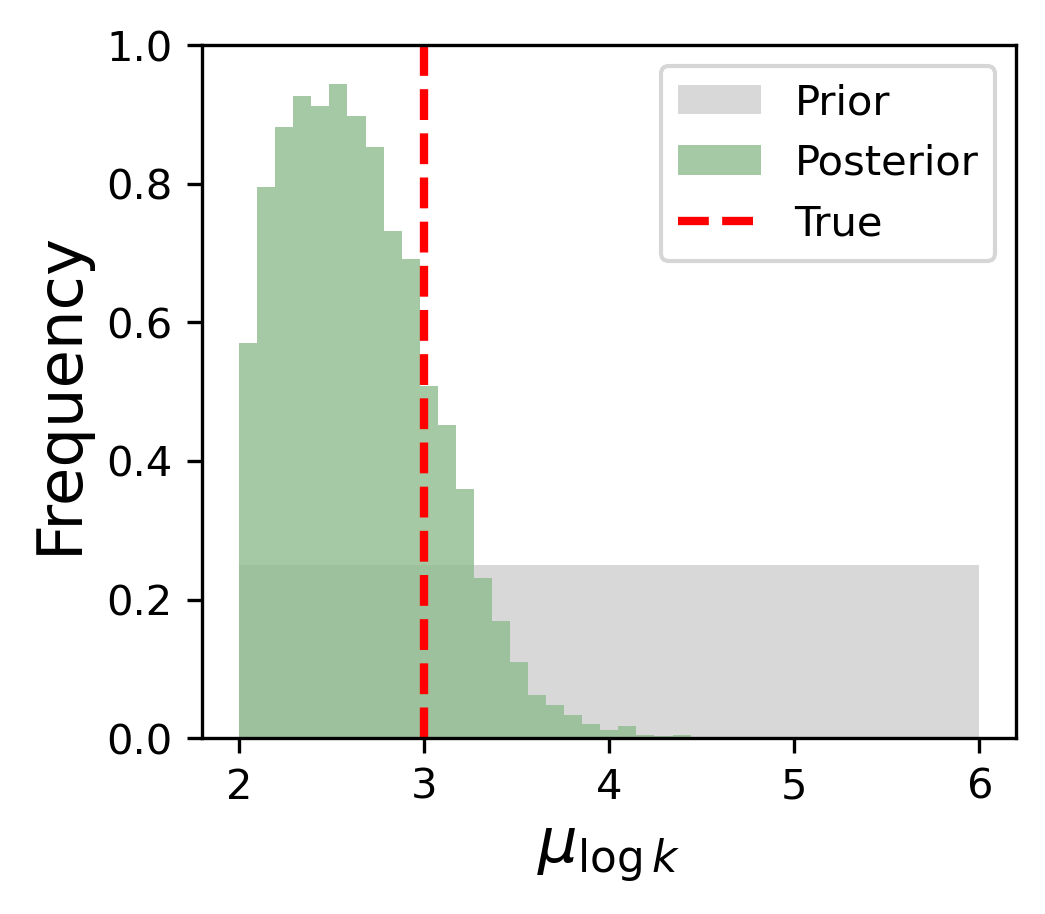}}
\subcaptionbox{$\mu_{\log k}$ with interpreted $S_g$}{
\includegraphics[height=0.275\linewidth]{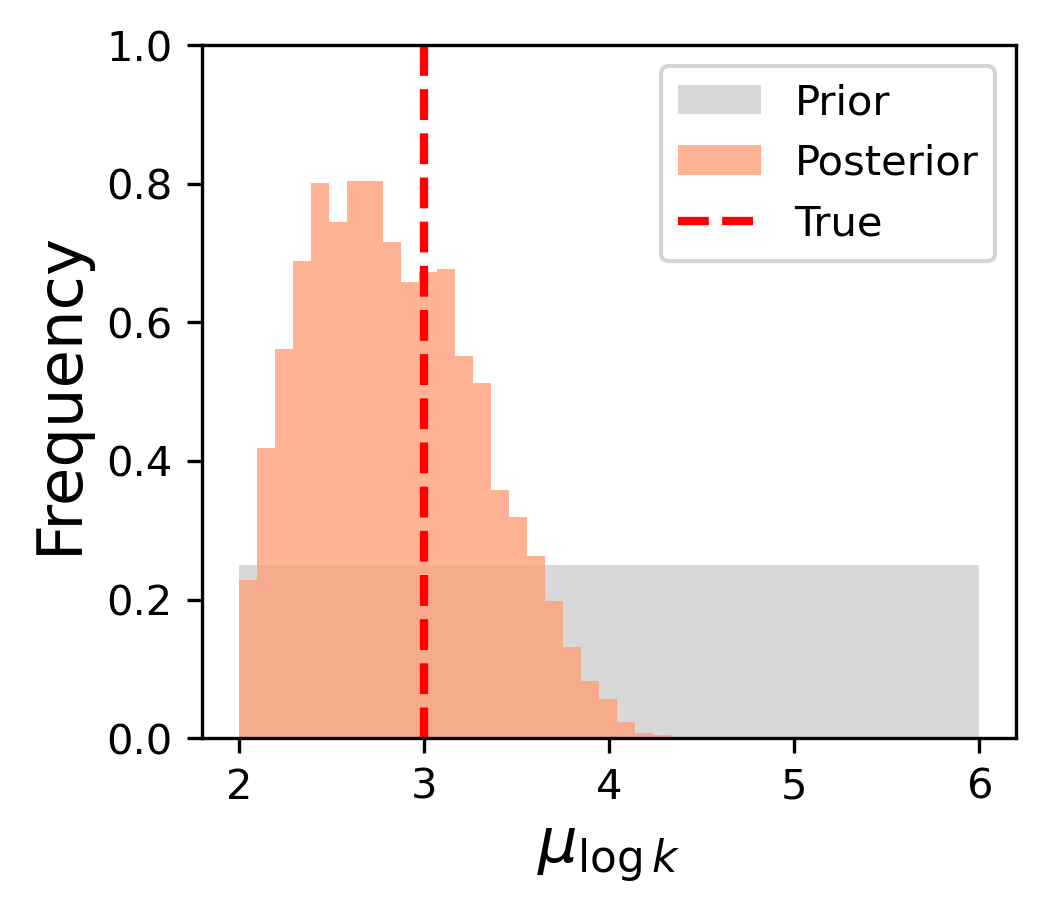}}

\subcaptionbox{$\sigma_{\log k}$ without seismic}{
\includegraphics[height=0.272\linewidth]{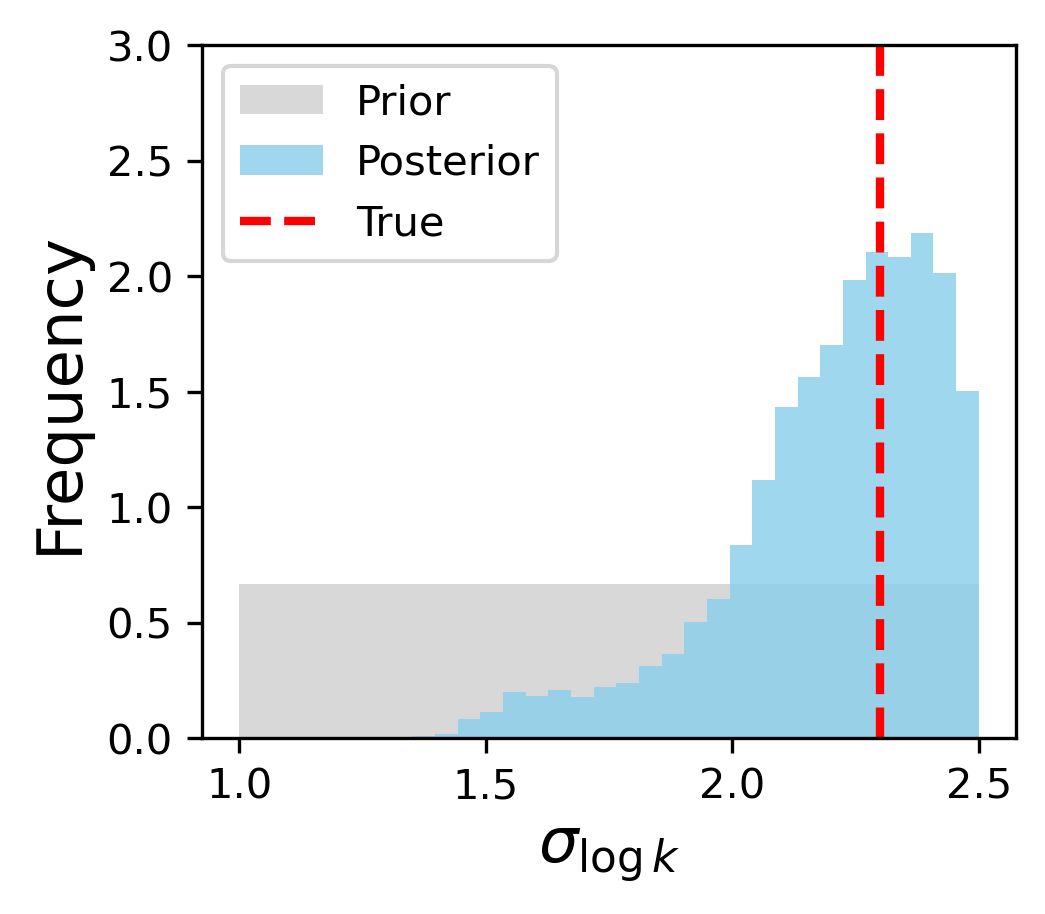}}
\subcaptionbox{$\sigma_{\log k}$ with plume location}{
\includegraphics[height=0.272\linewidth]{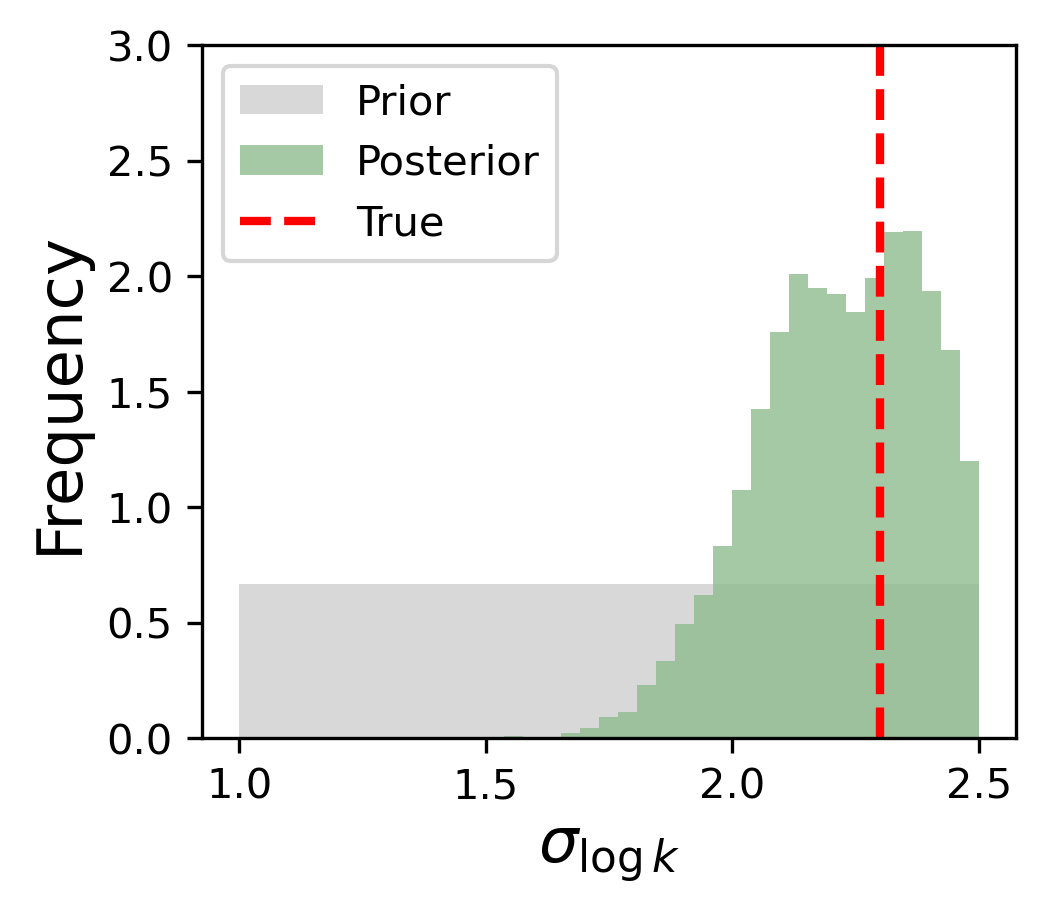}}
\subcaptionbox{$\sigma_{\log k}$ with interpreted $S_g$}{
\includegraphics[height=0.272\linewidth]{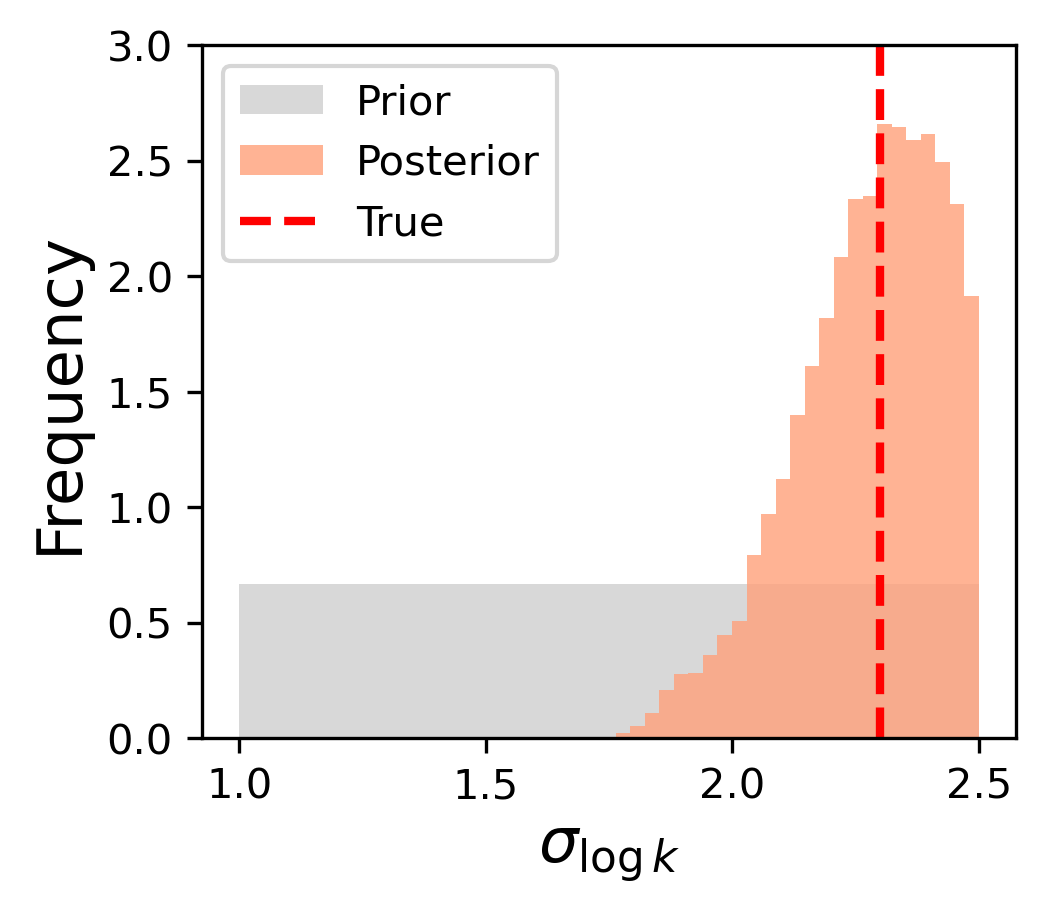}}

\subcaptionbox{$\log_{10}a_r$ without seismic}{
\includegraphics[height=0.275\linewidth]{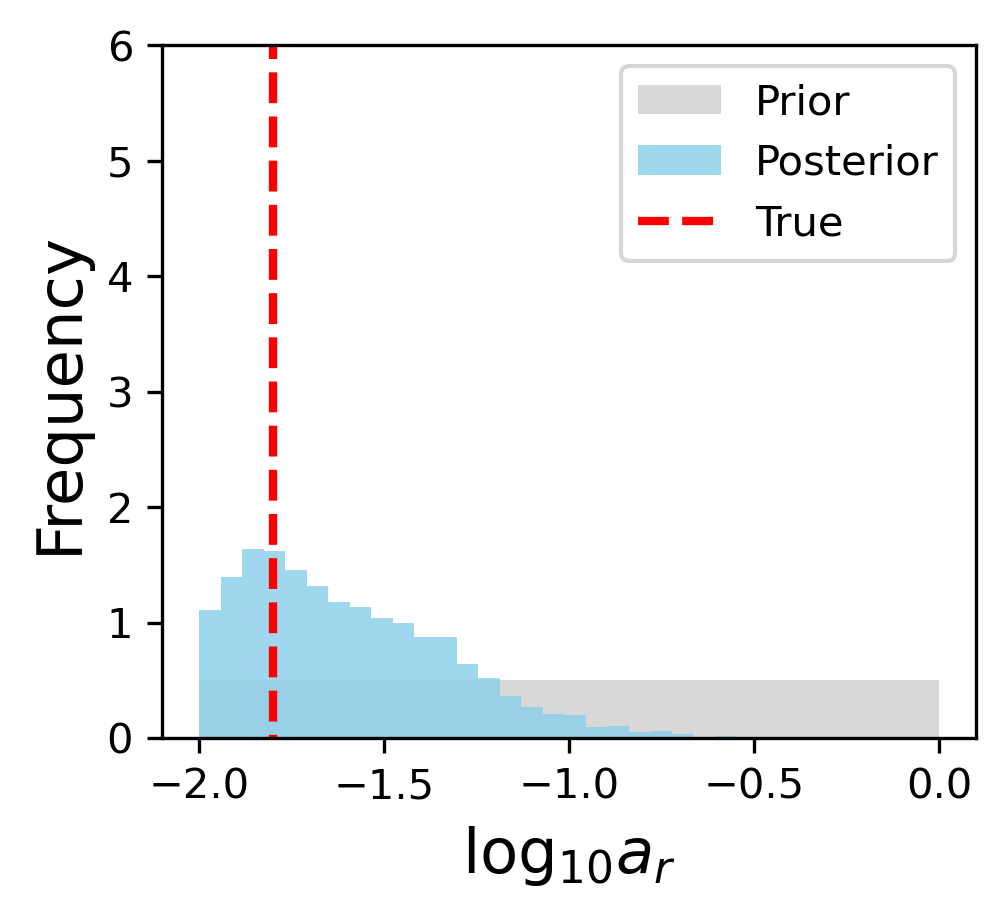}}
\subcaptionbox{$\log_{10}a_r$ with plume location}{
\includegraphics[height=0.275\linewidth]{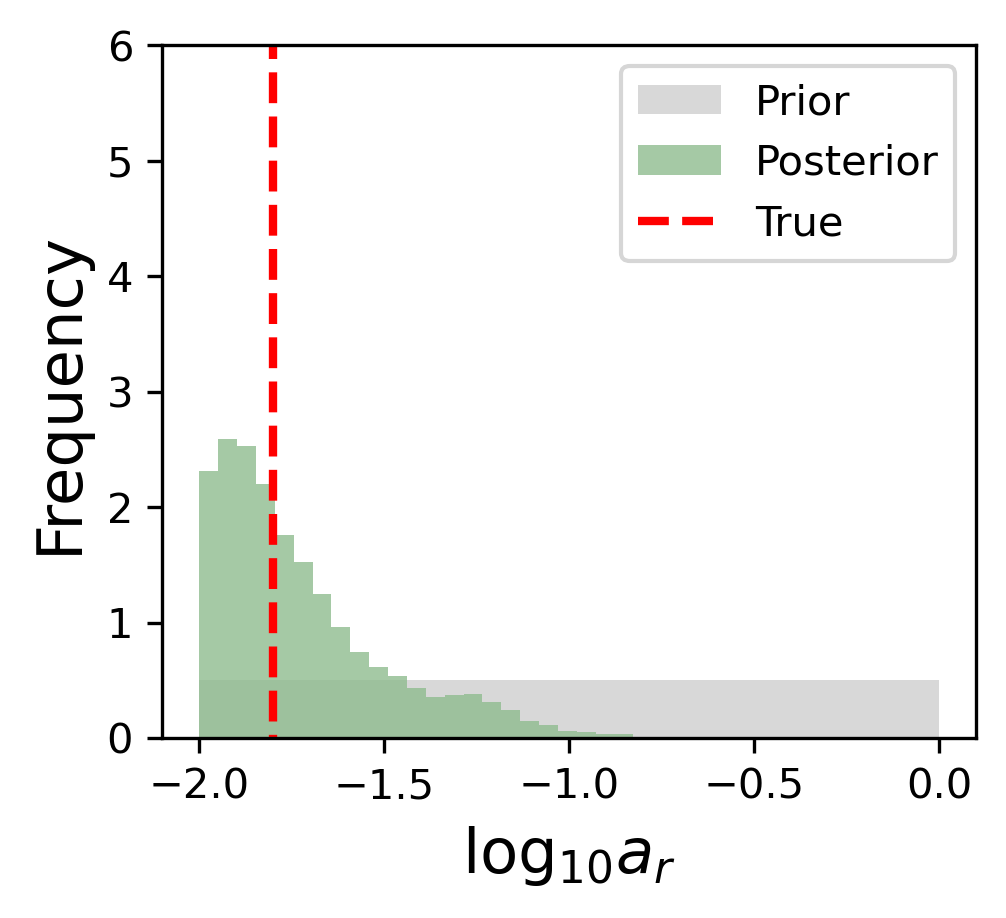}}
\subcaptionbox{$\log_{10}a_r$ with interpreted $S_g$}{
\includegraphics[height=0.275\linewidth]{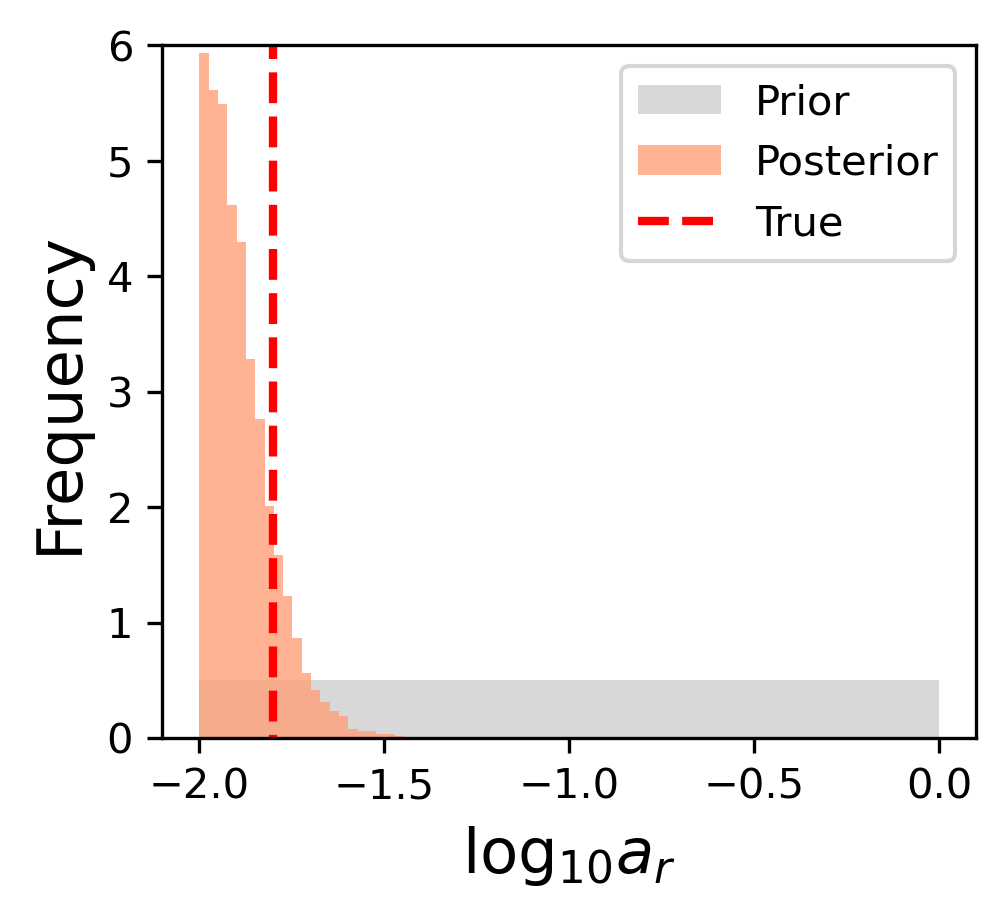}}
\caption{History matching results for metaparameters \texorpdfstring{$\mu_{\log k}$}{}, \texorpdfstring{$\sigma_{\log k}$}{} and \texorpdfstring{$\log_{10}a_r$}{} with monitoring well data only (left column, blue histograms), with monitoring well data and interpreted plume location seismic data (middle column, green histograms), and with monitoring well data and interpreted saturation seismic  data (right column, orange histograms). Priors shown in gray and true values indicated by the vertical red line.}
\label{fig:inversion_comparison1}
\end{figure}

\begin{figure}[htbp] 
\centering 
\vspace{0.35cm} 
\setlength{\lineskip}{\medskipamount}
\subcaptionbox{$d$ without seismic}{
\includegraphics[height=0.271\linewidth]{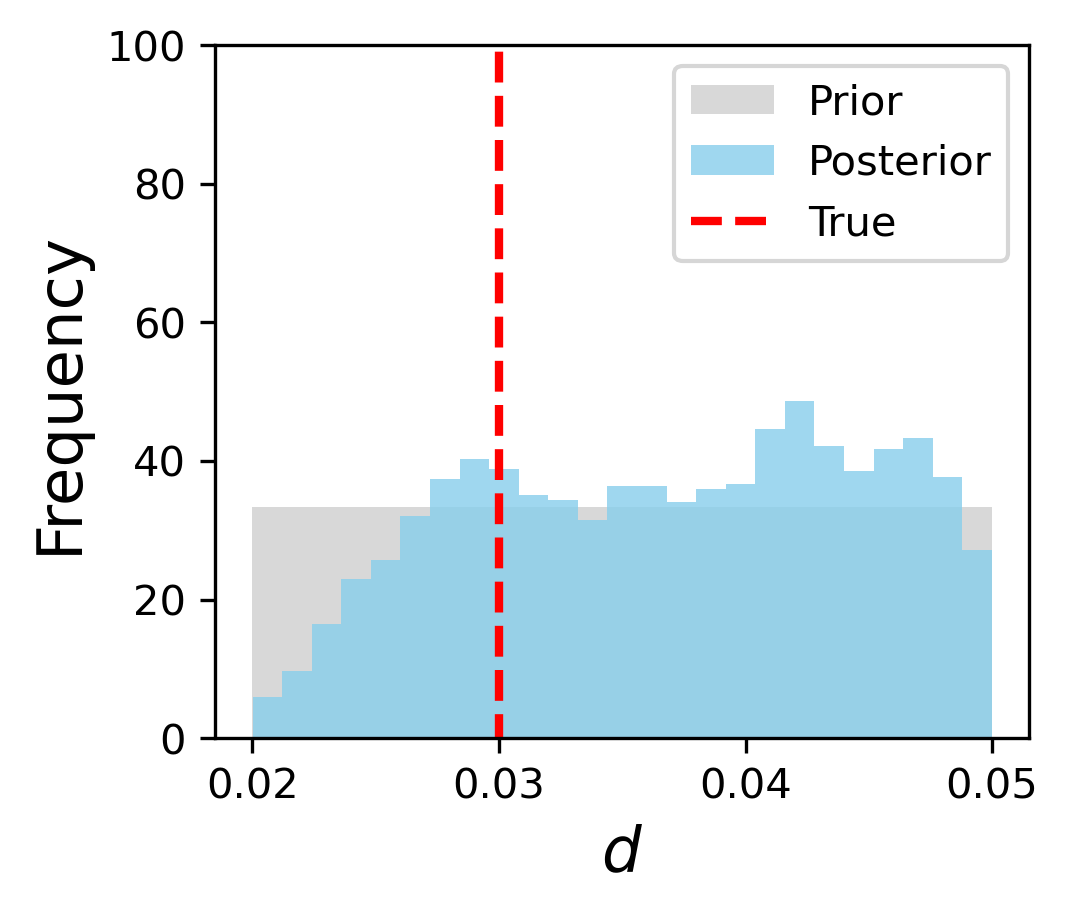}}
\subcaptionbox{$d$ with plume location}{
\includegraphics[height=0.271\linewidth]{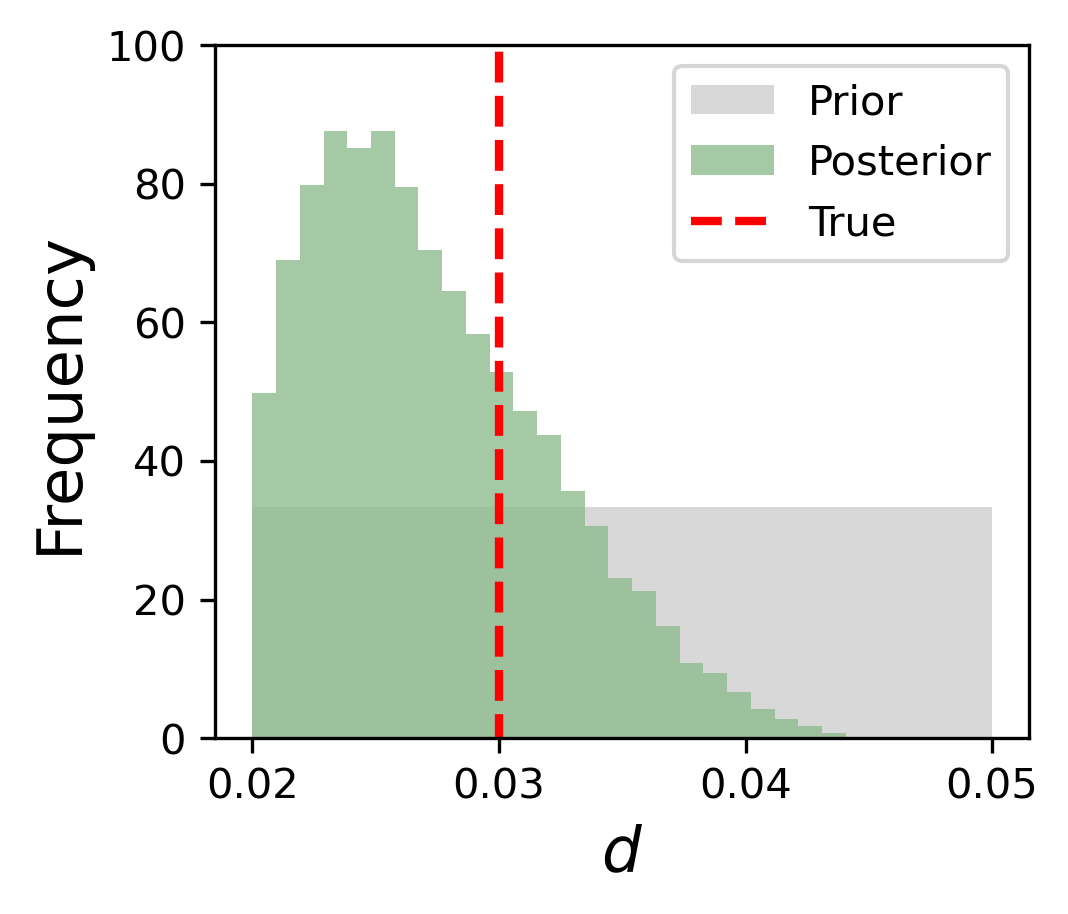}}
\subcaptionbox{$d$ with interpreted $S_g$}{
\includegraphics[height=0.271\linewidth]{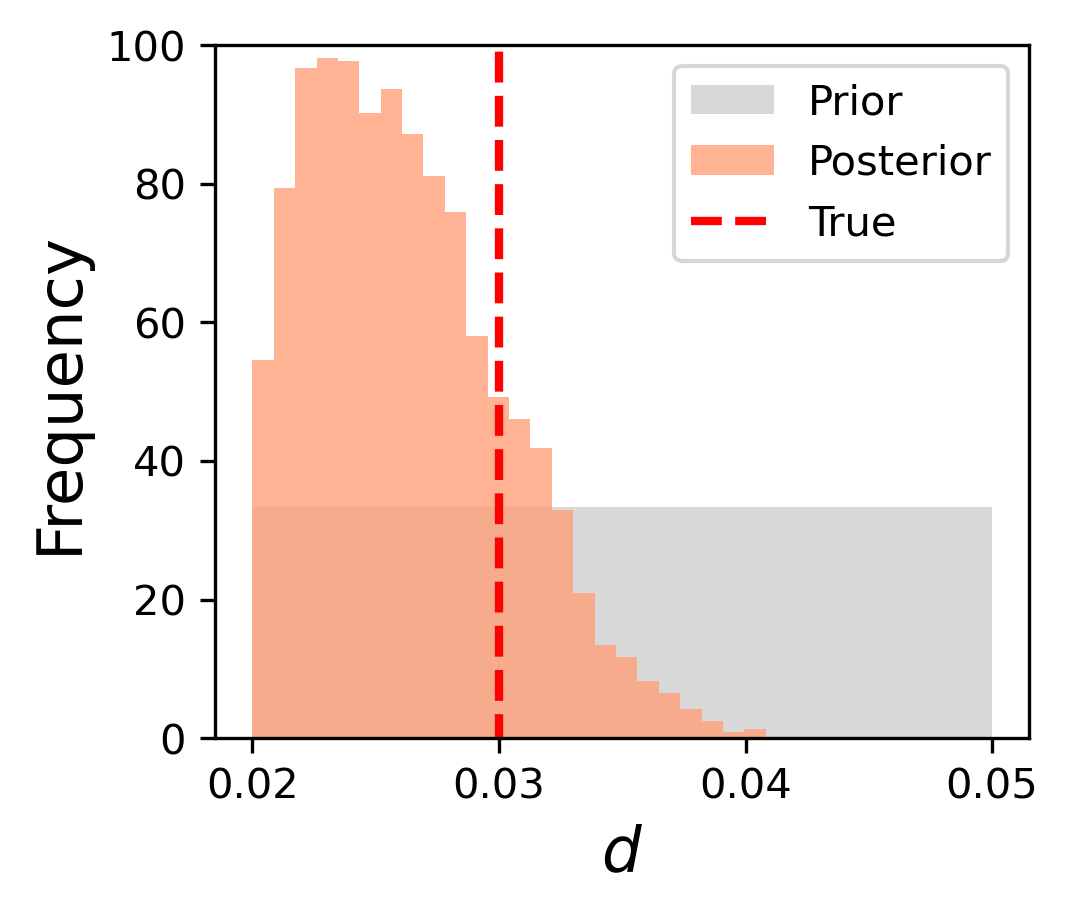}}

\subcaptionbox{$e$ without seismic}{
\includegraphics[height=0.275\linewidth]{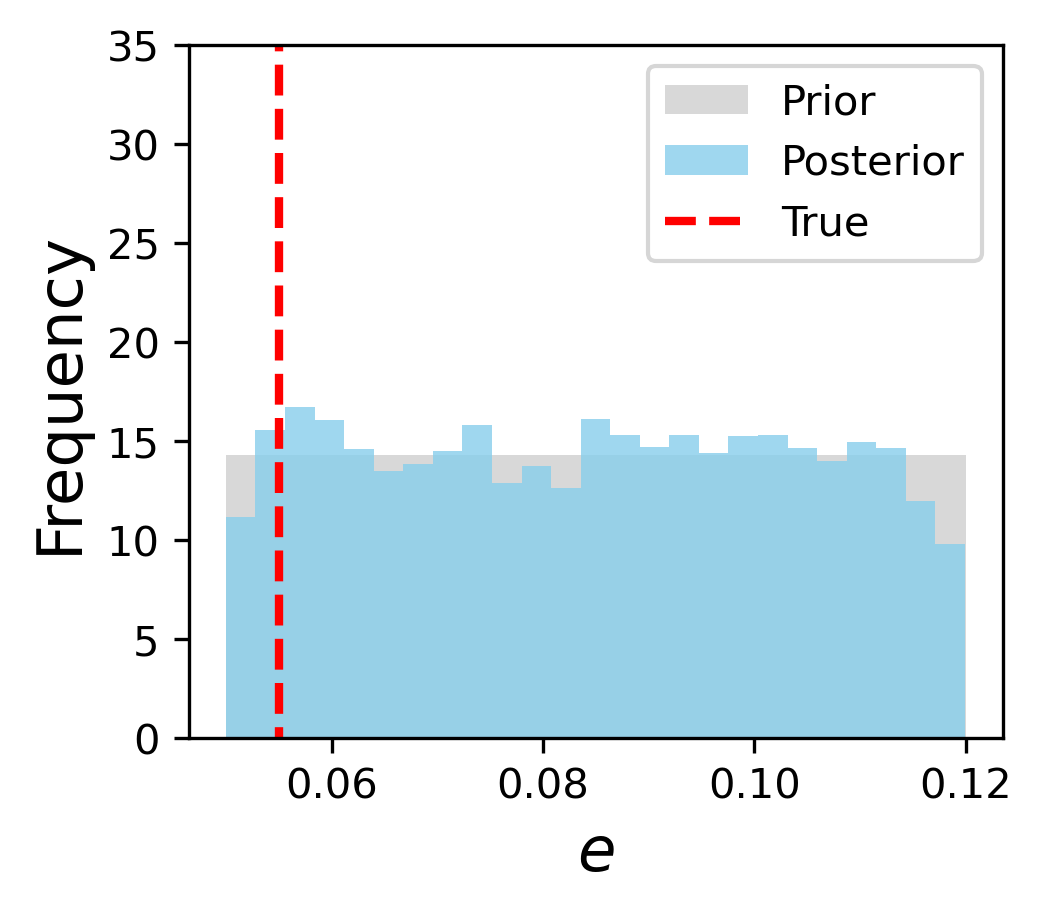}}
\subcaptionbox{$e$ with plume location}{
\includegraphics[height=0.275\linewidth]{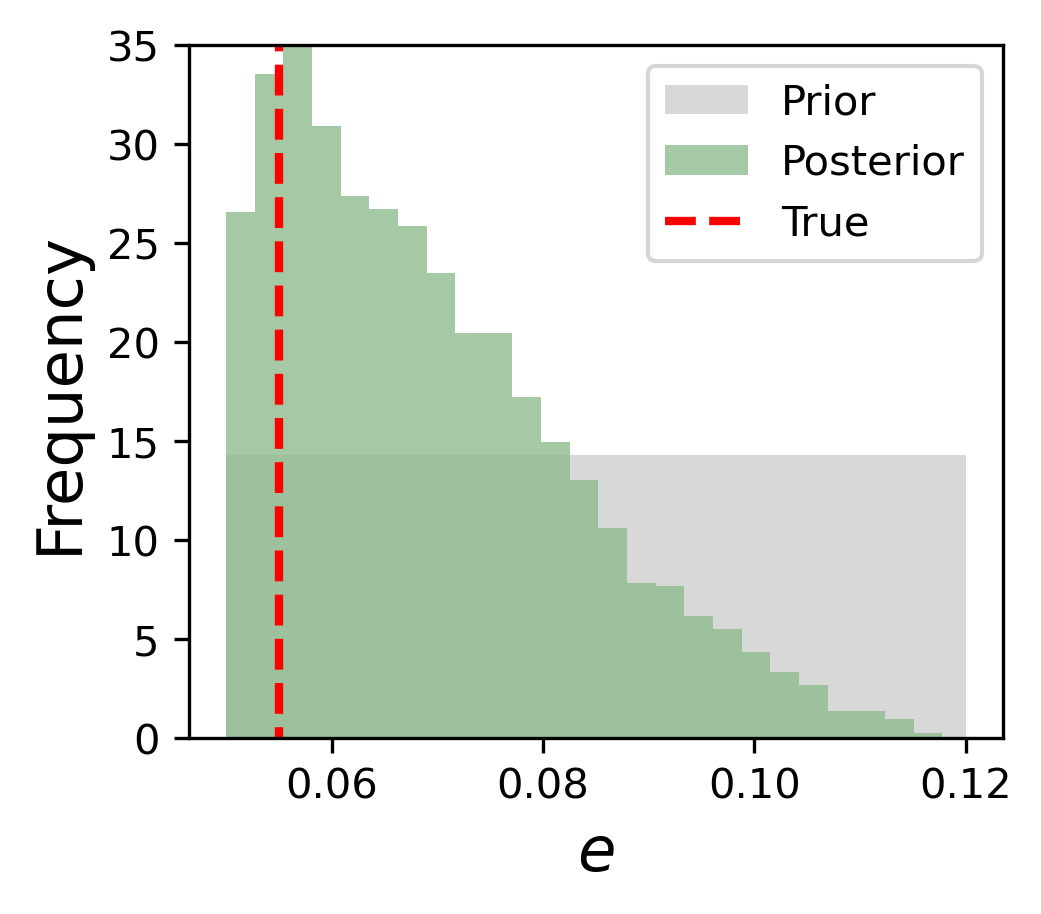}}
\subcaptionbox{$e$ with interpreted $S_g$}{
\includegraphics[height=0.275\linewidth]{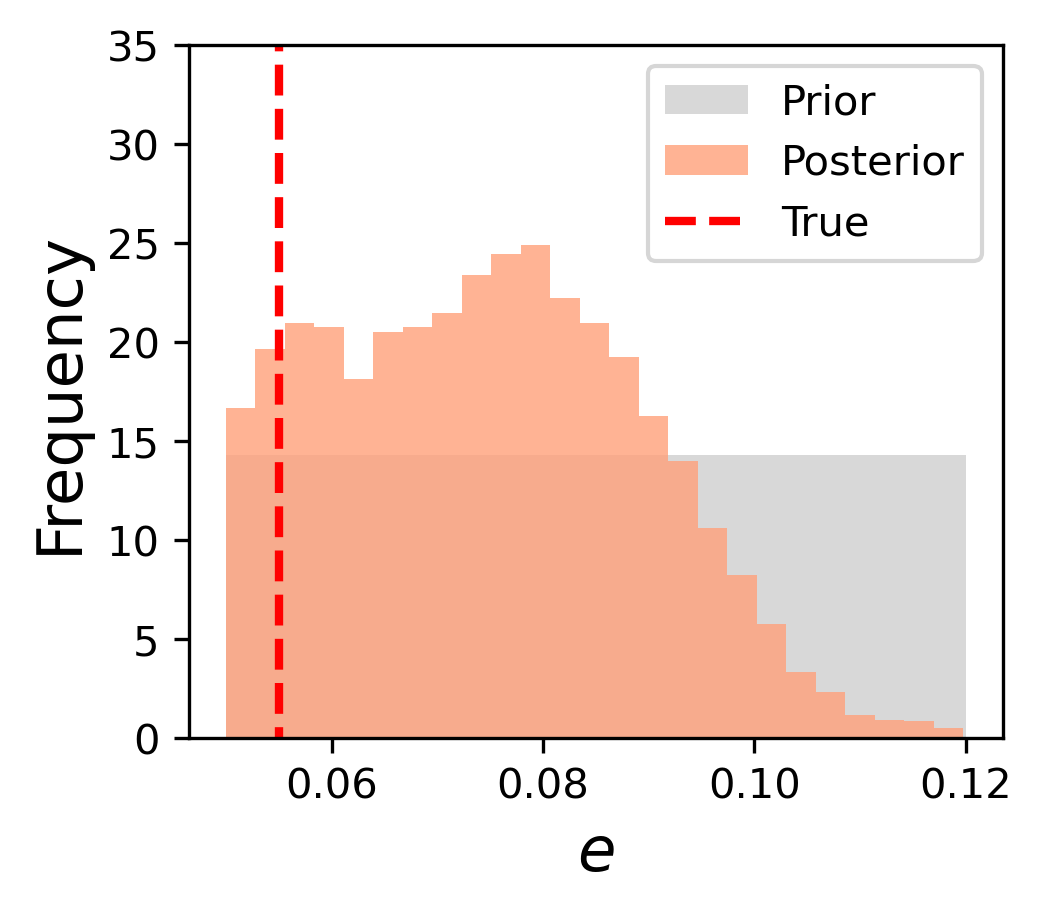}}

\subcaptionbox{$\mu_\phi$ without seismic}{
\includegraphics[height=0.275\linewidth]{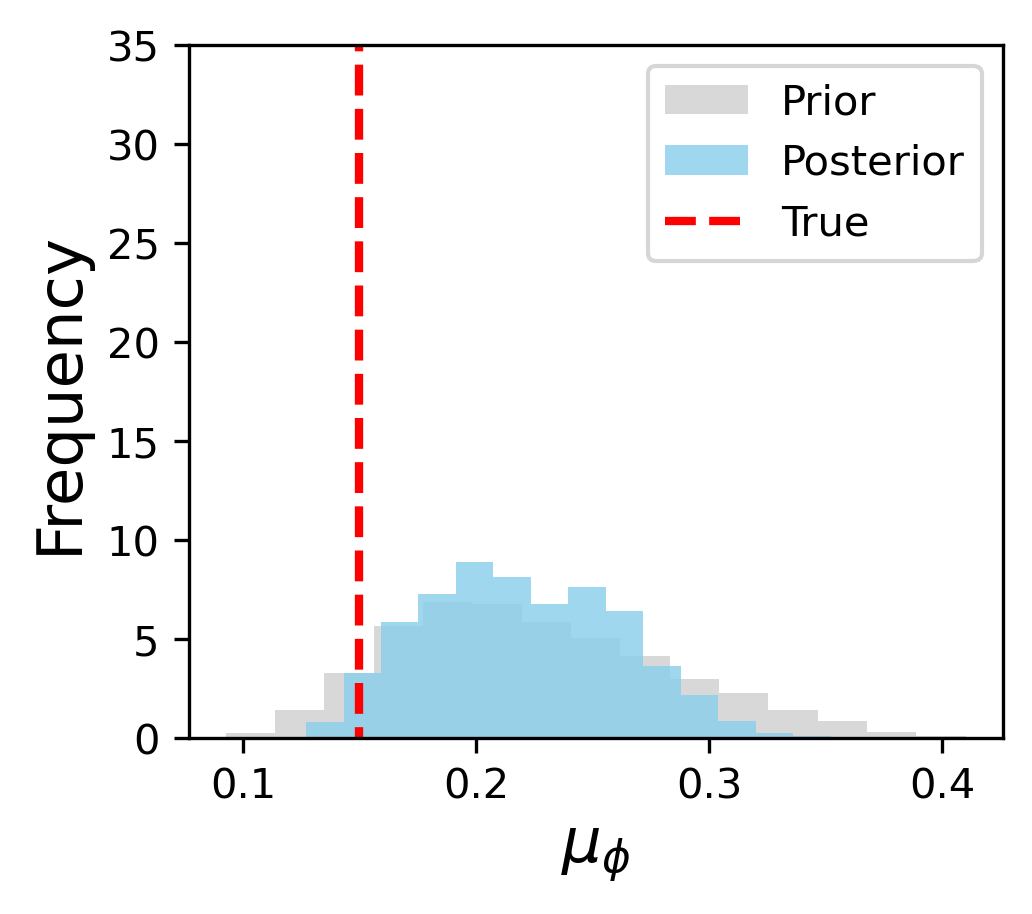}}
\subcaptionbox{$\mu_\phi$ with plume location}{
\includegraphics[height=0.275\linewidth]{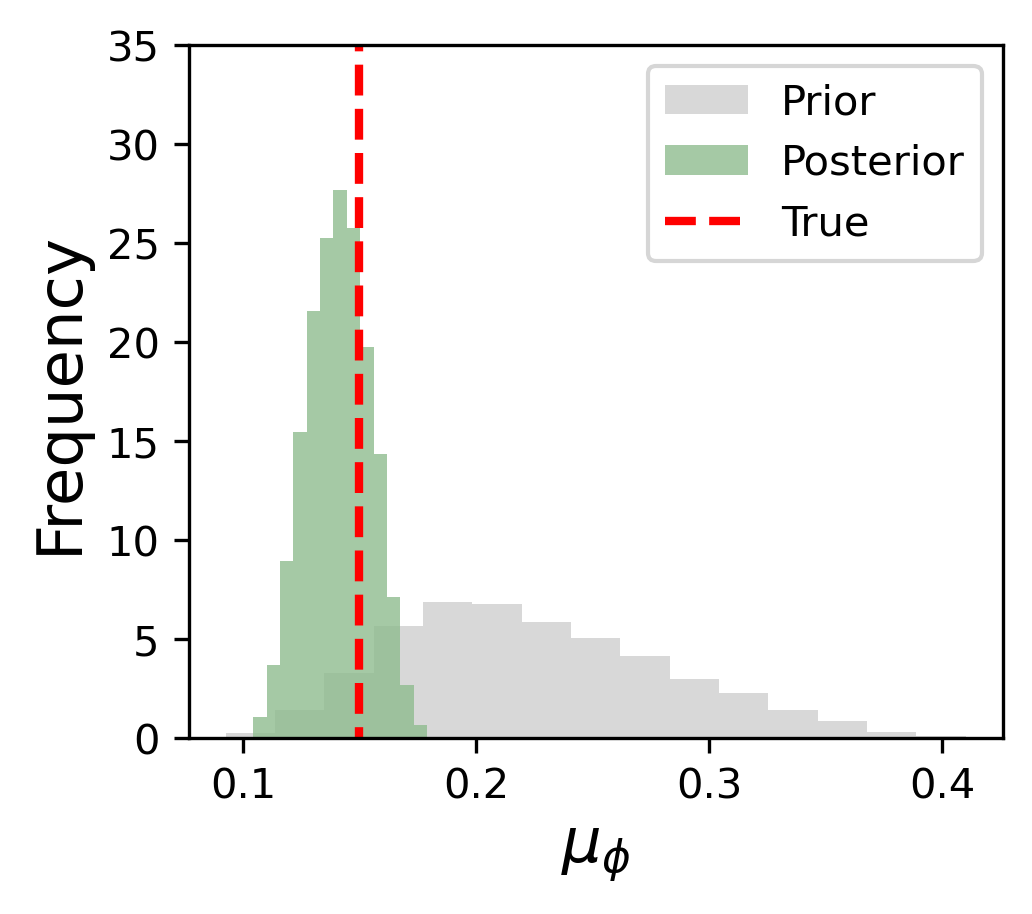}}
\subcaptionbox{$\mu_\phi$ with interpreted $S_g$}{
\includegraphics[height=0.275\linewidth]{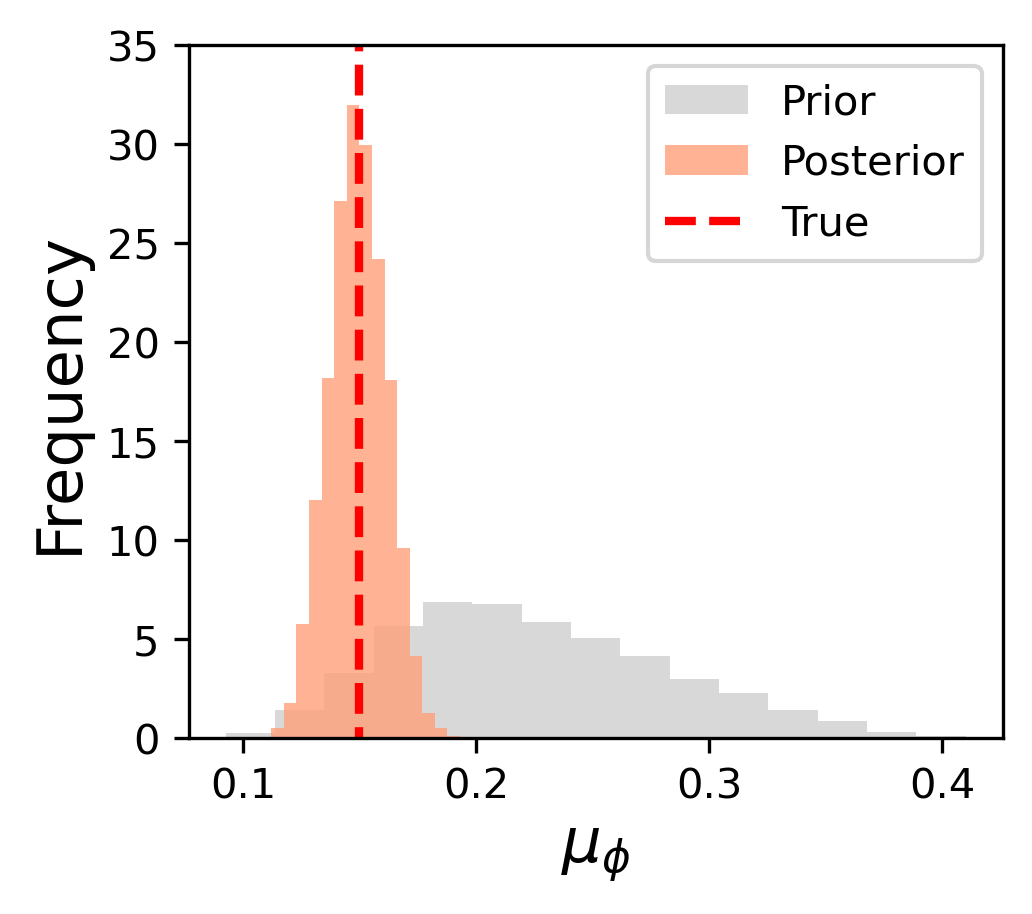}}
\caption{History matching results for metaparameters \texorpdfstring{$d$}{} and \texorpdfstring{$e$}{} and average porosity \texorpdfstring{$\mu_\phi$}{} with monitoring well data only (left column, blue histograms), with monitoring well data and interpreted plume location seismic data (middle column, green histograms), and with monitoring well data and interpreted saturation seismic data (right column, orange histograms). Priors shown in gray and true values indicated by the vertical red line.}
\label{fig:inversion_comparison2}
\end{figure}

Posterior results for the metaparameters, obtained from the last 10,000 accepted samples, for cases both with and without seismic data, are presented in Figures~\ref{fig:inversion_comparison1} and \ref{fig:inversion_comparison2}. The gray region in each figure represents the prior distribution of the metaparameter and the red dashed line indicates the true value of the metaparameter. The posterior histograms appear in blue for results in which only monitoring well data are used, in green for results where monitoring well data and interpreted plume location data are used, and in orange for results where monitoring well data and interpreted seismic saturation data are used. A reasonable degree of uncertainty reduction is achieved for $\sigma_{\log k}$ and $\log_{10}a_r$ when only monitoring well data are used (Figure~\ref{fig:inversion_comparison1}(d) and (g)). In addition, it is evident that the mode of the posterior histogram is near the true value for these quantities. We observe a small amount of uncertainty reduction for $\sigma_{\log k}$ (Figure~\ref{fig:inversion_comparison1}(a)) in this case.

Very little uncertainty reduction is observed for parameters $d$ and $e$ in this case (Figure~\ref{fig:inversion_comparison2}(a) and (d)), indicating that the monitoring well data are not informative in terms of the relationship between $\phi$ and $\log k$. This is consistent with the observations of \citet{han2024surrogate}. A small degree of uncertainty reduction is achieved, however, in the porosity field itself. This is demonstrated in Figure~\ref{fig:inversion_comparison2}(g), where we show the prior and posterior distributions for the mean of the porosity, $\mu_\phi$. The results in Figure~\ref{fig:inversion_comparison2}(g) are determined by using the relationship between $\phi$ and $\log k$ in Eq.~\ref{eq:phi_logk}.

Considerably more uncertainty reduction in the key metaparameters is achieved when both monitoring well data and seismic data are used. This is the case using either interpreted plume location or seismic saturation data. Specifically, with both monitoring well data and plume location data, we observe narrower distributions, which include the true value, for $\mu_{\log k}$, $\sigma_{\log k}$ and $\log_{10}a_r$ (Figure~\ref{fig:inversion_comparison1}(b), (e), (h)). The posterior distributions for $\sigma_{\log k}$ and $\log_{10}a_r$ are narrower, and include the true value, when monitoring well data and interpreted saturation seismic data are used (Figure~\ref{fig:inversion_comparison1}(f) and (i)). Interestingly, for $\mu_{\log k}$ there is  more uncertainty reduction using plume location data, though the distribution is more centered around the true value for the case with seismic saturation data. In Figure~\ref{fig:inversion_comparison1}(i), the maximum frequency is observed at the edge of the allowable range for the metaparameter. This (slightly) anomalous behavior may be due to correlated errors or to the small biases that exist in the surrogate models. In any event, the true value of $a_r$ is 0.0158, and the minimum allowable value is 0.01, so these values are close.

There is also more uncertainty reduction in $d$ and $e$ in the cases with seismic data (Figure~\ref{fig:inversion_comparison2}(b), (c), (e), and (f)), though significant posterior uncertainty in these quantities remains. As is apparent in Figure~\ref{fig:inversion_comparison2}(h), the use of both monitoring well data and plume location data provides substantial uncertainty reduction in mean porosity itself. A slightly narrower histogram is observed with interpreted saturation data (compare Figure~\ref{fig:inversion_comparison2}(h) and (i). Taken in total, the results in Figures~\ref{fig:inversion_comparison1} and \ref{fig:inversion_comparison2} clearly show the additional (significant) uncertainty reduction achieved when interpreted seismic data are incorporated into the data assimilation procedure, even if these data provide only plume location rather than interpreted saturation.

We next present saturation fields (CO$_2$ plumes) for prior and posterior geomodels. In order to identify `representative' interpreted seismic saturation fields, and thus enable meaningful comparisons, we proceed as follows. We randomly select 1000 prior geomodel realizations from the training set, 1000 accepted realizations using only monitoring well data, 1000 accepted realizations using monitoring well data and plume location seismic data, and 1000 accepted realizations using monitoring well data and saturation seismic data. Interpreted seismic data are then generated for each of these 4000 realizations using the seismic surrogate model. Then, for each set of 1000 models, we apply k-means clustering to construct four clusters. The medoid from each cluster, identified with a k-medoids method, is then viewed as a representative interpreted seismic saturation field.

\begin{figure}[htbp] 
\centering 
\vspace{0.35cm} 
\setlength{\lineskip}{\medskipamount}
\subcaptionbox{Prior 1}{
\includegraphics[height=0.1\linewidth]{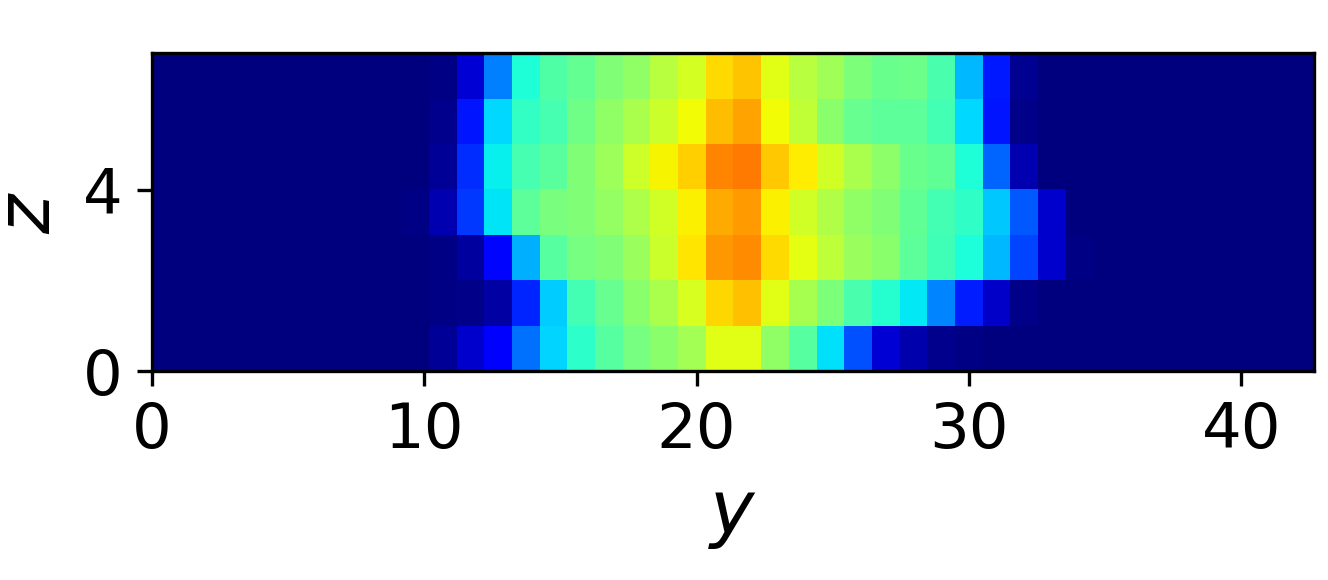}}
\subcaptionbox{Prior 2}{
\includegraphics[height=0.1\linewidth]{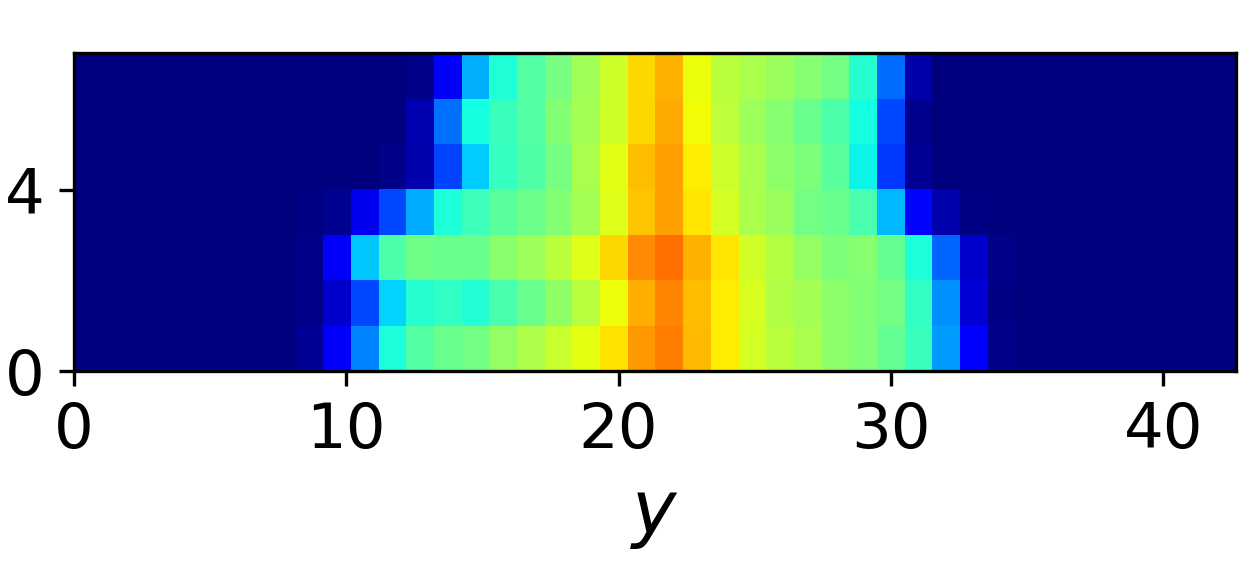}}
\subcaptionbox{Prior 3}{
\includegraphics[height=0.1\linewidth]{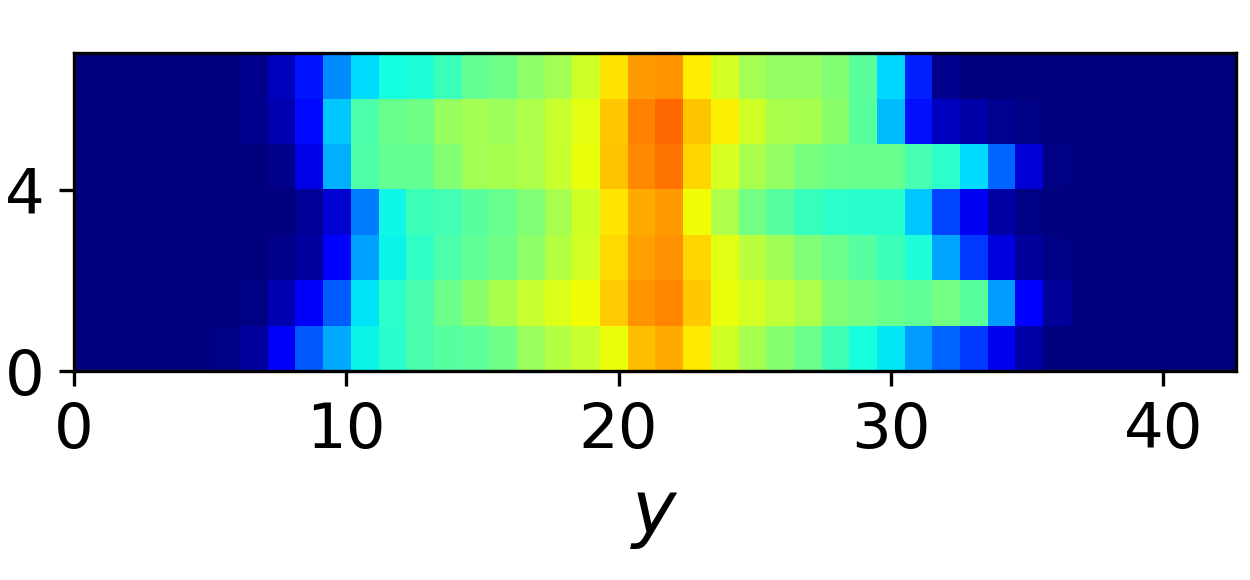}}
\subcaptionbox{Prior 4}{
\includegraphics[height=0.1\linewidth]{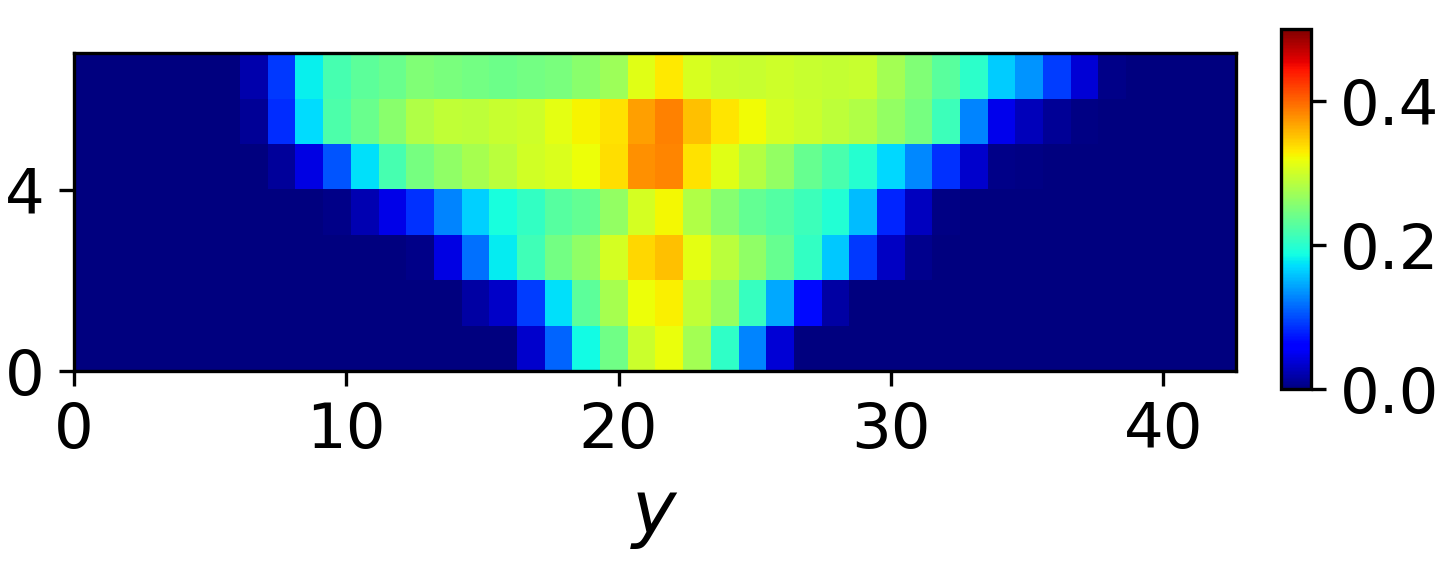}}

\subcaptionbox{Post.~1 (w/o seis)}{
\includegraphics[height=0.1\linewidth]{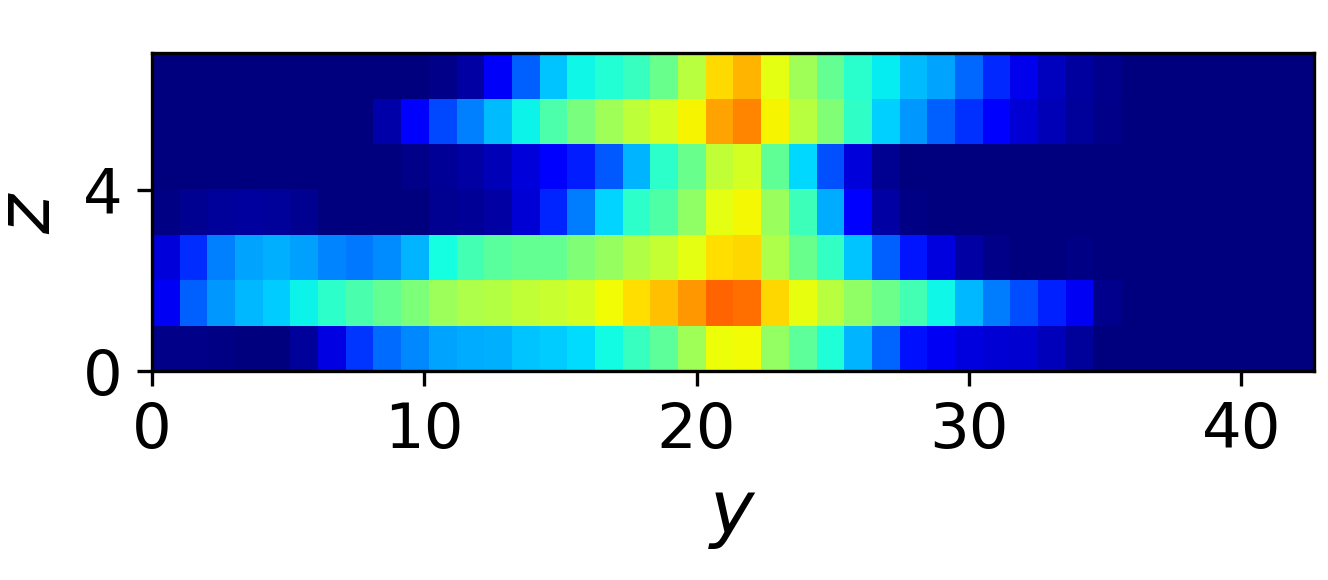}}
\subcaptionbox{Post.~2 (w/o seis)}{
\includegraphics[height=0.1\linewidth]{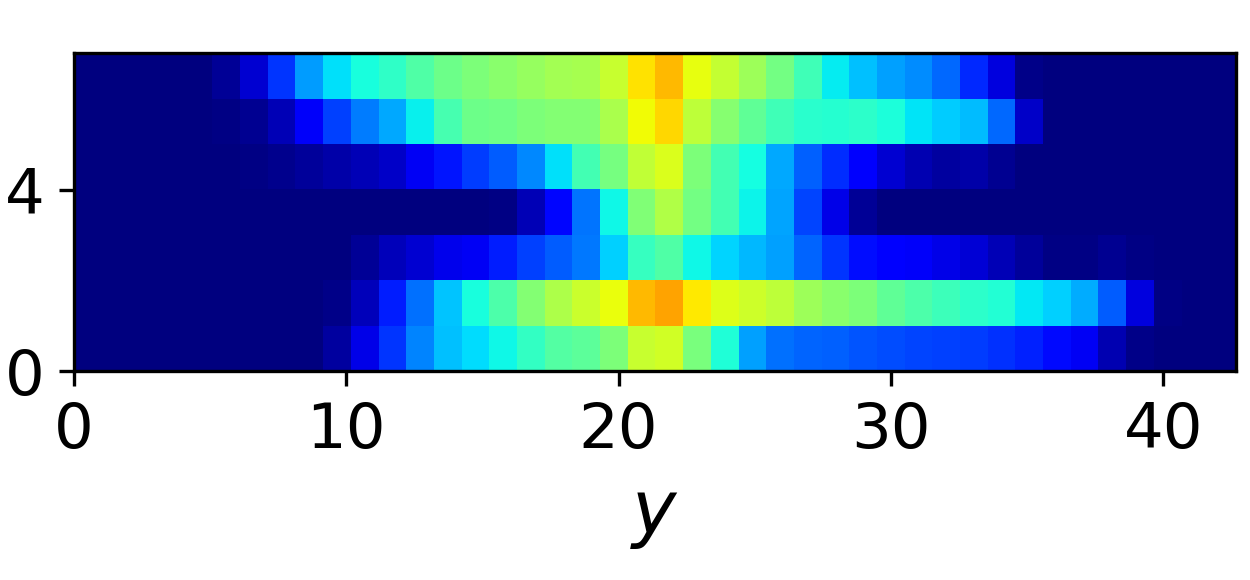}}
\subcaptionbox{Post.~3 (w/o seis)}{
\includegraphics[height=0.1\linewidth]{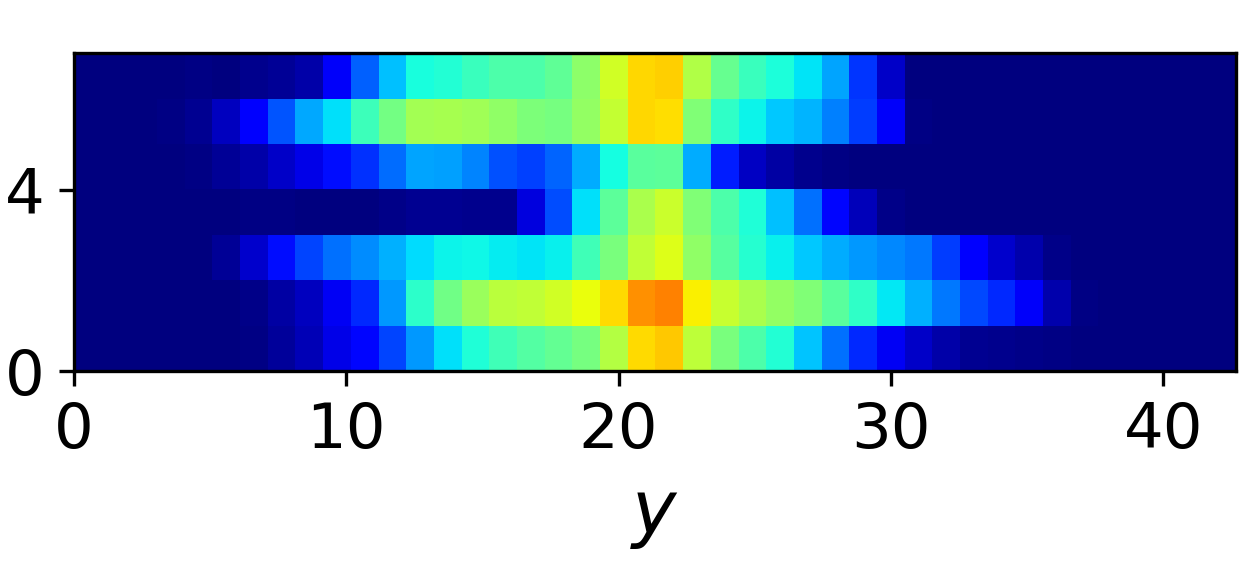}}
\subcaptionbox{Post.~4 (w/o seis)}{
\includegraphics[height=0.1\linewidth]{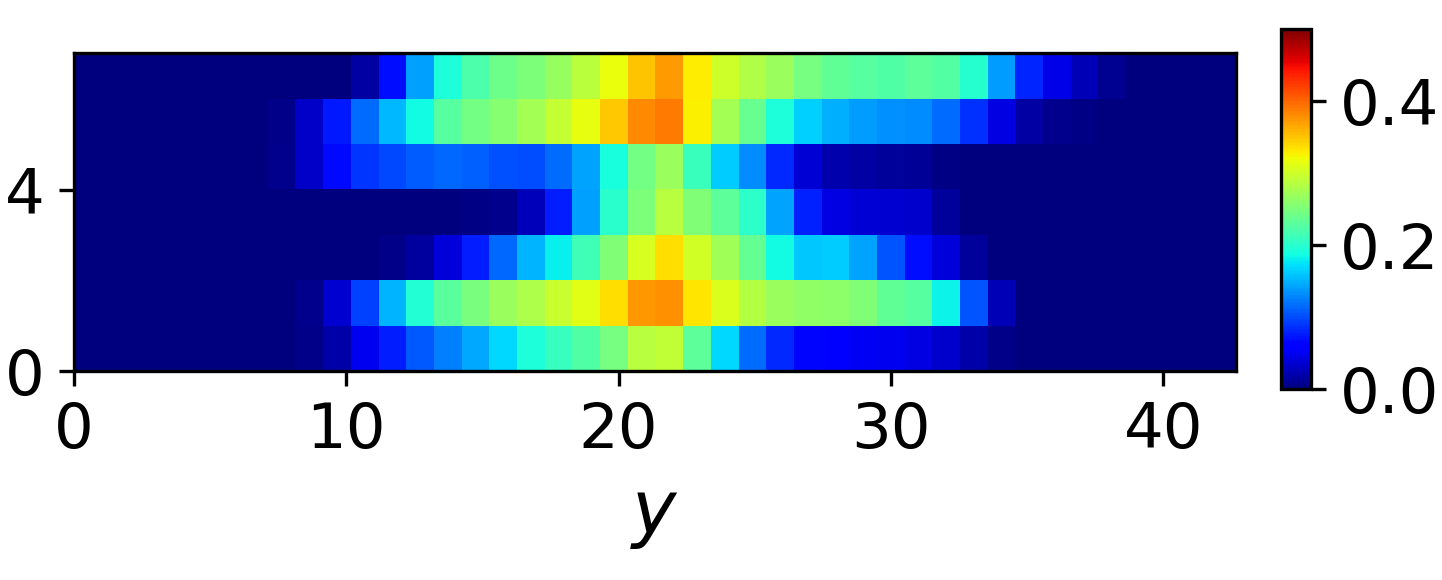}}

\subcaptionbox{Post.~1 (w/ loc seis)\centering}{
\includegraphics[height=0.1\linewidth]{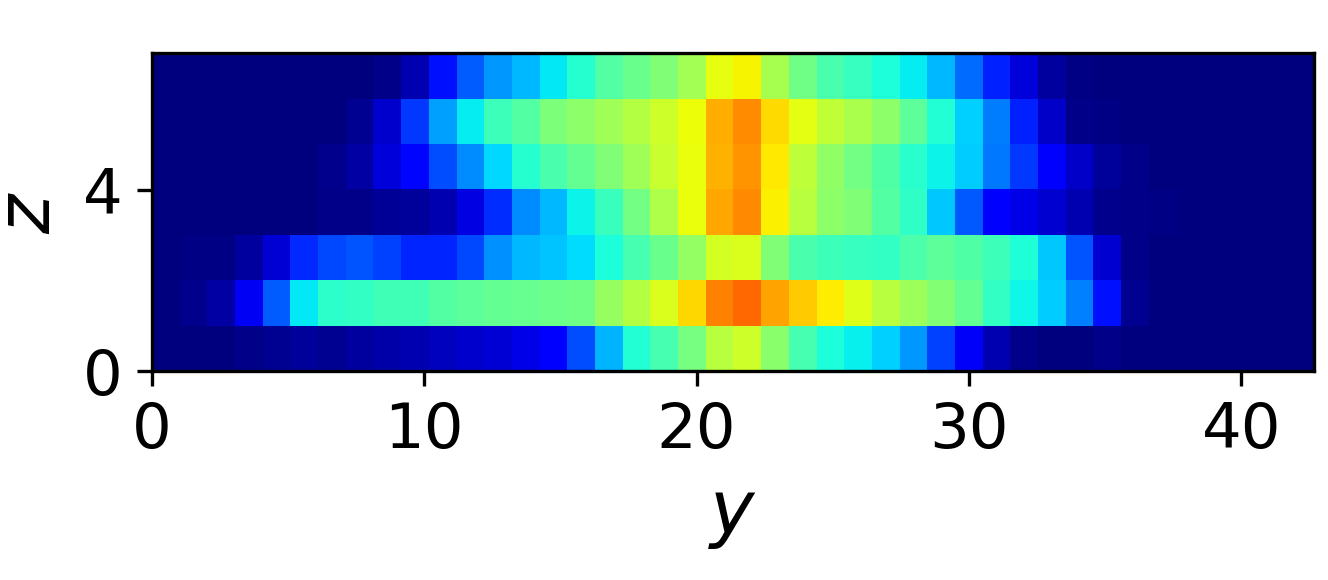}}
\subcaptionbox{Post.~2 (w/ loc seis)\centering}{
\includegraphics[height=0.1\linewidth]{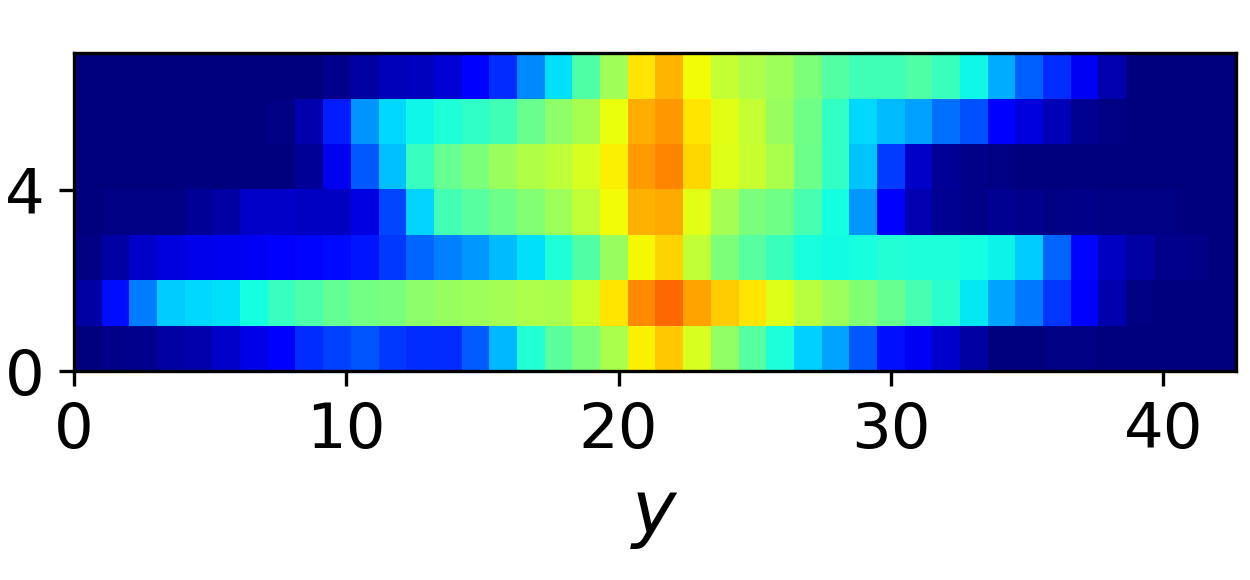}}
\subcaptionbox{Post.~3 (w/ loc seis)\centering}{
\includegraphics[height=0.1\linewidth]{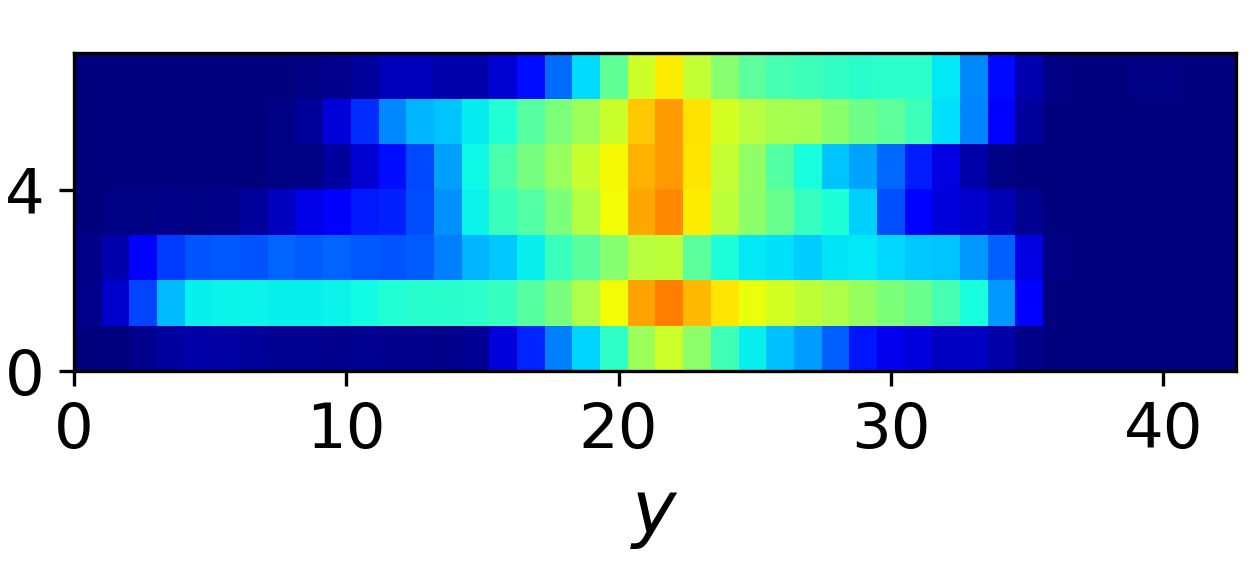}}
\subcaptionbox{Post.~4 (w/ loc seis)\centering}{
\includegraphics[height=0.1\linewidth]{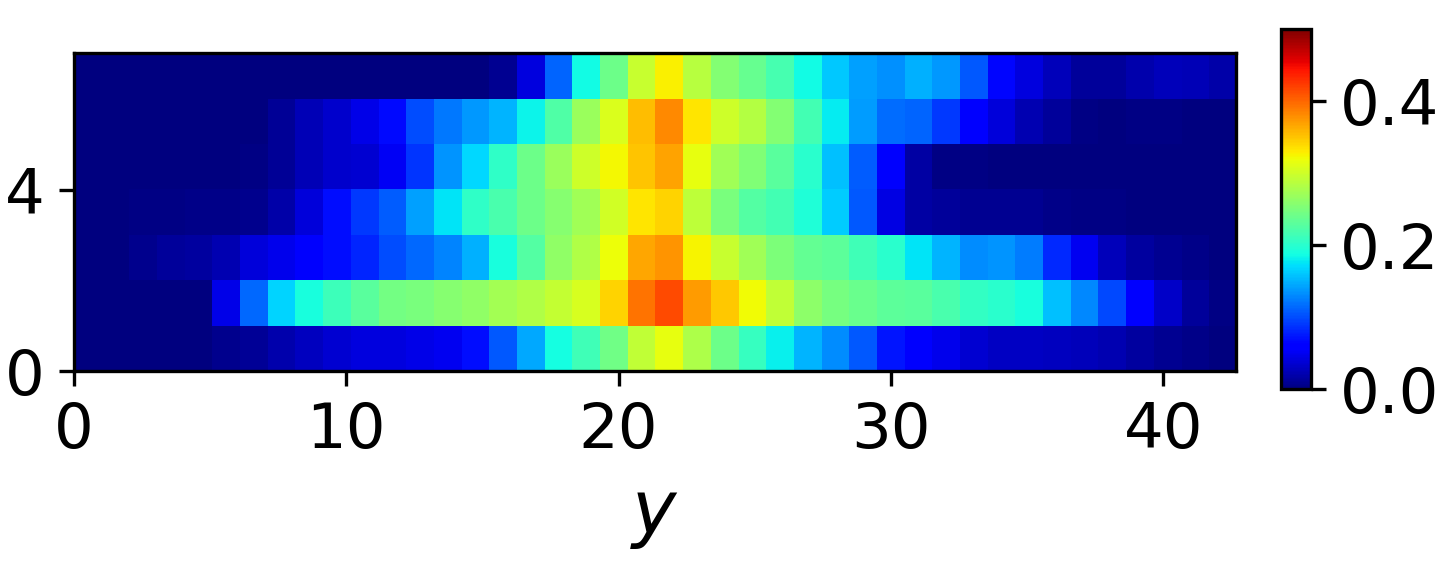}}

\subcaptionbox{Post.~1 (w/ $S_g$ seis)\centering}{
\includegraphics[height=0.1\linewidth]{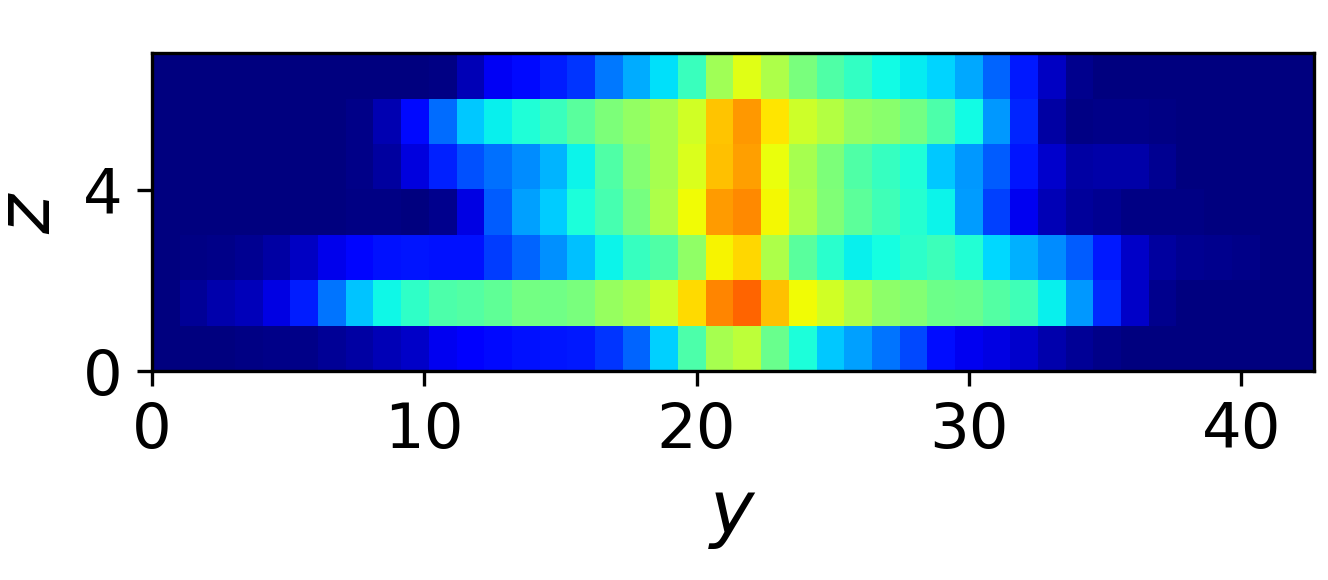}}
\subcaptionbox{Post.~2 (w/ $S_g$ seis)\centering}{
\includegraphics[height=0.1\linewidth]{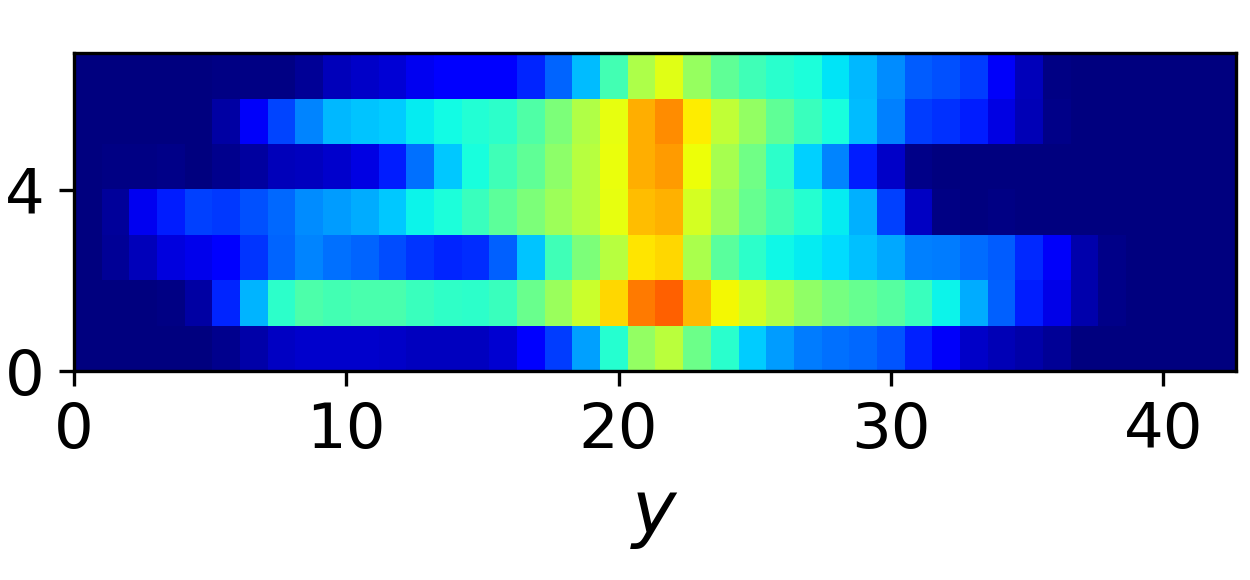}}
\subcaptionbox{Post.~3 (w/ $S_g$ seis)\centering}{
\includegraphics[height=0.1\linewidth]{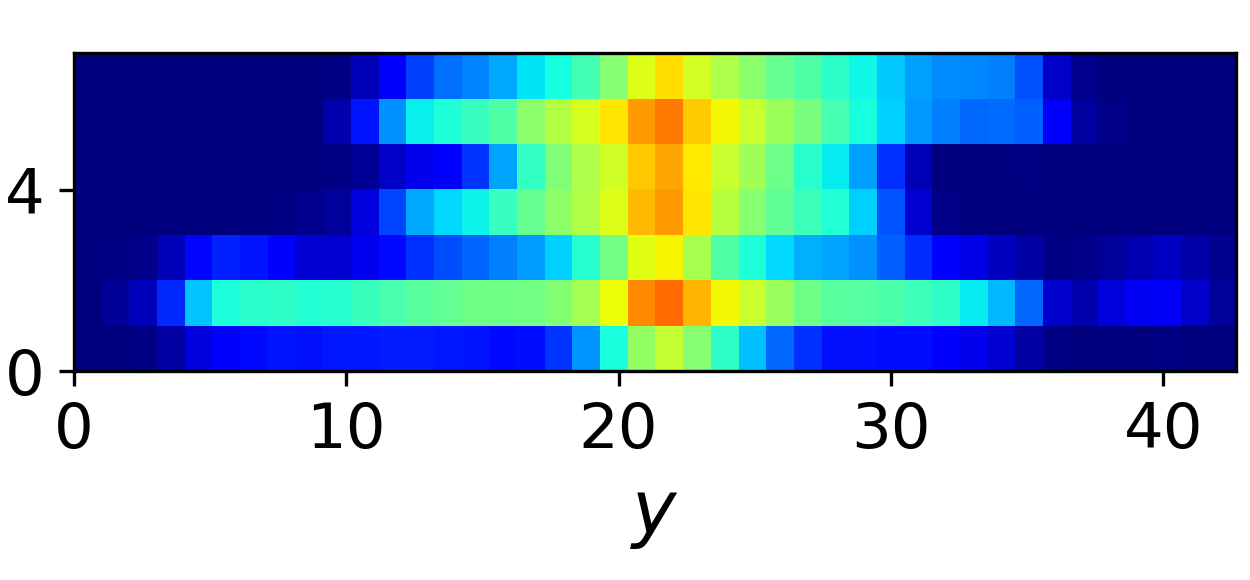}}
\subcaptionbox{Post.~4 (w/ $S_g$ seis)\centering}{
\includegraphics[height=0.1\linewidth]{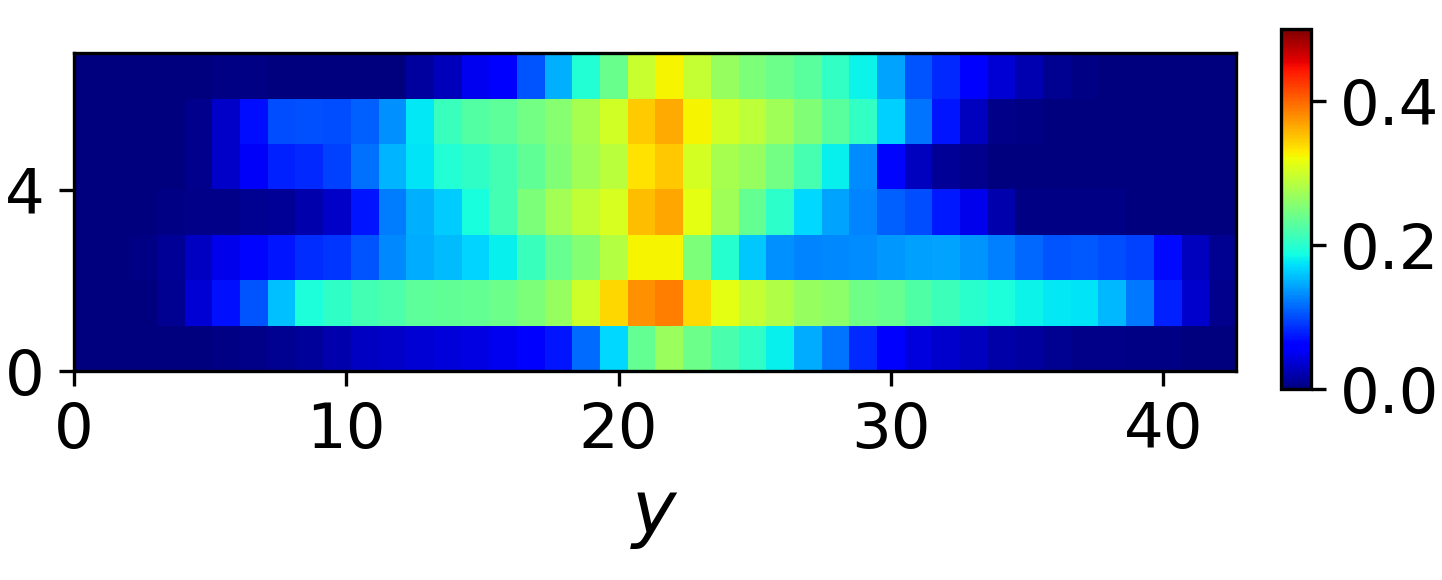}}

\caption{Representative interpreted seismic saturation fields in \texorpdfstring{$y$-$z$}{} cross sections through the injector at 1~year for prior realizations (first row), posterior realizations obtained using only monitoring well data (second row), posterior realizations obtained using monitoring well and plume location seismic data (third row), and posterior realizations obtained using monitoring well and saturation seismic data (fourth row). True result shown in Figure~\ref{fig:plume_true}(c).}
\label{fig:representative_x}
\end{figure}

\begin{figure}[htbp] 
\centering 
\vspace{0.35cm} 
\setlength{\lineskip}{\medskipamount}
\subcaptionbox{Prior 1}{
\includegraphics[height=0.1\linewidth]{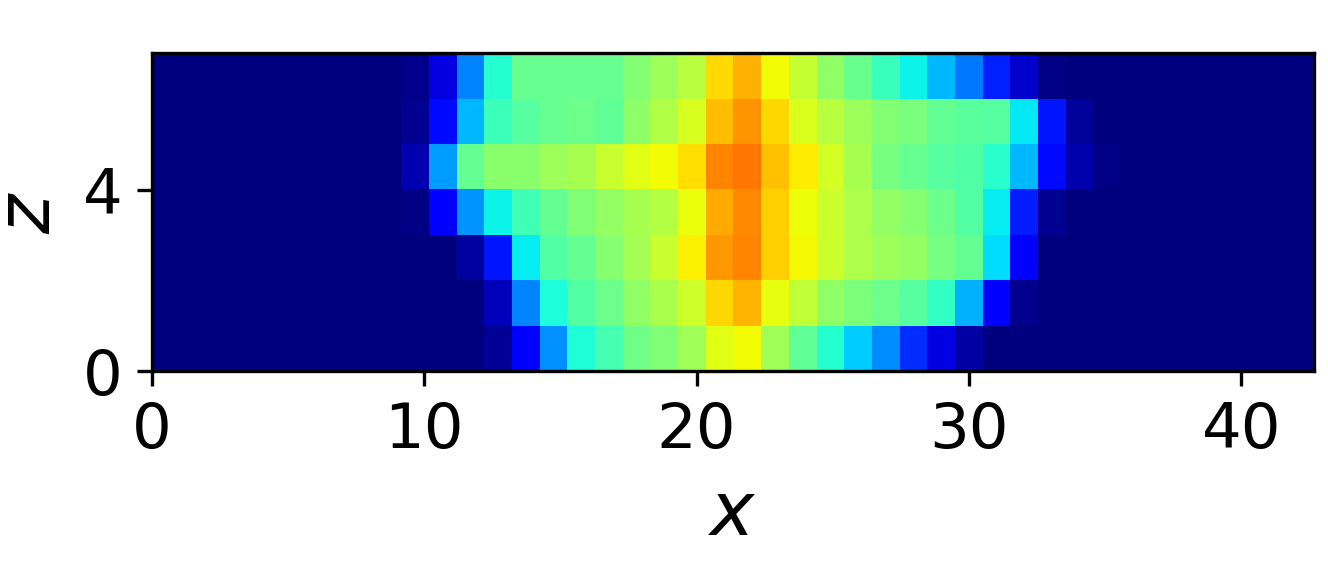}}
\subcaptionbox{Prior 2}{
\includegraphics[height=0.1\linewidth]{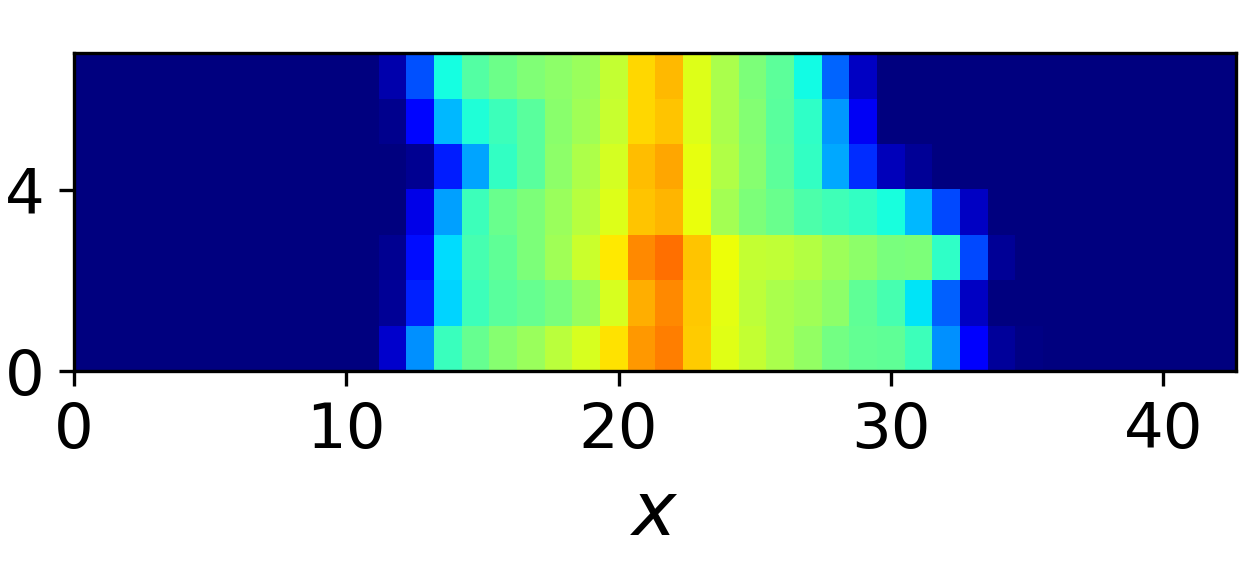}}
\subcaptionbox{Prior 3}{
\includegraphics[height=0.1\linewidth]{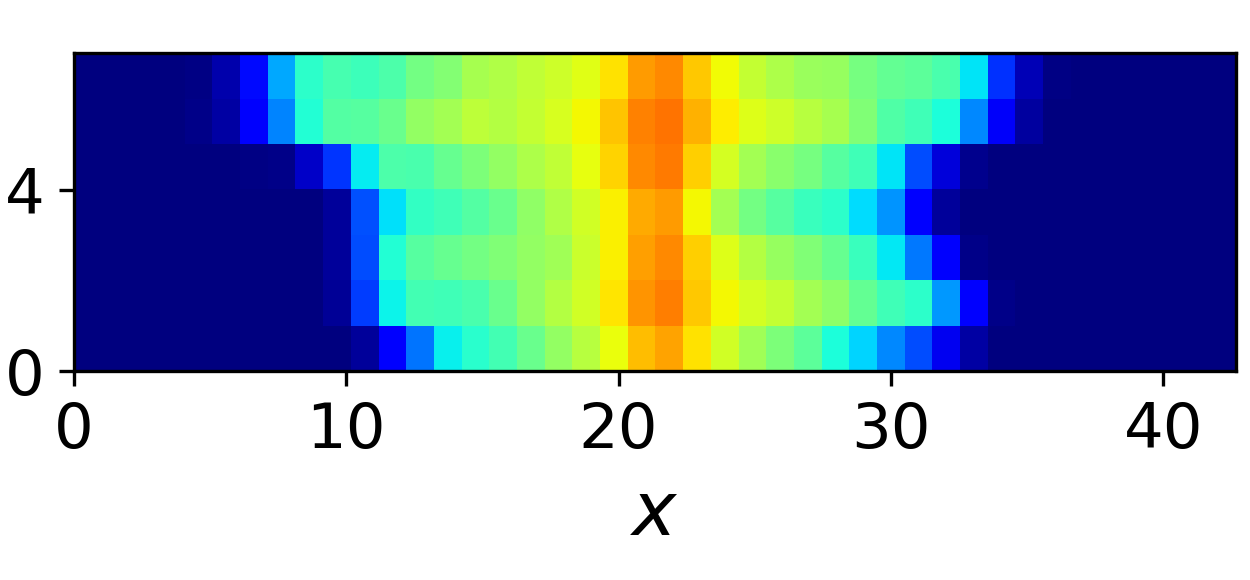}}
\subcaptionbox{Prior 4}{
\includegraphics[height=0.1\linewidth]{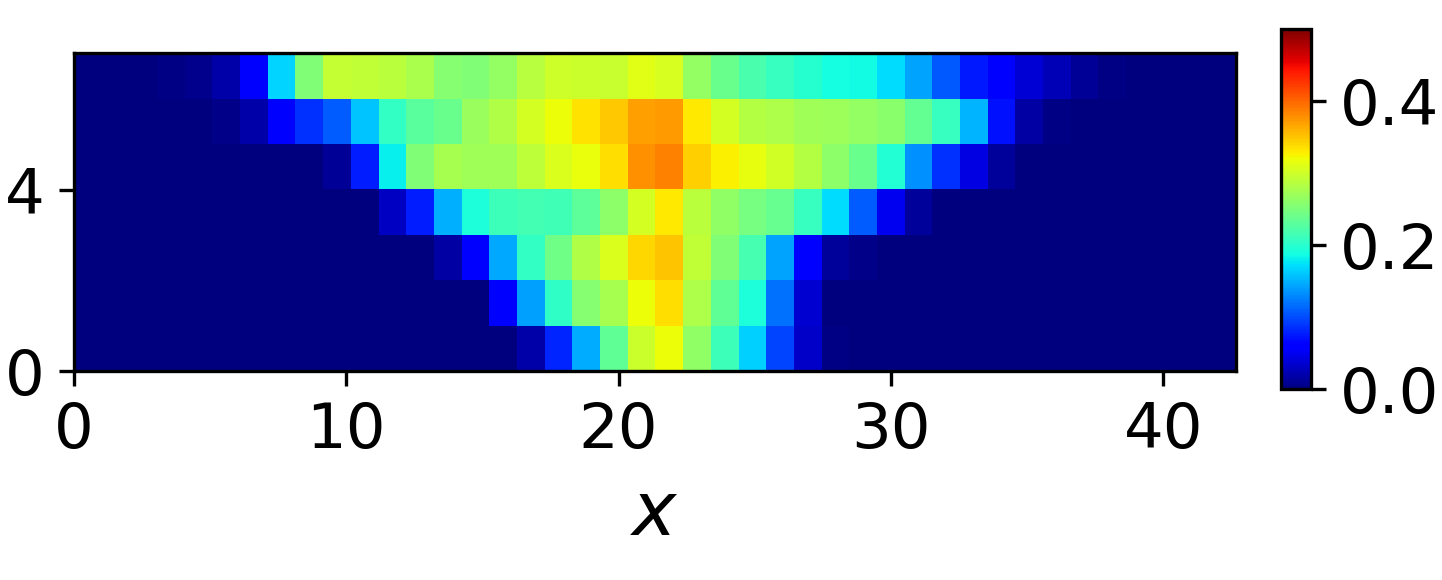}}

\subcaptionbox{Post.~1 (w/o seis)}{
\includegraphics[height=0.1\linewidth]{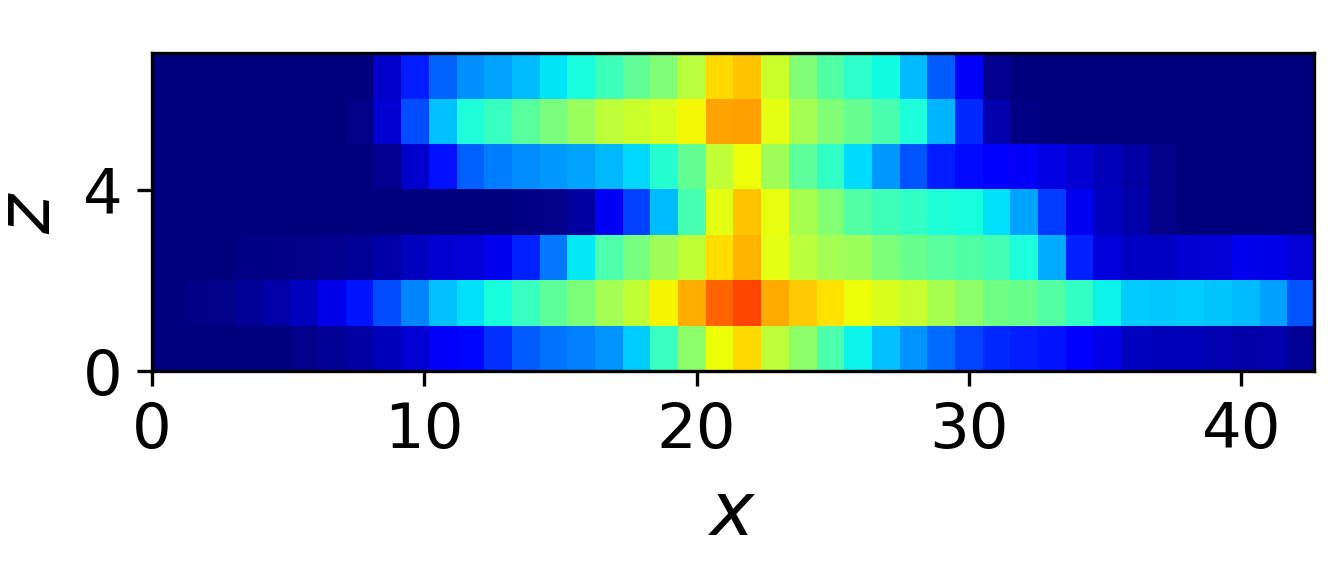}}
\subcaptionbox{Post.~2 (w/o seis)}{
\includegraphics[height=0.1\linewidth]{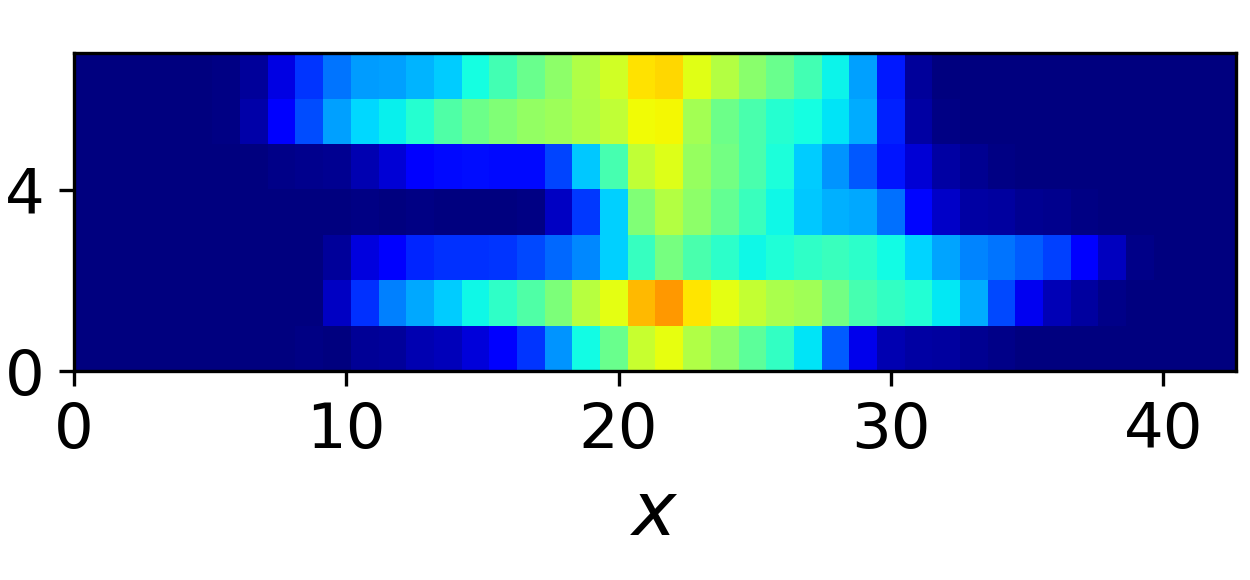}}
\subcaptionbox{Post.~3 (w/o seis)}{
\includegraphics[height=0.1\linewidth]{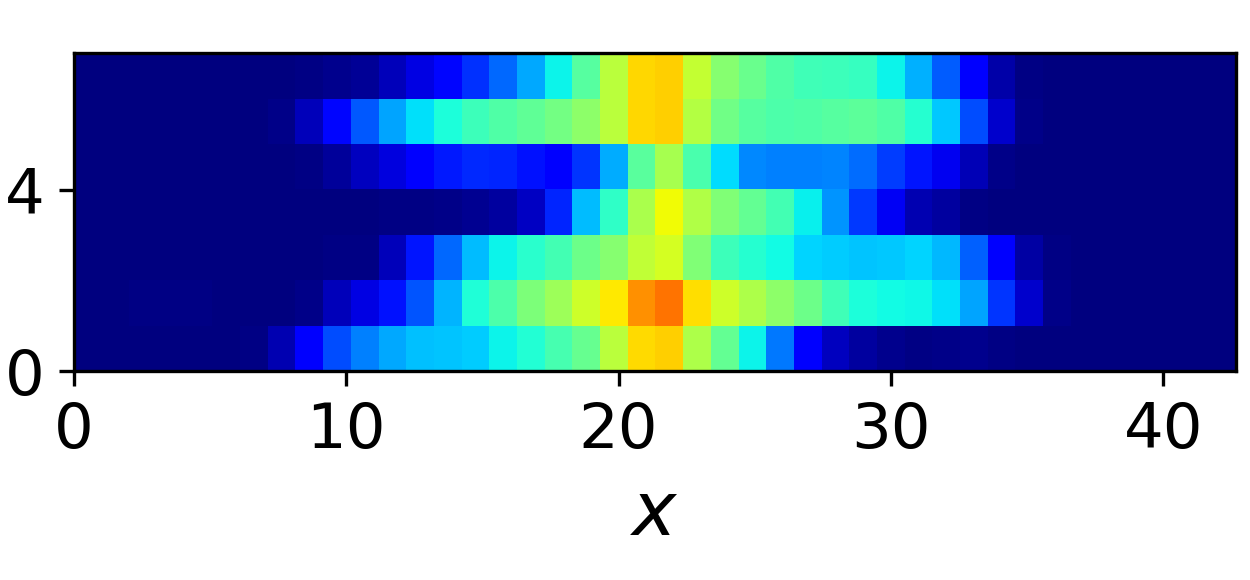}}
\subcaptionbox{Post.~4 (w/o seis)}{
\includegraphics[height=0.1\linewidth]{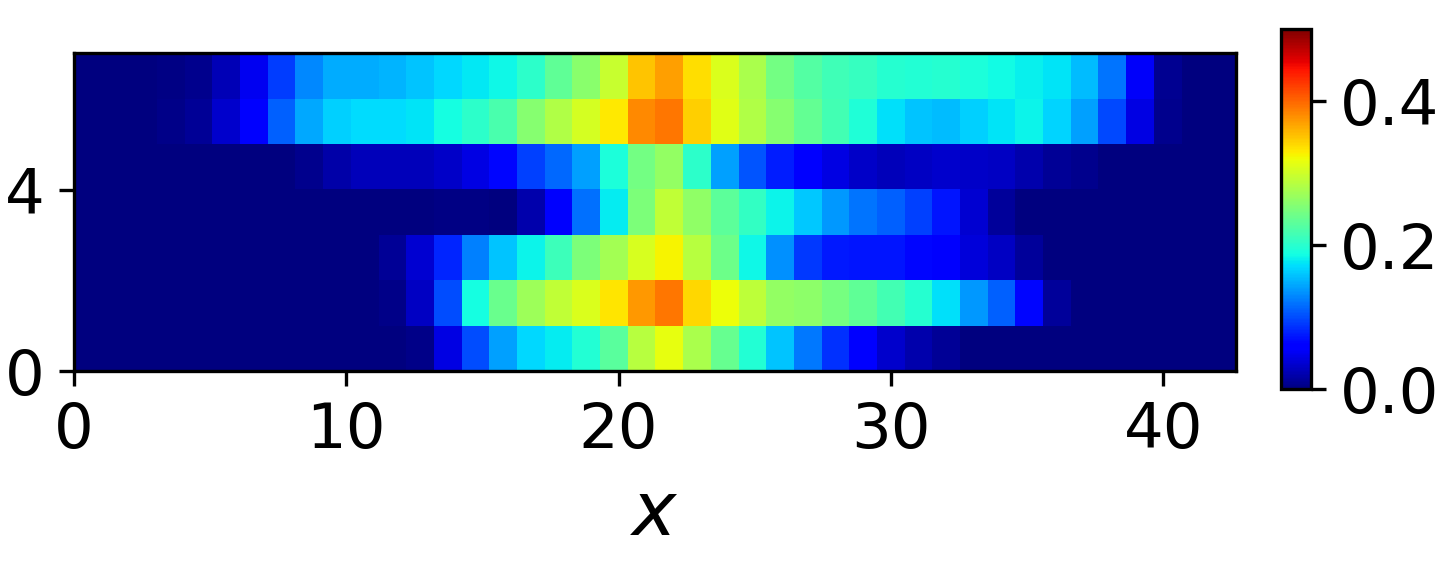}}

\subcaptionbox{Post.~1 (w/ loc seis)\centering}{
\includegraphics[height=0.1\linewidth]{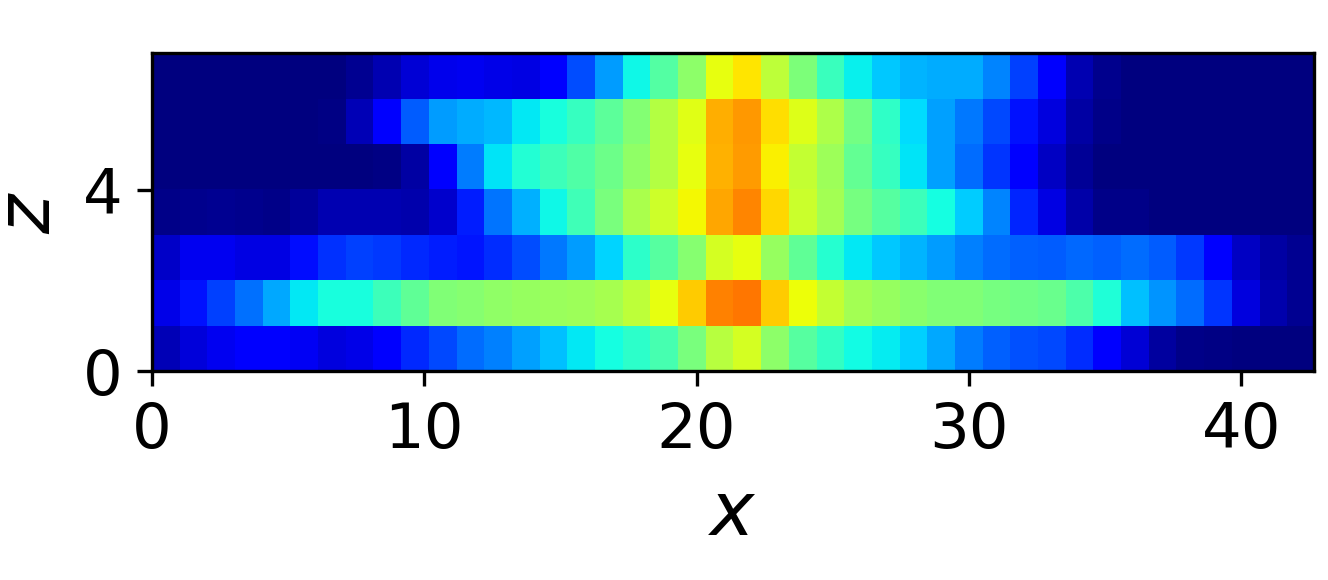}}
\subcaptionbox{Post.~2 (w/ loc seis)\centering}{
\includegraphics[height=0.1\linewidth]{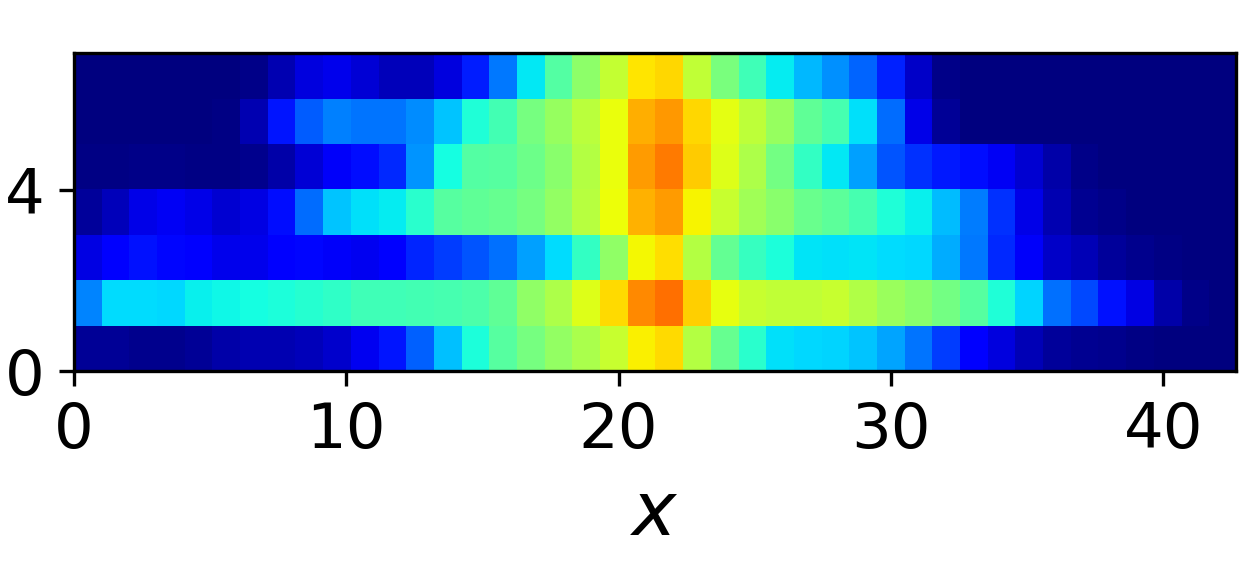}}
\subcaptionbox{Post.~3 (w/ loc seis)\centering}{
\includegraphics[height=0.1\linewidth]{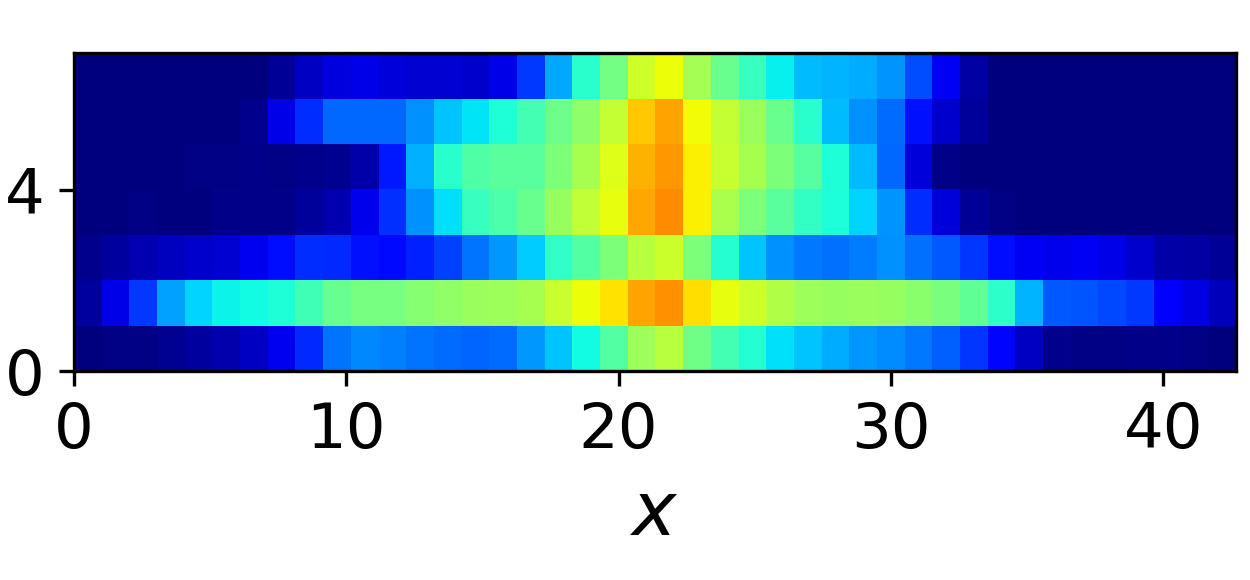}}
\subcaptionbox{Post.~4 (w/ loc seis)\centering}{
\includegraphics[height=0.1\linewidth]{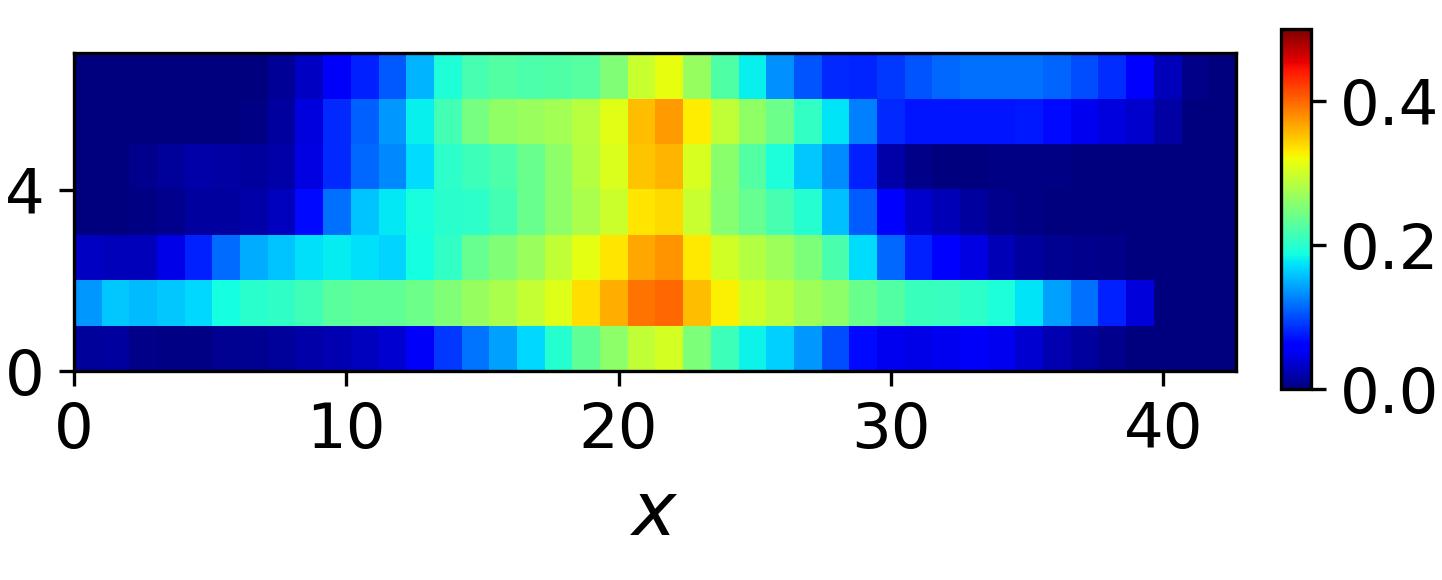}}

\subcaptionbox{Post.~1 (w/ $S_g$ seis)\centering}{
\includegraphics[height=0.1\linewidth]{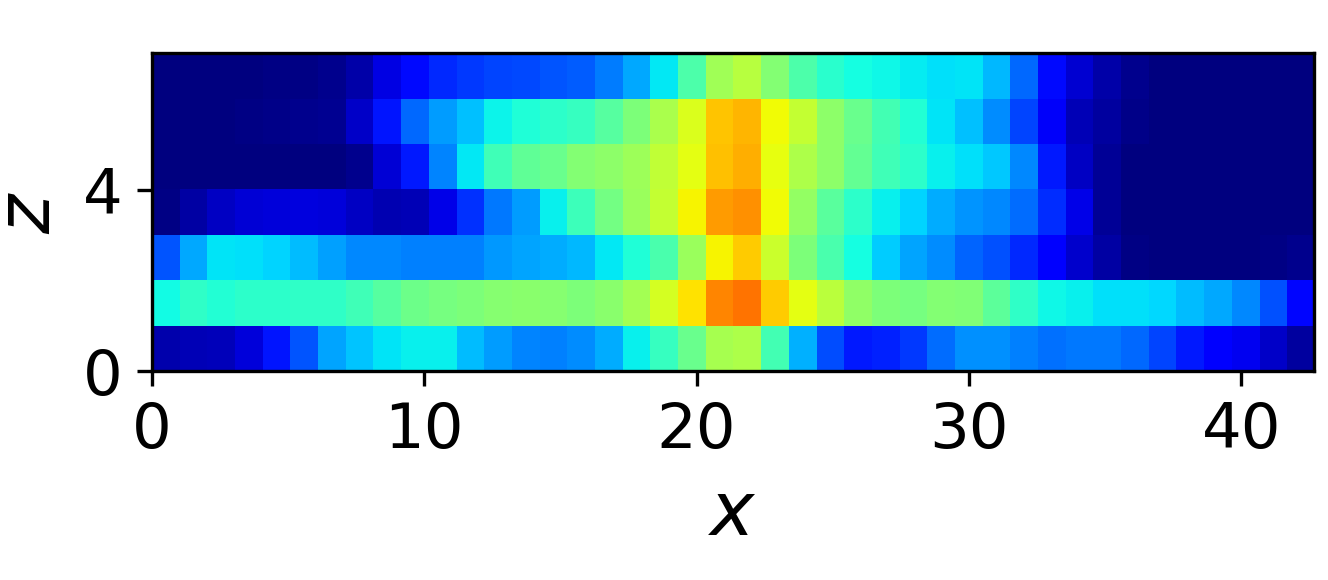}}
\subcaptionbox{Post.~2 (w/ $S_g$ seis)\centering}{
\includegraphics[height=0.1\linewidth]{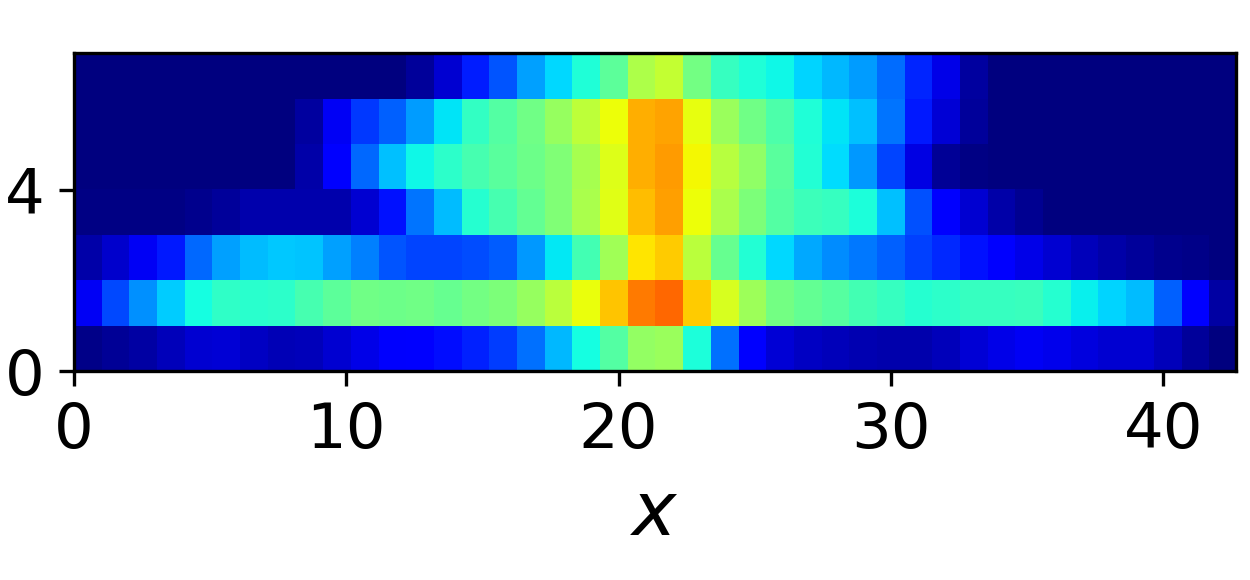}}
\subcaptionbox{Post.~3 (w/ $S_g$ seis)\centering}{
\includegraphics[height=0.1\linewidth]{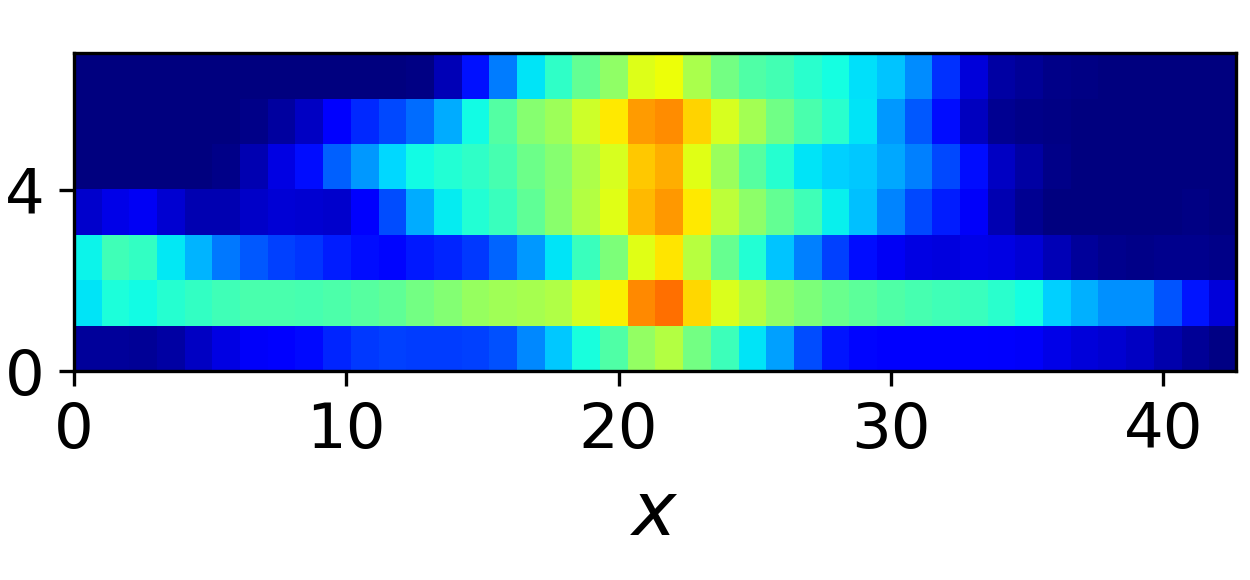}}
\subcaptionbox{Post.~4 (w/ $S_g$ seis)\centering}{
\includegraphics[height=0.1\linewidth]{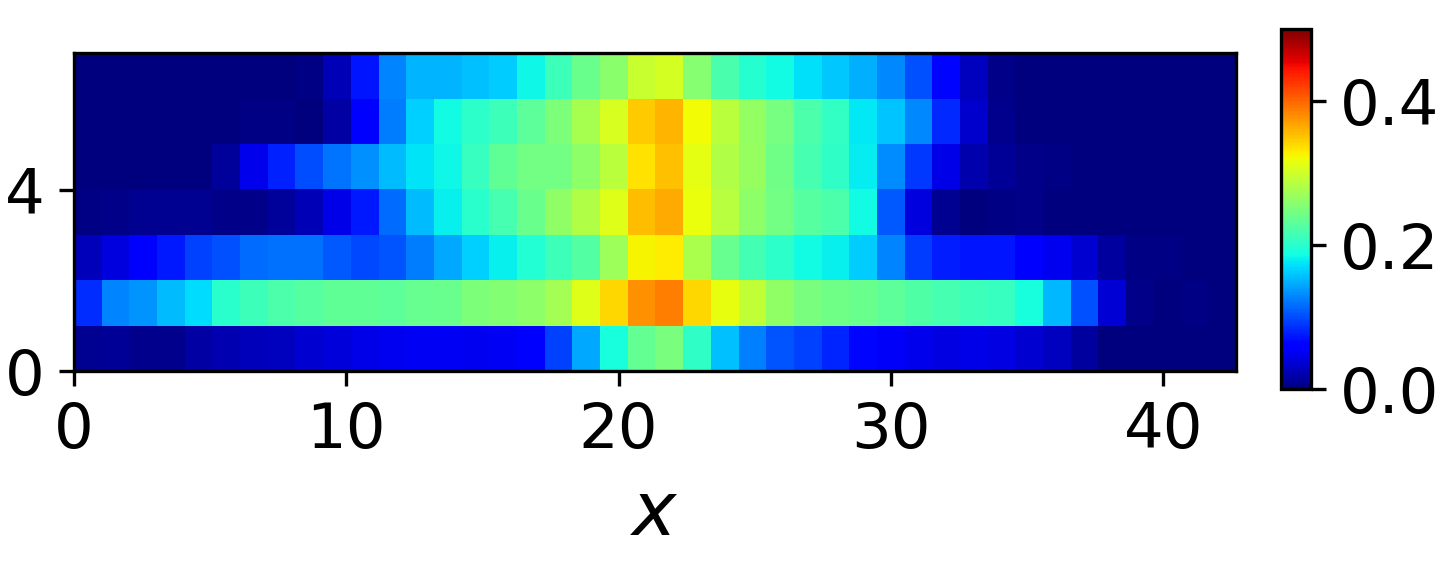}}
\caption{Representative interpreted seismic saturation fields in \texorpdfstring{$x$-$z$}{} cross sections through the injector at 1~year for prior realizations (first row), posterior realizations obtained using only monitoring well data (second row), posterior realizations obtained using monitoring well and plume location seismic data (third row), and posterior realizations obtained using monitoring well and saturation seismic data (fourth row). True result shown in Figure~\ref{fig:plume_true}(d).}
\label{fig:representative_y}
\end{figure}

The resulting 16 saturation fields are shown in Figure~\ref{fig:representative_x} ($y$-$z$ cross sections through the injection well) and in Figure~\ref{fig:representative_y} ($x$-$z$ cross sections through the injection well). These results are all at a time of 1~year. The true results for this case are shown in Figure~\ref{fig:plume_true}(c) and (d). The plumes in the prior realizations are fairly regular, varying from near-cylindrical to cone-shaped. The plumes for the posterior models all differ considerably from any of the prior plumes, and all display some amount of CO$_2$ channeling toward the bottom of the model.

Posterior saturation fields in which seismic data (of either type) are used are closer to the true saturation fields than those obtained with just monitoring well data. For example, the plumes in Figure~\ref{fig:representative_x}(e)-(h) are limited in extent in some of the middle layers, which is not consistent with the true result in Figure~\ref{fig:plume_true}(c). With either type of seismic data (Figure~\ref{fig:representative_x}(i)-(p)), the saturation fields more closely resemble the true results, both in terms of the shape of the CO$_2$ plume in the middle layers and the channeling toward the bottom of the model. Posterior results obtained with interpreted saturation seismic data appear to provide slightly more realistic saturation fields than those using only plume location data. For example, in Figure~\ref{fig:representative_x}(m)-(p), we see three somewhat distinct channels to the left of the injector along with CO$_2$ moving further to the left than the right. These behaviors, consistent with Figure~\ref{fig:plume_true}(c), are not consistently observed in Figure~\ref{fig:representative_x}(i)-(l). Analogous observations can be made in the $x$-$z$ cross sections shown in Figure~\ref{fig:representative_y}(m)-(p) in relation to those in Figure~\ref{fig:representative_y}(i)-(l), i.e., the extensive channeling at the bottom of the model evident in Figure~\ref{fig:plume_true}(d) is better resolved in Figure~\ref{fig:representative_y}(m)-(p).

\begin{figure}[H] 
\centering 
\vspace{0.35cm} 
\setlength{\lineskip}{\medskipamount}
\subcaptionbox{True model}{
\includegraphics[height=0.18\linewidth]{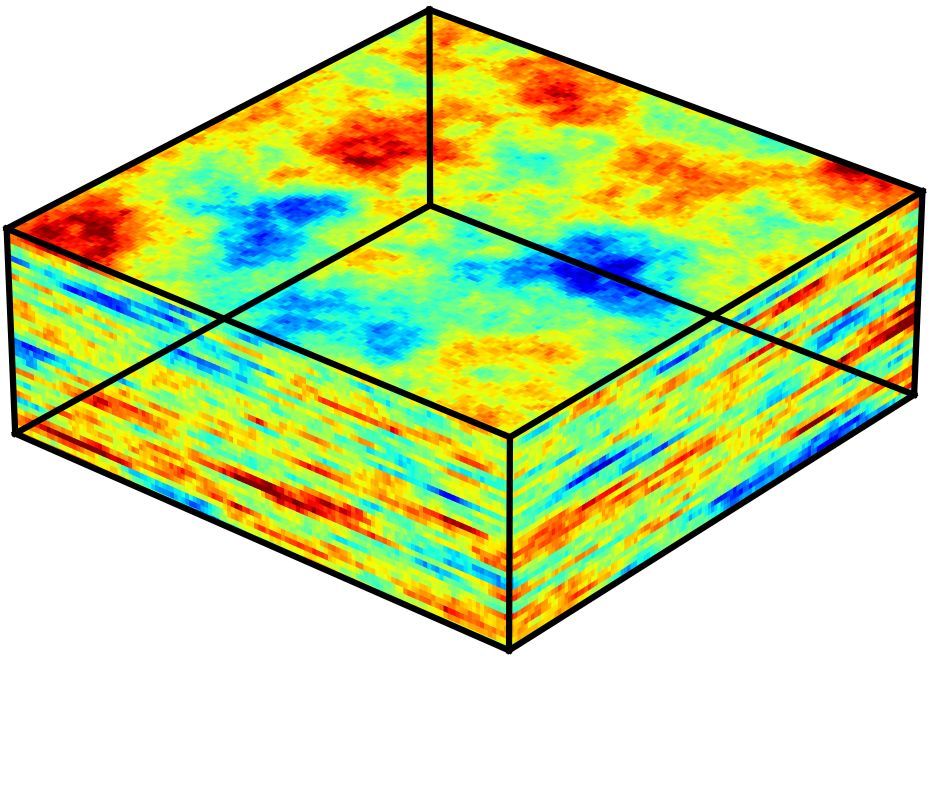}}
\subcaptionbox{Prior 1}{
\includegraphics[height=0.18\linewidth]{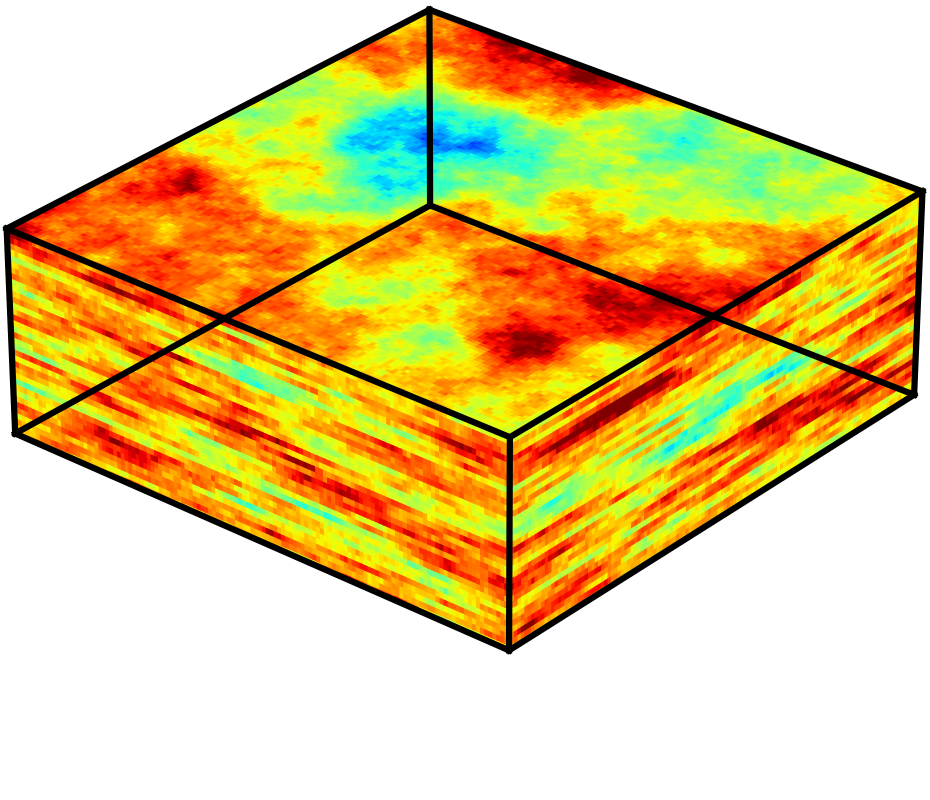}}
\subcaptionbox{Prior 2}{
\includegraphics[height=0.18\linewidth]{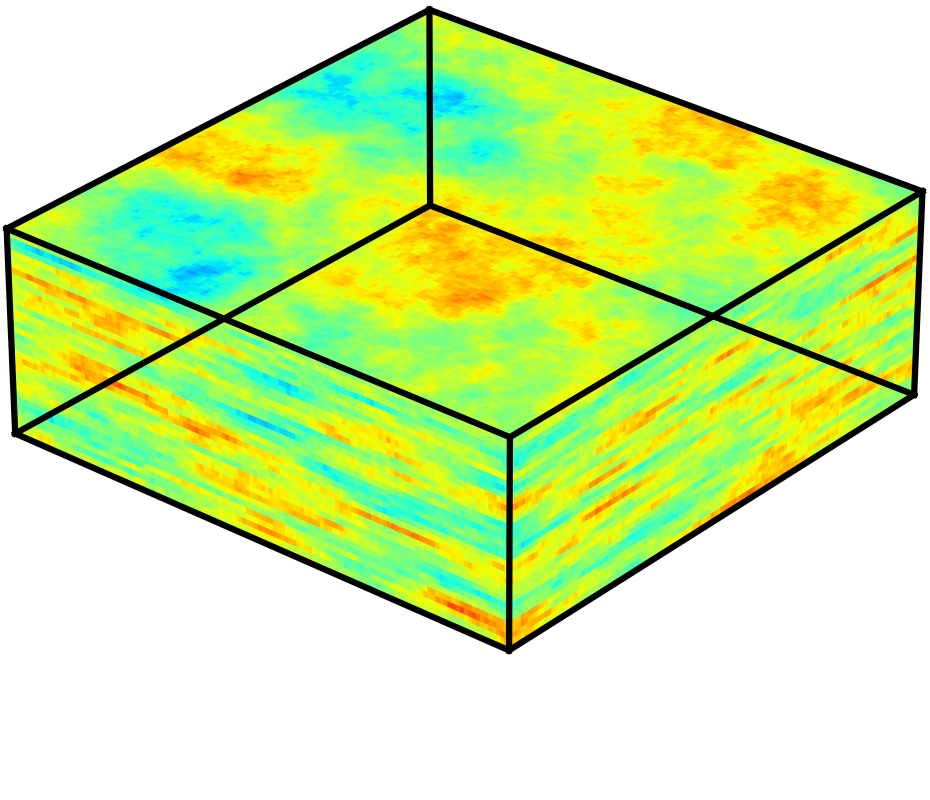}}
\subcaptionbox{Prior 3}{
\includegraphics[height=0.18\linewidth]{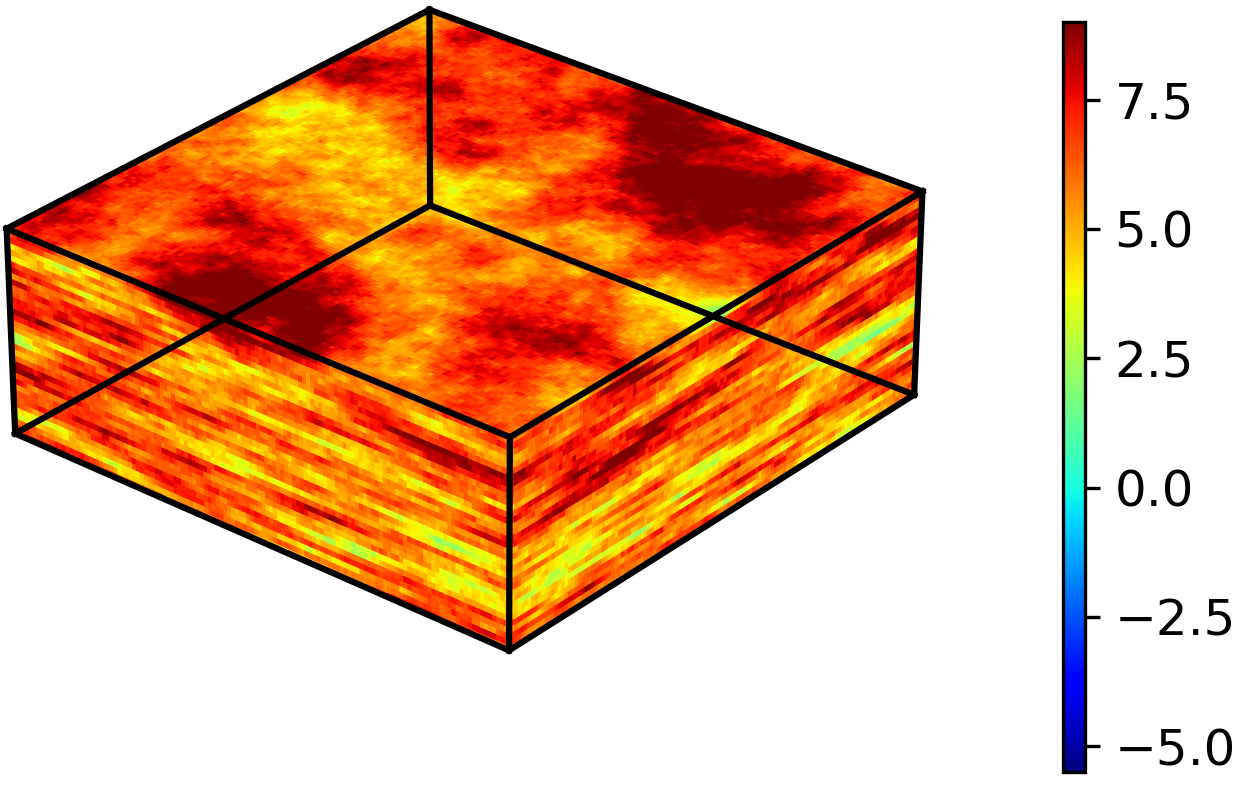}}

\hspace{0.20925\linewidth}
\subcaptionbox{Posterior 1}{
\includegraphics[height=0.18\linewidth]{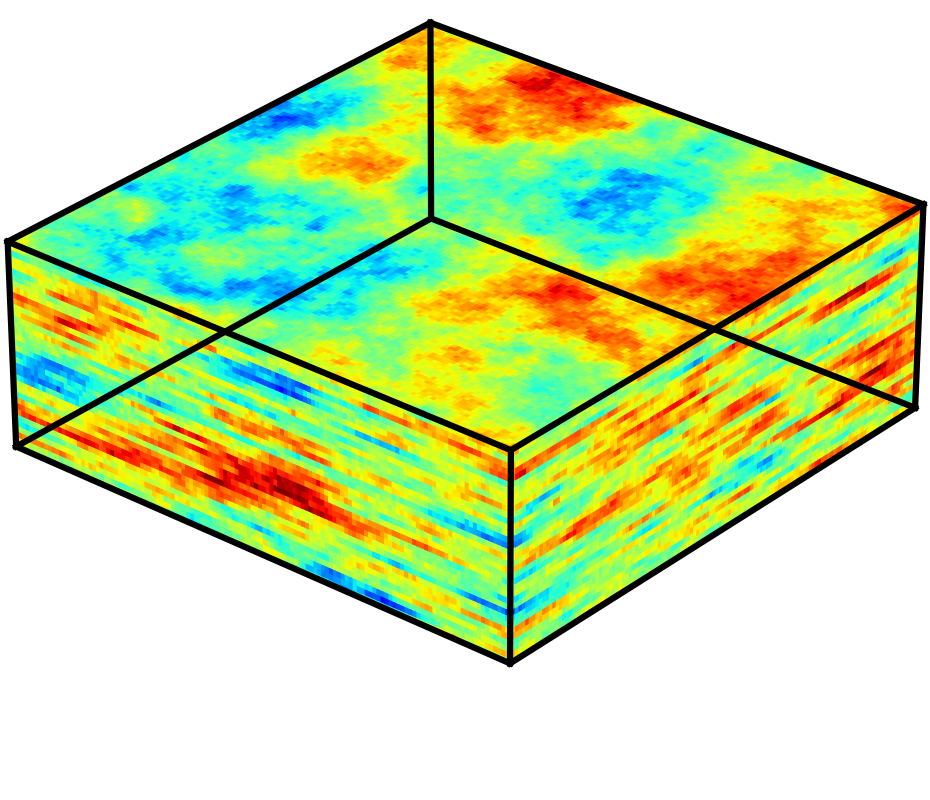}}
\subcaptionbox{Posterior 2}{
\includegraphics[height=0.18\linewidth]{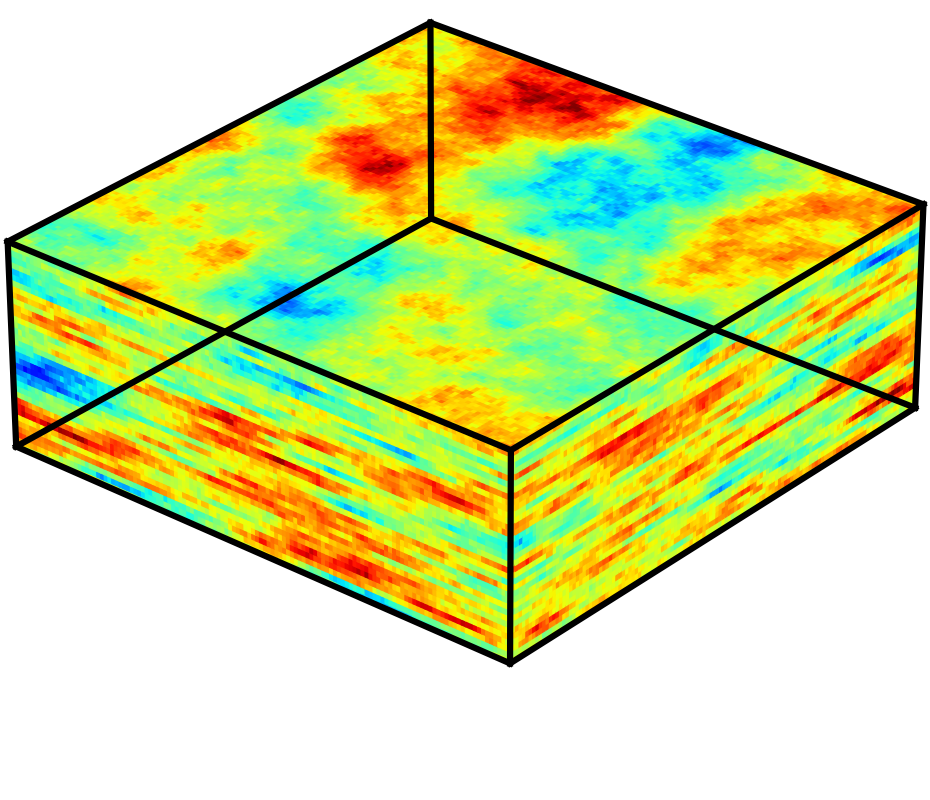}}
\subcaptionbox{Posterior 3}{
\includegraphics[height=0.18\linewidth]{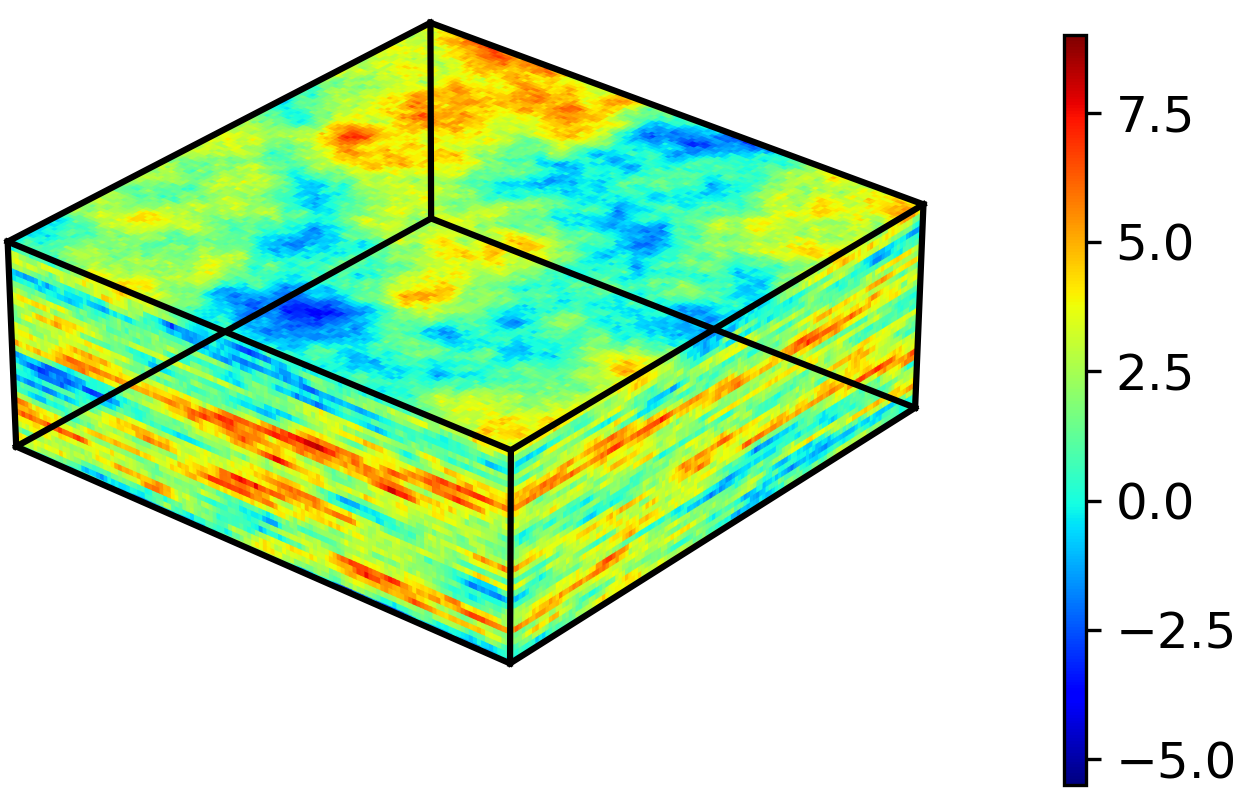}}

\caption{True model, three randomly sampled prior realizations, and three randomly sampled posterior realizations obtained using monitoring well and interpreted saturation seismic data. Geomodels shown in terms of \texorpdfstring{$\log k_x$}{}, with \texorpdfstring{$k_x$}{} in md.}
\label{fig:logk_posterior}
\end{figure}

Although the posterior saturation fields shown in Figures~\ref{fig:representative_x} and \ref{fig:representative_y} are at the coarse (interpreted seismic) scale, it is important to reiterate that our procedure provides high-resolution posterior geomodels. This is illustrated in Figure~\ref{fig:logk_posterior}, where we display randomly sampled prior and posterior geomodel realizations ($\log k_x$ is the quantity displayed) along with the true model. The posterior models derive from history matching using monitoring well and interpreted saturation seismic data. It is evident that the three prior models show very different ranges for $\log k_x$, with notable scale variation between Prior~2 and Prior~3 (Figure~\ref{fig:logk_posterior}(c) and (d)). The posterior realizations, by contrast, show similar ranges and variability in $\log k_x$, as well as general correspondence with the true model (Figure~\ref{fig:logk_posterior}(a)). This is as expected since the posterior uncertainty in $\mu_{\log k}$ and $\sigma_{\log k}$ is relatively small.

\begin{figure}[H] 
\centering 
\vspace{0.35cm} 
\setlength{\lineskip}{\medskipamount}
\subcaptionbox{Posterior 1 ($y$-$z$, sim)}{
\includegraphics[height=0.135\linewidth]{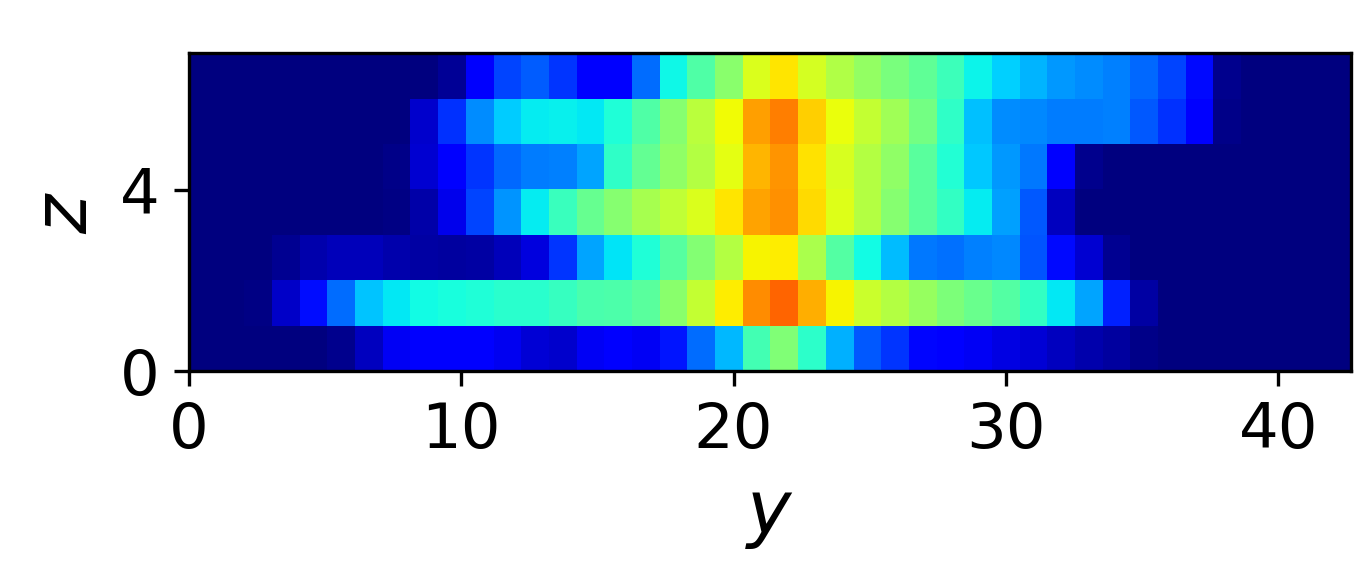}}
\subcaptionbox{Posterior 2 ($y$-$z$, sim)}{
\includegraphics[height=0.135\linewidth]{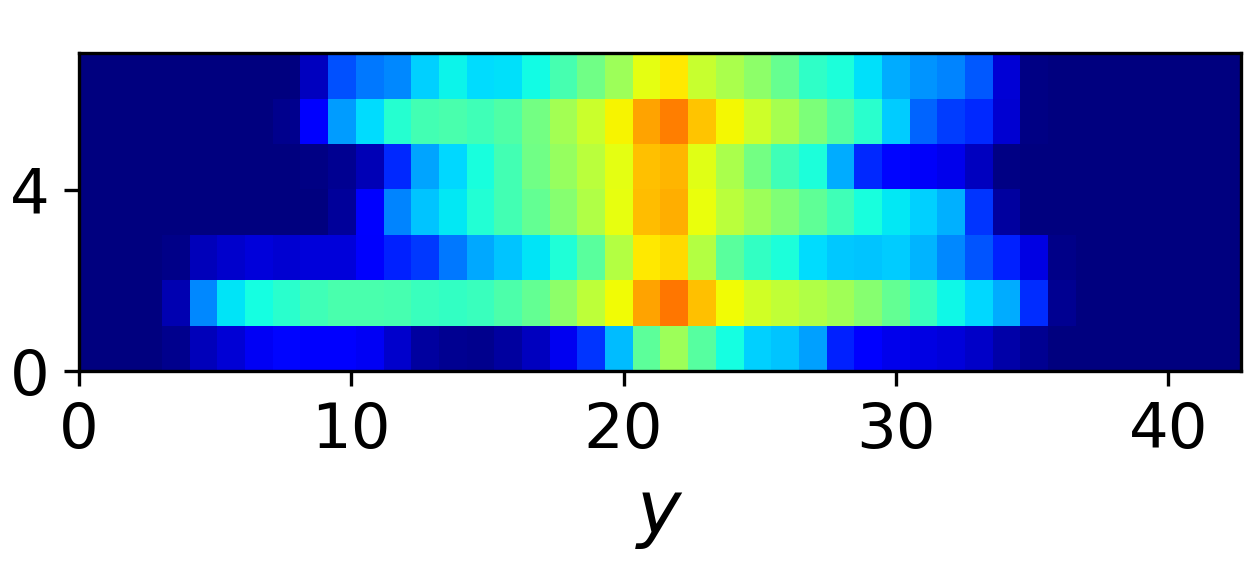}}
\subcaptionbox{Posterior 3 ($y$-$z$, sim)}{
\includegraphics[height=0.135\linewidth]{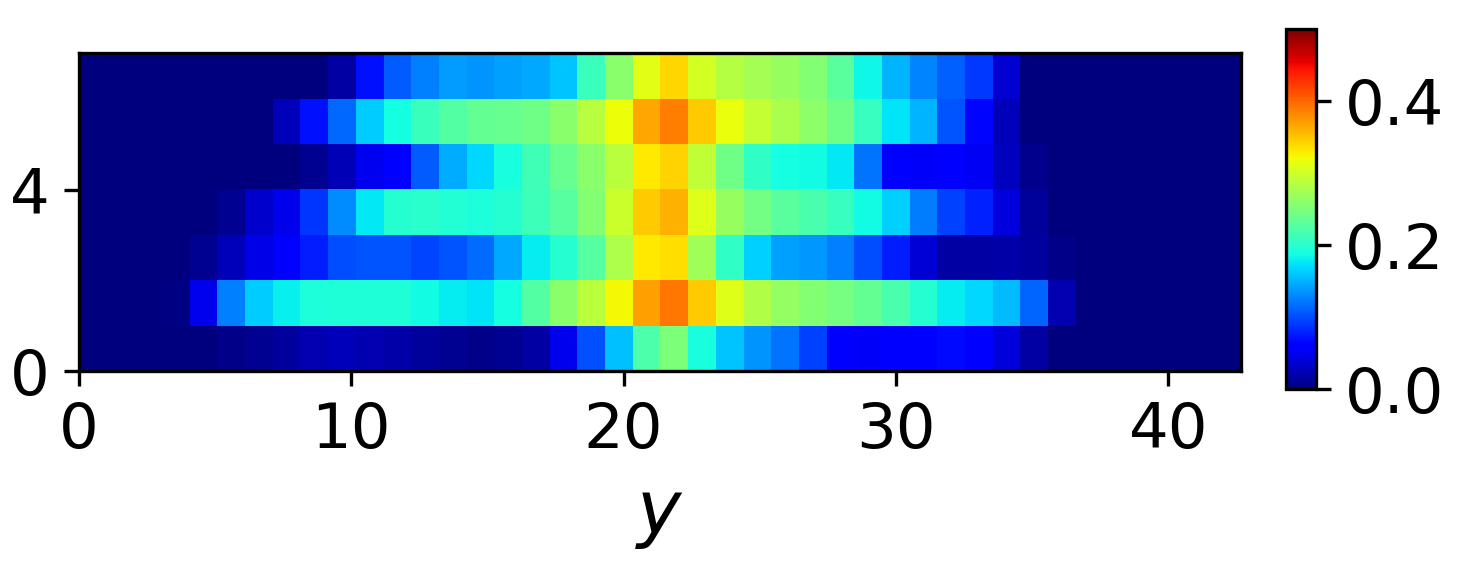}}

\subcaptionbox{Posterior 1 ($y$-$z$, surr)}{
\includegraphics[height=0.135\linewidth]{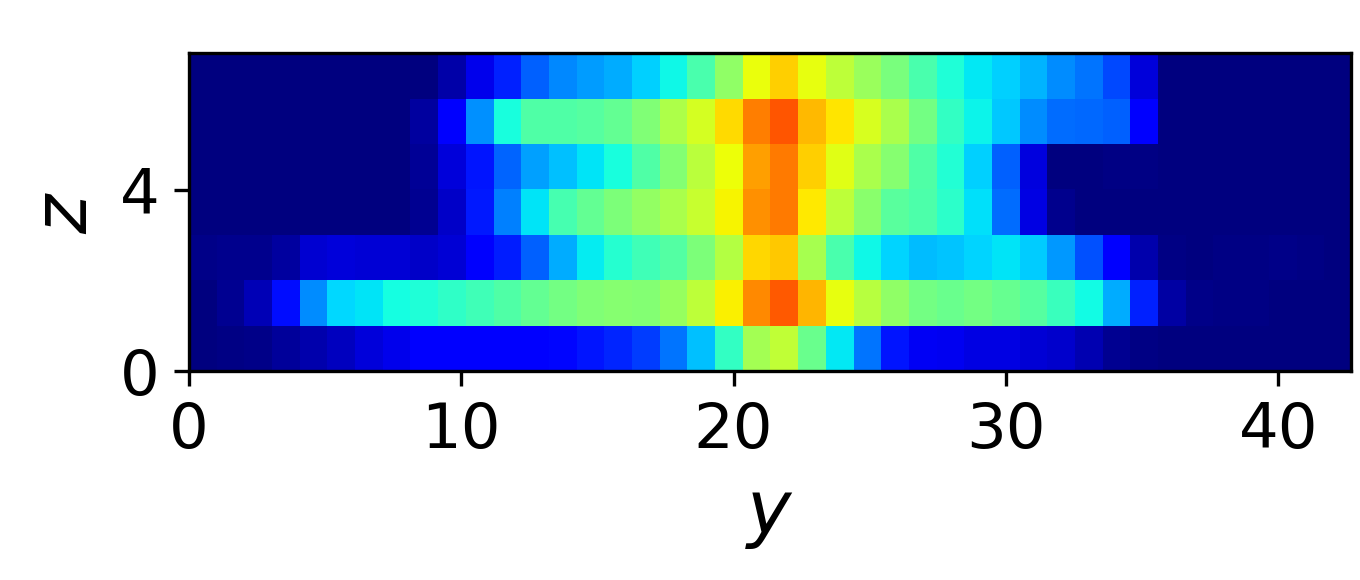}}
\subcaptionbox{Posterior 2 ($y$-$z$, surr)}{
\includegraphics[height=0.135\linewidth]{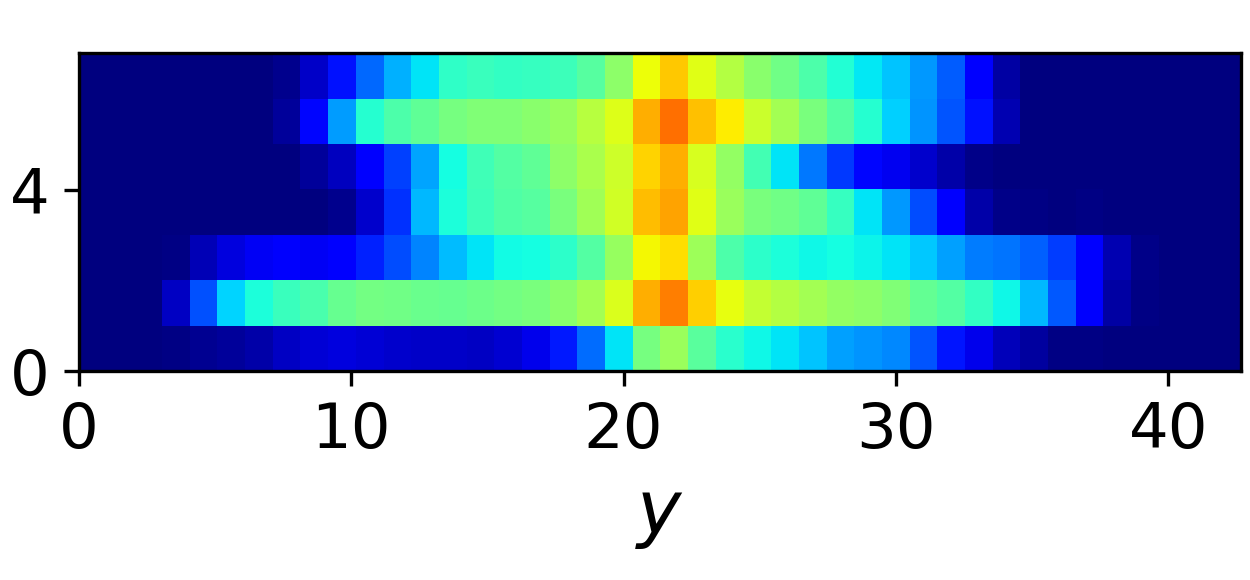}}
\subcaptionbox{Posterior 3 ($y$-$z$, surr)}{
\includegraphics[height=0.135\linewidth]{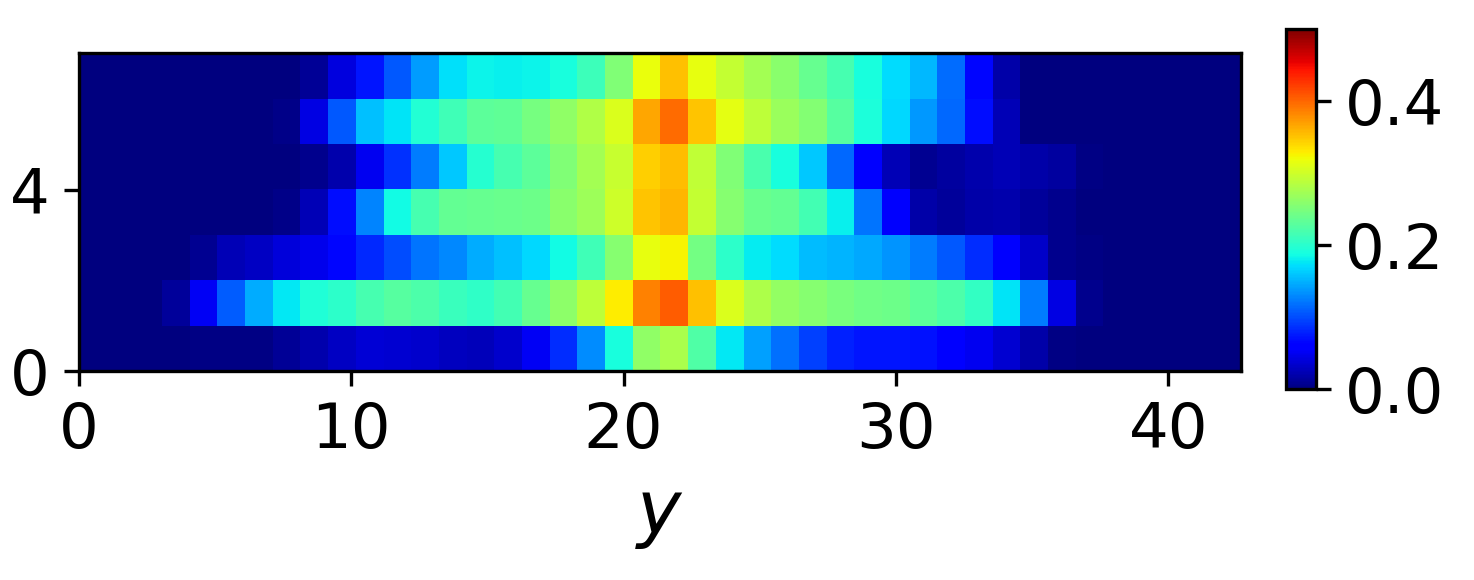}}

\subcaptionbox{Posterior 1 ($x$-$z$, sim)}{
\includegraphics[height=0.135\linewidth]{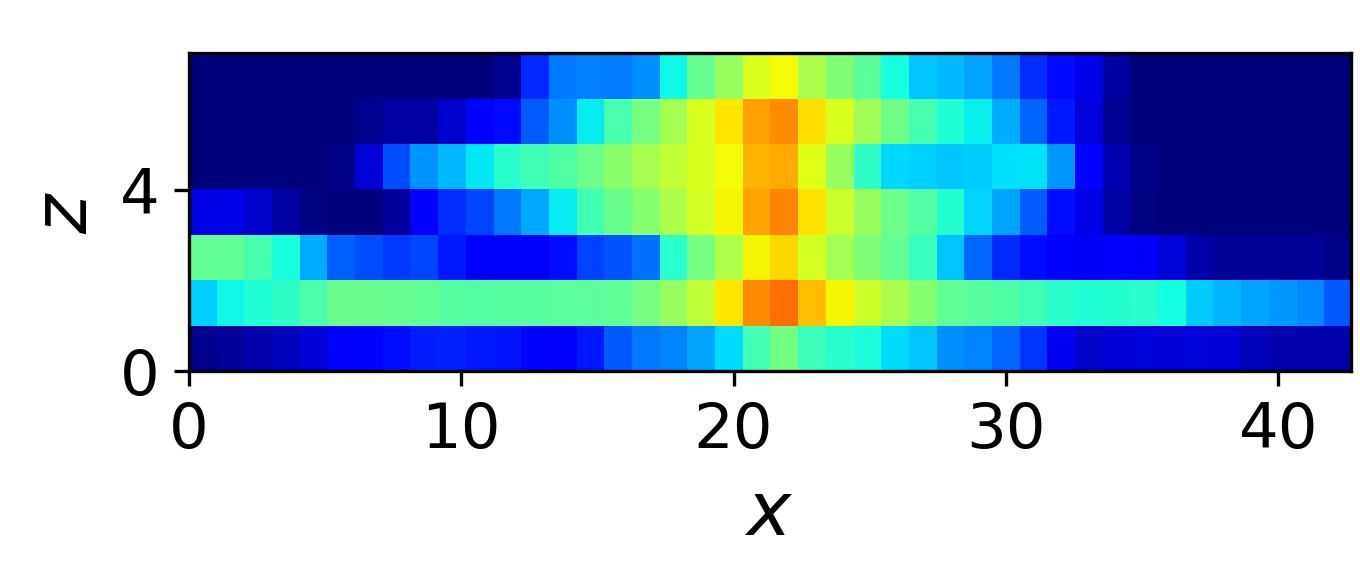}}
\subcaptionbox{Posterior 2 ($x$-$z$, sim)}{
\includegraphics[height=0.135\linewidth]{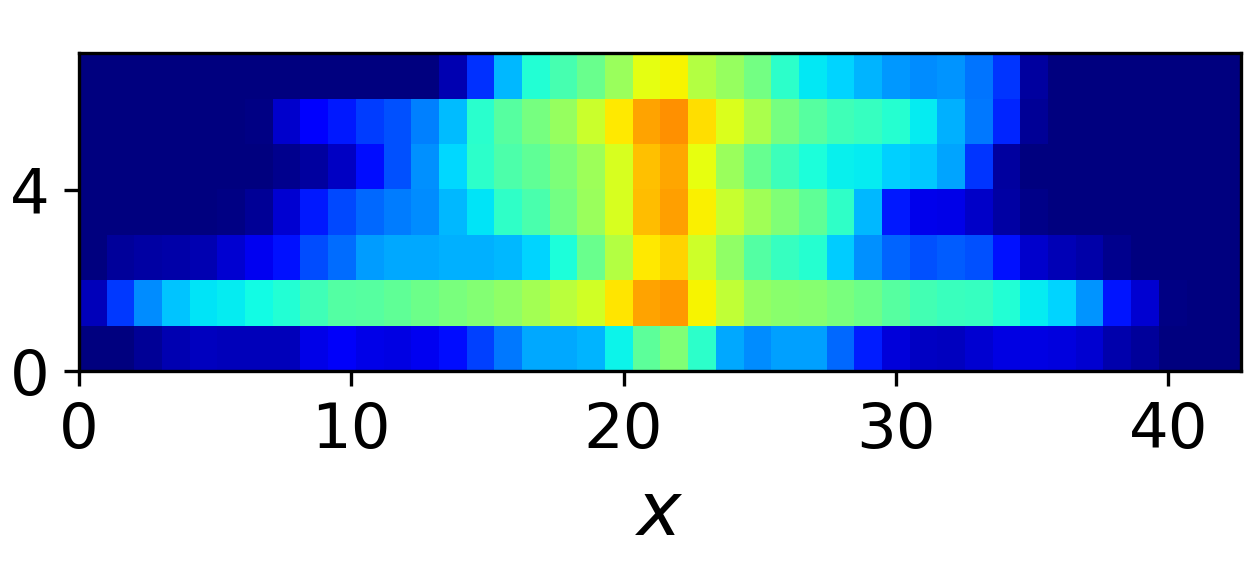}}
\subcaptionbox{Posterior 3 ($x$-$z$, sim)}{
\includegraphics[height=0.135\linewidth]{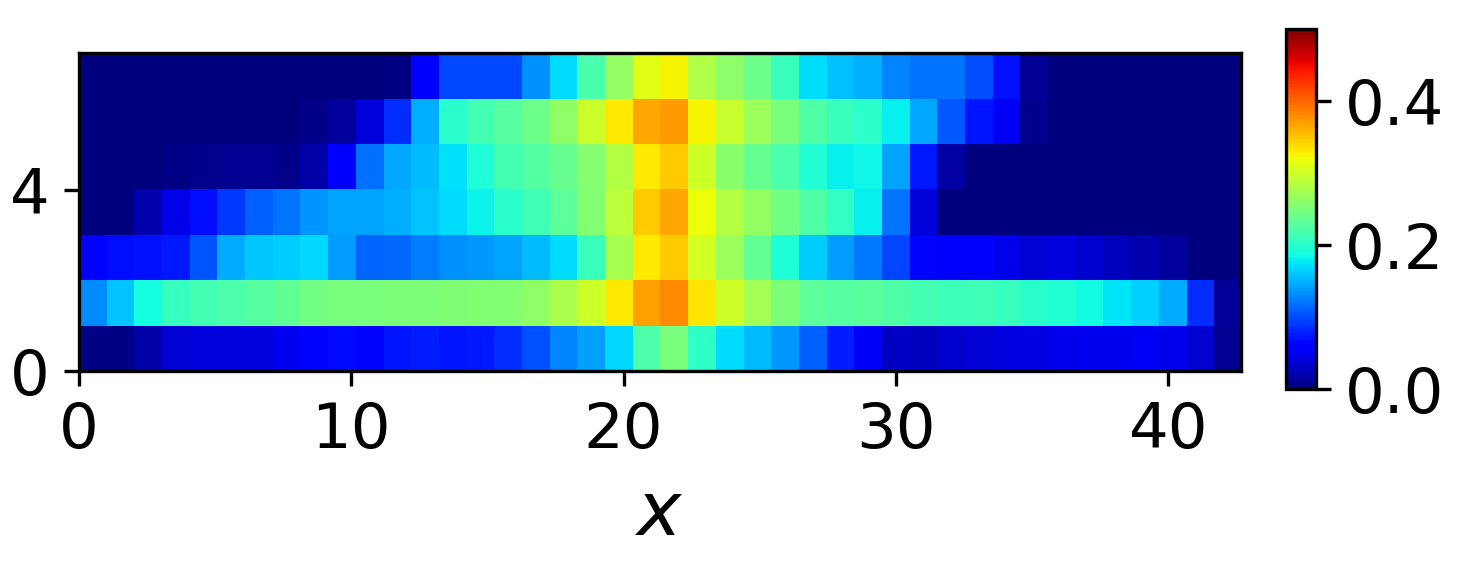}}

\subcaptionbox{Posterior 1 ($x$-$z$, surr)}{
\includegraphics[height=0.135\linewidth]{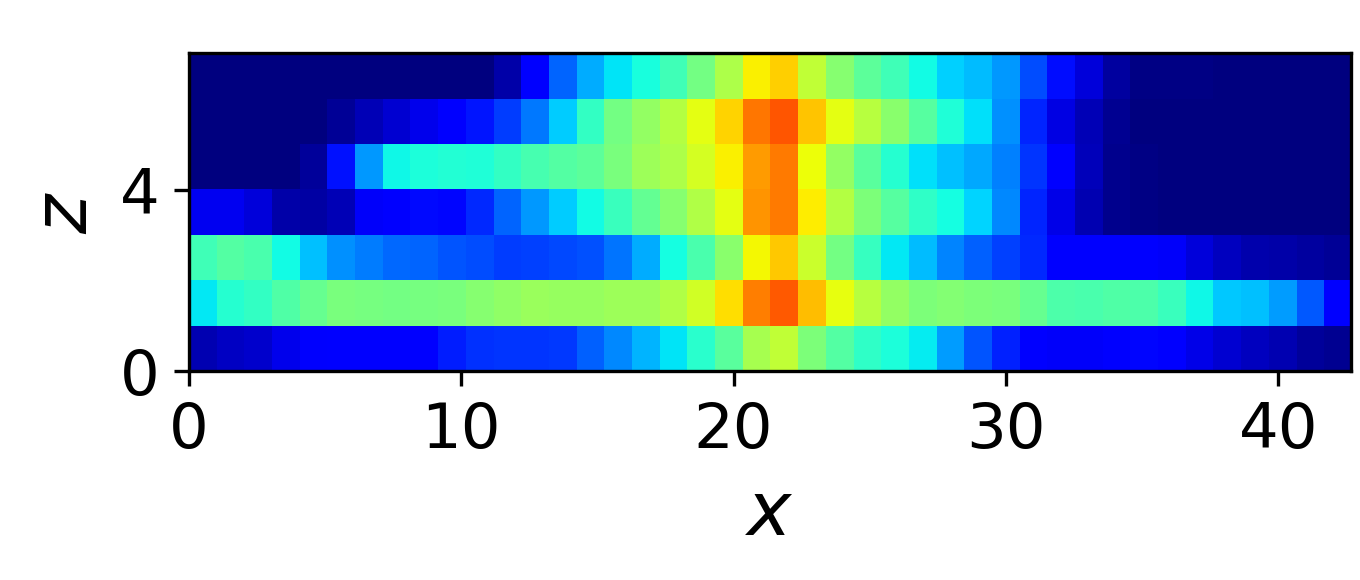}}
\subcaptionbox{Posterior 2 ($x$-$z$, surr)}{
\includegraphics[height=0.135\linewidth]{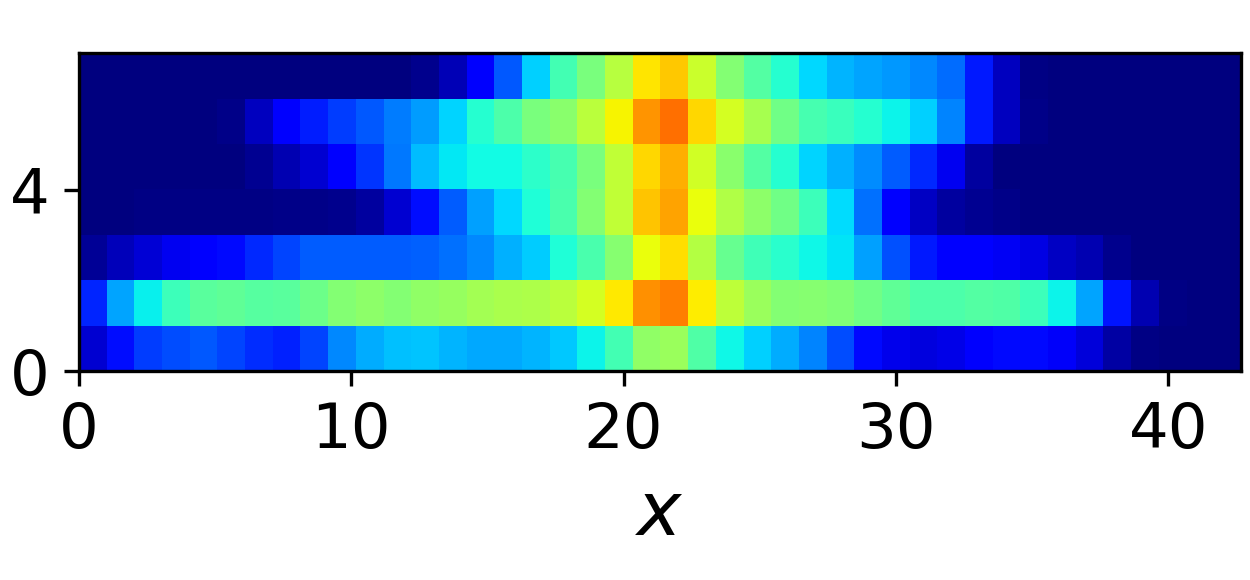}}
\subcaptionbox{Posterior 3 ($x$-$z$, surr)}{
\includegraphics[height=0.135\linewidth]{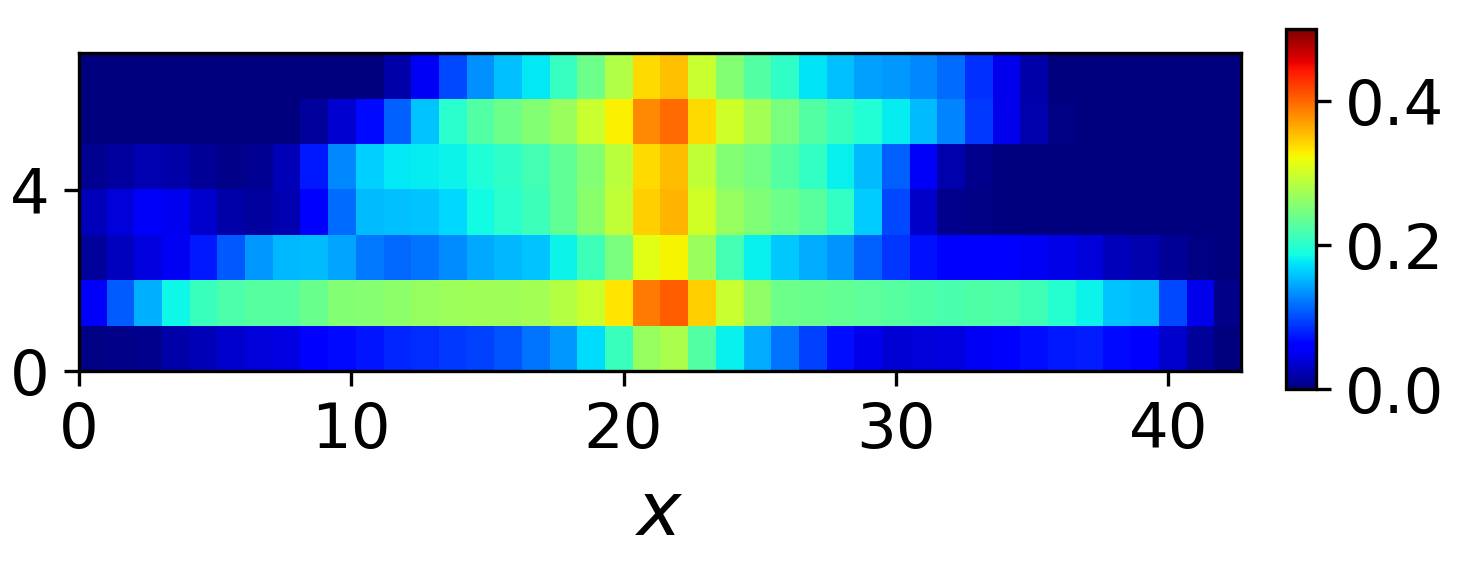}}

\caption{Simulation and surrogate model results for saturation at 1~year for posterior realizations shown in Figure~\ref{fig:logk_posterior}. Surrogate model results on row~2 correspond to simulation results on row~1, and surrogate model results on row~4 correspond to simulation results on row~3.}
\label{fig:plume_of_logk_posterior}
\end{figure}

As a final assessment of the applicability of the seismic surrogate model, we simulate the posterior realizations in Figure~\ref{fig:logk_posterior}(e)-(g) and then apply the filtering procedure to obtain interpreted seismic results. Saturation fields at 1~year are shown in Figure~\ref{fig:plume_of_logk_posterior}. The upper two rows show $y$-$z$ cross sectional results from simulation and surrogate model predictions, and the lower two rows are analogous results for $x$-$z$ cross sections. As expected, all of these fields resemble, to some degree, the true results in Figure~\ref{fig:plume_true}(c) and (d). There is also clear correspondence between the simulation results and the surrogate model results, though slight differences are evident. 

We note finally that, if the surrogate model is not sufficiently accurate for predictions with posterior realizations, some amount of retraining could be performed. The new training realizations in this case would be generated by sampling from the posterior distributions of the metaparameters. The MCMC history matching would then be repeated, in this case with a surrogate model that is more accurate in the region of interest. Such a workflow would be quite manageable given the speed of the surrogate models used in this work.

\section{Concluding remarks}
\label{sec:conclusion}

In this work, we implemented a deep-learning-based framework for history matching of geological carbon storage operations when both 4D seismic and monitoring well data are available. Our emphasis was on early time predictions, which are particularly important because uncertainty is high and unexpected behavior must be quickly identified, though our workflow is also applicable for longer time frames. Two different fit-for-purpose deep learning surrogate models were constructed -- one for predicting interpreted (coarse-scale) 4D seismic saturation fields, and one for predicting monitoring well data. Both networks accept high-resolution geomodels as input. The network for interpreted 4D seismic involves a 3D U-Net architecture, while the network for monitoring well data applies a 1D U-Net. Both networks have multiple output channels, with each channel corresponding to a different time step. The two specialized networks are simpler to construct and less time consuming to train than a single overall network able to provide global high-fidelity predictions.  

The training samples for the two surrogate models were obtained by performing high-fidelity flow simulation, using the GEOS simulator, on 3500 geomodel realizations. These realizations were characterized by different geological scenario parameters (referred to as metaparameters), resulting in a high degree of variability. The interpreted seismic data used for training were constructed through application of a filtering procedure, which provides results at the scale informed by 4D seismic data. The performance of the surrogate models was evaluated on a test set of 500 new geomodels. Both surrogates were found to provide accurate predictions and both were shown to capture a variety of flow behaviors, as can occur given the high degree of prior uncertainty considered.

The deep learning surrogate models were then applied for data assimilation using a hierarchical MCMC method. History matching results were constructed using only monitoring well data and using both monitoring well and interpreted 4D seismic data. The latter entailed either (binary) interpreted plume location data or interpreted saturation data. The impact of either type of 4D seismic data, in terms of reducing posterior uncertainty in the metaparameters over that achieved using only monitoring well data, was clearly demonstrated. Predictions for CO$_2$ plume shape and extent were also shown to be enhanced through inclusion of seismic data. A major advantage of our methodology is that the benefit of each data type can be quantified, and different strategies for combining highly-resolved local (monitoring well) data and coarse-resolution global (4D seismic) data can be quickly evaluated and thus optimized.

There are several topics that should be considered in future research. In this study, interpreted seismic saturation fields, assumed to be provided via petro-elastic inversion, were utilized. A higher degree of consistency would be achieved by working directly with seismic data rather than with interpreted data. In this case, the joint assimilation of both monitoring and seismic data could be addressed. Other data types, including pressure and surface displacement (which will require coupled flow-geomechanics simulations), are also expected to be informative and should be considered. Strategies for weighting and optimally combining these multiple data types, as well as error models, will also require investigation. Finally, the methodology developed here should be extended as necessary and then applied to practical storage operations.

\section*{Acknowledgments}
We are grateful to the Stanford Center for Carbon Storage for funding. We thank the SDSS Center for Computation for providing the HPC resources used in this work. We also thank Yifu Han for providing the hierarchical MCMC code and for assistance in its use, Oleg Volkov for assistance with GEOS, and Tapan Mukerji for discussions on interpreted 4D seismic data.









\bibliographystyle{elsarticle-harv} 
\bibliography{Ref}

\end{document}